\documentclass[12pt]{article}

\usepackage[margin=1in]{geometry}

\usepackage{setspace}

\usepackage{graphicx}

\usepackage{enumitem}

\usepackage{microtype}
\usepackage{subfigure}
\usepackage{booktabs}
\usepackage{comment}
\usepackage{wrapfig}
\usepackage{arydshln}
\usepackage{enumitem}
\usepackage{multirow}
\usepackage{url}
\usepackage{natbib}
\usepackage{authblk}

\RequirePackage[colorlinks,citecolor=blue,linkcolor=blue,urlcolor=blue,pagebackref]{hyperref}

\usepackage{amsmath}
\usepackage{amssymb}
\usepackage{mathtools}
\usepackage{amsfonts}
\usepackage{bm}
\usepackage{bbm}

\usepackage{algorithm}
\usepackage{algorithmic}

\def\1{\bm{1}}

\def\rmd{{\mathrm{d}}}

\def\rmp{{\mathrm{p}}}

\DeclareMathAlphabet{\mathsfit}{\encodingdefault}{\sfdefault}{m}{sl}
\SetMathAlphabet{\mathsfit}{bold}{\encodingdefault}{\sfdefault}{bx}{n}

\def\calA{{\mathcal{A}}}

\def\calC{{\mathcal{C}}}

\def\calF{{\mathcal{F}}}
\def\calG{{\mathcal{G}}}
\def\calH{{\mathcal{H}}}
\def\calI{{\mathcal{I}}}
\def\calJ{{\mathcal{J}}}

\def\calL{{\mathcal{L}}}

\def\calQ{{\mathcal{Q}}}
\def\calR{{\mathcal{R}}}

\def\calT{{\mathcal{T}}}

\def\calX{{\mathcal{X}}}

\def\bbE{{\mathbb{E}}}

\def\bbR{{\mathbb{R}}}

\def\tildeX{{\widetilde{X}}}

\newcommand{\Var}{\mathrm{Var}}

\newcommand{\p}[1]{\left(#1\right)}
\newcommand{\sqb}[1]{\left[#1\right]}
\newcommand{\cb}[1]{\left\{#1\right\}}

\newcommand{\bigp}[1]{\big(#1\big)}

\newcommand{\Bigp}[1]{\Big(#1\Big)}

\newcommand{\abs}[1]{\left|#1\right|}

\newcommand{\iid}{ \stackrel{\mathrm{i.i.d}}{\sim} }

\newcommand{\Prb}[1]{\mathbb{P}\left[#1\right]}

\mathtoolsset{showonlyrefs}
\usepackage{amsthm}

\theoremstyle{plain}

\newtheorem{theorem}{Theorem}[section]

\newtheorem{corollary}[theorem]{Corollary}
\newtheorem{proposition}[theorem]{Proposition}

\newtheorem{assumption}{Assumption}[section]
\newtheorem*{example}{Example}
\newtheorem*{remark}{Remark}

\usepackage[textsize=tiny]{todonotes}
\usepackage{multirow}
\usepackage{wrapfig}
\usepackage{subfigure}

\usepackage{xcolor}
\newcount\Comments  
\newcommand{\kibitz}[2]{\ifnum\Comments=1\textcolor{#1}{#2}\fi}

\usepackage{listings}

\usepackage[capitalize,noabbrev]{cleveref}

\allowdisplaybreaks

\title{Prediction-Powered Causal Inference by Automatic Debiased Machine Learning and Semi-Supervised Riesz Regression}

\author{Masahiro Kato\thanks{Email: \texttt{mkato-csecon@g.ecc.u-tokyo.ac.jp}}$\,$}

\affil{Data Analytics Department, Mizuho-DL Financial Technology, Co., Ltd.}

\date{\today}

\begin{document}

\maketitle 

\begin{abstract}
This study investigates semiparametric efficient estimation of causal and structural parameters in a semi-supervised setting. In our setting, unlabeled auxiliary regressors are available in addition to labeled observations consisting of outcomes and regressors. Our goal is to construct estimators of causal and structural parameters whose asymptotic variances are smaller than those of estimators constructed using only labeled data. We refer to this framework as \emph{prediction-powered causal inference} (PPCI). We first derive the efficient influence function and the efficiency bound, which imply that the use of auxiliary regressors can attain a smaller asymptotic variance than the efficiency bound attainable from labeled observations alone. Then, by combining the efficient influence function with the debiased machine learning (DML) framework, we propose methods that we call \emph{DML-PPCI}. If we construct an estimating-equation estimator, we refer to the method as EE-DML-PPCI; if we construct a targeted-learning estimator, we refer to the method as TMLE-DML-PPCI. The asymptotic variances of both estimators match our derived efficiency bound. In the construction of the estimators, estimation of the efficient influence function plays an important role. In our study, the efficient influence function is also a Neyman orthogonal score, which depends on the Riesz representer and the regression function. For Riesz representer estimation, we develop semi-supervised generalized Riesz regression with convergence rate guarantees.
\end{abstract}


\section{Introduction}
This study investigates efficient causal parameter estimation in a semi-supervised setting. The estimation targets are causal parameters defined via functionals of regression functions, including the average treatment effect (ATE), the average marginal effect (AME), and the average policy effect (APE) as special cases. We aim to construct estimators of these causal parameters with smaller variances by using an auxiliary unlabeled dataset of regressors, in addition to a conventional labeled dataset containing outcomes and regressors. This setup is closely related to the literature on prediction-powered inference \citep{Ilker2024predictionpowered}, but our goal is to estimate causal parameters. Hence, we refer to this framework as prediction-powered causal inference (PPCI). We show that, by using such an auxiliary unlabeled dataset, we can construct estimators whose asymptotic variances are smaller than those of estimators constructed using only the labeled dataset.

Efficient estimation of causal parameters has been a core interest in causal inference. To discuss efficiency, we consider the asymptotic efficiency bound, also called the semiparametric efficiency bound or the H\'{a}jek--Le Cam bound \citep{LeCam1986asymptoticmethods,VanderVaart1998asymptoticstatistics}, which characterizes the theoretically best asymptotic variance among regular estimators. We refer to causal parameter estimators whose asymptotic variances attain the efficiency bound as asymptotically efficient estimators. Asymptotically efficient estimators yield accurate estimation of causal parameters in terms of asymptotic mean squared error and tight confidence intervals. Therefore, constructing asymptotically efficient estimators is a standard goal in causal parameter estimation.

In this study, we consider the semi-supervised setup and develop efficiency bounds and asymptotically efficient estimators for this setup. The efficiency bound has been intensively studied in the standard setup with only labeled data, while it has not been fully studied in settings where unlabeled data can be used for general causal and structural parameters represented as functionals of regression functions. In this study, we show that if such unlabeled data are used appropriately, then we can construct estimators whose asymptotic variances are smaller than those of estimators constructed without using unlabeled data.

For example, consider estimating the effect of a new medicine in clinical trials. If population characteristics of patients without outcomes are available in addition to clinical trial data, then these unlabeled patient characteristics provide information about the regressor distribution over which the causal parameter is averaged. This information reduces the regressor-averaging component of the efficiency bound, even though it does not reduce the conditional outcome-noise component. This result shows that efficiency gains can arise from data that contain no outcome information, provided that such data inform the regressor distribution over which the target functional is averaged. Such approaches can be understood as semi-supervised causal inference or prediction-powered inference for causal inference.

In constructing asymptotically efficient estimators, we propose debiased machine learning (DML)-PPCI. DML is a framework for constructing asymptotically efficient estimators through Neyman orthogonal scores. Several approaches can be used in DML and in more general constructions of asymptotically efficient estimators. In this study, among them, we focus on estimating equations and targeted maximum likelihood (TMLE) \citep{VanderVaart2002semiparametricstatistics,vanderLaan2011targetedlearning,vanderLaan2006targetedmaximum}. We refer to DML-PPCI based on the estimating-equation approach as EE-DML-PPCI, and we refer to DML-PPCI based on TMLE as TMLE-DML-PPCI. For these estimators, we show asymptotic efficiency by proving that their asymptotic variances match our derived efficiency bounds.

\subsection{Contribution}
We summarize our contributions. First, we formulate PPCI under the two-sample and one-sample scenarios within the automatic DML (ADML) framework. Second, we derive the efficient influence functions and asymptotic efficiency bounds in our semi-supervised setup. Third, we construct asymptotically efficient estimators whose asymptotic variances match the derived efficiency bound. Fourth, we develop a semi-supervised version of generalized Riesz regression for estimating the Riesz representer appearing in the Neyman orthogonal score.

\paragraph{Problem formulation.} In the problem formulation, we define our parameters of interest as functionals of regression functions, as in ADML \citep{Chernozhukov2022automaticdebiased}. This framework covers causal and structural parameters that can be written as functionals of regression functions, including ATE, AME, and APE. In addition, we define the data-generating process (DGP) using the two-sample and one-sample scenarios \citep{Niu2016theoreticalcomparisons,Kato2025puate}. In the two-sample scenario, we assume that there exist two independent datasets, while in the one-sample scenario, we assume that there exists one dataset and that, from this dataset, we can observe labeled data. The two-sample scenario is closely related to the stratified sampling scheme \citep{Wooldridge2001asymptoticproperties,Lancaster1996casecontrolstudies}, while the one-sample scenario is closely related to the missing-value literature \citep{Rubin1974estimatingcausal,Kennedy2020efficientnonparametric}.

\paragraph{Efficiency bound.} For this setup, we derive the efficiency bounds, which can be computed from the efficient influence function. The efficiency bounds depend on the DGP. Therefore, for the two-sample and one-sample scenarios, we derive the corresponding efficiency bounds separately \citep{Uehara2020offpolicy}. We show that using an auxiliary unlabeled dataset can reduce the asymptotic variance compared with the case where only the labeled dataset is used. The efficiency bound implies the possibility of an efficiency gain, and feasibility is confirmed by constructing estimators whose asymptotic variances match the efficiency bounds.

\paragraph{Asymptotically efficient estimators.}
We refer to our estimation method as DML-PPCI. We develop an estimating-equation version, called EE-DML-PPCI, and a targeted maximum likelihood version, called TMLE-DML-PPCI. We show that these estimators have asymptotic variances matching the efficiency bounds derived in this study. Therefore, these estimators are asymptotically efficient \citep{Schuler2024introductionmodern}.

\paragraph{Semi-supervised generalized Riesz regression.} In DML-PPCI, we need to estimate the efficient influence function. The efficient influence function in our setup depends on two nuisance parameters, the regression function and the Riesz representer. To estimate the Riesz representer, we employ Riesz regression proposed by \citet{Chernozhukov2022automaticdebiased} or its generalization, generalized Riesz regression, proposed by \citet{Kato2026aunified}. Note that, under some restrictions on Riesz representer models, Riesz regression is mathematically equivalent to covariate balancing \citep{BrunsSmith2025augmentedbalancing,Zhao2019covariatebalancing}. We extend generalized Riesz regression to the semi-supervised setup so that an unlabeled dataset is also used to estimate the Riesz representer. This extension can be interpreted as providing an implementable nuisance-parameter estimation method for prediction-powered causal inference. For semi-supervised generalized Riesz regression, we provide convergence rates, which cover finite pseudo-dimension classes and deep ReLU sieves, including the H\"{o}lder-smooth case, the unbounded-support case, and the approximate low-dimensional manifold case. We then translate these Riesz representer rates into sufficient conditions for the product-rate requirement in DML-PPCI.

\subsection{Related Work}
This study is closely related to asymptotic efficiency theory and semi-supervised learning.

Asymptotic efficiency theory aims to construct estimators whose asymptotic variances match the asymptotic efficiency bound, especially in the sense of the semiparametric efficiency bound or the H\'{a}jek--Le Cam asymptotic efficiency bound. Various approaches and techniques have been proposed, such as estimating equations, TMLE, and sample splitting \citep{Klaassen1987consistentestimation}. The DML framework organizes these approaches by introducing Neyman orthogonal scores and cross-fitting \citep{Chernozhukov2018doubledebiased}. The ADML framework is a generalization of DML that deals with many causal and structural parameters written as functionals of a regression function and the corresponding Riesz representer. Under the ADML framework, even if we do not know the closed form of the efficient influence function or the Neyman orthogonal score, we can estimate it using Riesz regression \citep{Chernozhukov2022automaticdebiased}, which is related to semiparametric and sieve Riesz modeling \citep{Chen2015sievewald,Chen2015sievesemiparametric} and density-ratio estimation \citep{Sugiyama2011densityratio}.

Semi-supervised learning studies how labeled and unlabeled data can be combined to improve learning \citep{Zhu2005semisupervised,Chapelle2006semisupervisedlearning}. Covariate-shift methods are a variant of this setup and use information on a target covariate distribution to improve prediction or estimation under distributional changes \citep{Shimodaira2000improvingpredictive}. \citet{Angelopoulos2023predictionpowered} proposes prediction-powered inference as a framework for using machine-learning predictions together with gold-standard data to conduct valid inference. This idea has already been extended to causal inference by studies such as \citet{Cadei2026predictionpowered} and \citet{Ilker2024predictionpowered}. However, this study differs from these studies in its focus. We derive the semiparametric efficiency bound for causal and structural regression functionals when the auxiliary observations are unlabeled regressors and then construct ADML-type estimators that attain this bound.

Our study is also related to the literature on semi-supervised regression. \citet{Azriel2022semisupervised} studies best linear approximation under misspecification and shows that unlabeled regressors can improve the asymptotic variance of least-squares-type estimators when nonlinear features of the conditional mean interact with the marginal regressor distribution. \citet{Wasserman2007statisticalanalysis} analyzes semi-supervised regression through minimax theory and shows that unlabeled data do not automatically improve rates, especially for graph-Laplacian regularization. They emphasize that rate improvements require assumptions that connect the regression function and the regressor distribution. We use this insight to state primitive Riesz representer rates only under explicit smoothness, tail, or low-dimensional-structure conditions.

For estimating the Riesz representer, we employ Riesz regression. Note that there is a duality between Riesz regression and covariate balancing \citep{BrunsSmith2022outcomeassumptions,Zhao2019covariatebalancing,Hainmueller2012entropybalancing,Imai2013covariatebalancing,Zubizarreta2015stableweights,Sugiyama2008directimportance}, and based on this relationship, \citet{Kato2026aunified} develops generalized Riesz regression that formulates Riesz representer estimation as Riesz representer fitting under the Bregman divergence, which incorporates various existing methods such as Riesz regression, calibrated estimation \citep{Tan2019regularizedcalbrated}, tailored loss minimization \citep{Zhao2019covariatebalancing}, density-ratio estimation, and covariate balancing. In this study, we extend generalized Riesz regression to the semi-supervised setting so that we also utilize unlabeled datasets.

The convergence analysis for semi-supervised generalized Riesz regression is closely related to density-ratio estimation under the Bregman divergence \citep{Zheng2022anerror,Kato2021nonnegativebregman,Kato2025rieszregression}. \citet{Zheng2022anerror} establishes non-asymptotic error bounds for density-ratio estimation with deep ReLU feedforward neural networks, including minimax-optimal rates up to logarithmic factors under H\"{o}lder smoothness, extensions to unbounded support, and rates under approximate low-dimensional manifold structure. We use their result to derive error rates for our proposed semi-supervised generalized Riesz regression.

The stratified sampling scheme plays an important role in our theoretical analysis \citep{Wooldridge2001asymptoticproperties}. When the labeled and unlabeled datasets are independent samples from different strata with fixed sampling proportions, the usual one-sample efficiency bound is not directly applicable. Closely related work includes off-policy evaluation and learning for external validity under covariate shift \citep{Uehara2020offpolicy} and DML for covariate shift \citep{Chernozhukov2025automaticdebiased,Kato2024doubledebiasedcovariateshift}.

This study builds on our previous studies \citep{Kato2026aunified,Uehara2020offpolicy,Kato2024activeadaptive,Kato2025semisupervised,Kato2024doubledebiasedcovariateshift}. In particular, the idea of asymptotic efficiency under stratified sampling is inspired by \citet{Uehara2020offpolicy}. An early version of PPCI is presented in \citet{Kato2024activeadaptive} for adaptive experiments. Compared with those studies, we consider a general class of regression-functional causal and structural parameters, whereas those studies focus on specific applications, such as policy evaluation or adaptive experimental design under covariate shift.

\section{Setup}
Let $Y \in \bbR$ be an outcome and let $X,\tildeX\in\calX$ be regressors, where $\calX$ is the regressor space. We observe the outcome $Y$ for $X$, but we do not observe the corresponding outcome for $\widetilde{X}$. Therefore, we refer to $W=(X,Y)$ as labeled data and $\widetilde{X}$ as unlabeled data. In applications, unlabeled data are often test data or evaluation data for which the chosen treatment or policy will be implemented.

\subsection{Parameter of Interest}
Our goal is to estimate a parameter of the form
\begin{align}
    \theta_0
    \coloneqq
    \bbE_{V_{0X}}\sqb{m(X,\gamma_0)}
    \coloneqq
    \int m(x,\gamma_0) \rmd V_{0X}(x),
    \label{eq:target-parameter}
\end{align}
where $\gamma_0(x)\coloneqq\bbE_{P_0}\sqb{Y\mid X=x}$ and $m(X,\gamma)$ is a known functional map. 

The distribution $V_{0X}$ is the marginal regressor distribution used to evaluate the target parameter. For simplicity, we assume that $V_{0X}$ has the density
\begin{align}
    v_{0X}(x)
    \coloneqq
    \kappa p_{0X}(x)+(1-\kappa)q_{0X}(x),
    \qquad
    \kappa\in(0,1).
    \label{eq:evaluation-density}
\end{align}
We assume that the constant $\kappa$ is known. It is distinct from the labeled sampling proportion $\rho = n/(n + m)$ introduced in Section~\ref{sec:efficiency}.

We define the parameter of interest as a functional of the regression function, which includes various causal parameters as special cases. We give examples below.
\begin{example}[Examples of parameters of interest]
\label{ex:parameterofinterest}
By specifying the parameter functional $m$, we obtain the following estimands.
\begin{itemize}
    \item \textbf{ATE.} Let $X=(D,Z)$, where $D\in\{0,1\}$ is a treatment and $Z$ is a vector of covariates. Let $\gamma_0(d,z)=\bbE\sqb{Y\mid D=d,Z=z}$. Define
    \begin{align*}
        m^{\mathrm{ATE}}(X,\gamma)=\gamma(1,Z)-\gamma(0,Z).
    \end{align*}
    Then, the ATE is defined as
    \[\theta^{\mathrm{ATE}}_0=\bbE_{V_{0Z}}\sqb{\gamma_0(1,Z)-\gamma_0(0,Z)}.\]
    \item \textbf{AME.} Let $X=(D,Z)$, where $D$ is a continuously distributed treatment or policy variable and $Z$ is a vector of covariates. Suppose that $\gamma(d,z)$ is differentiable in $d$. Define
    \begin{align*}
        m^{\mathrm{AME}}(X,\gamma)=\partial_d\gamma(D,Z).
    \end{align*}
    Then, the AME is defined as 
    \[\theta^{\mathrm{AME}}_0=\bbE_{V_{0X}}\sqb{\partial_d\gamma_0(D,Z)}.\] This parameter is the average derivative of the regression function under the evaluation regressor distribution $V_{0X}$.
    \item \textbf{APE.} Let $X=(D,Z)$, where $D$ is a treatment or policy variable and $Z$ is a vector of covariates. Let $\pi_1(d\mid z)$ and $\pi_0(d\mid z)$ be two policy rules, written as conditional probability mass functions or conditional densities with respect to a measure $\nu$. Define
    \begin{align*}
        m^{\mathrm{APE}}(X,\gamma)
        =
        \int
        \p{\pi_1(d\mid Z)-\pi_0(d\mid Z)}\gamma(d,Z)\rmd\nu(d).
    \end{align*}
    Then, the APE is defined as 
    \[\theta^{\mathrm{APE}}_0
        =
        \bbE_{V_{0Z}}\sqb{
        \int
        \p{\pi_1(d\mid Z)-\pi_0(d\mid Z)}\gamma_0(d,Z)\rmd\nu(d)
        }.\] 
    If the policies are deterministic and $D$ is discrete, this reduces to $m^{\mathrm{APE}}(X,\gamma)
        =
        \gamma(\pi_1(Z),Z)-\gamma(\pi_0(Z),Z)$, 
    and hence the APE is defined as 
    \[\theta^{\mathrm{APE}}_0
        =
        \bbE_{V_{0Z}}\sqb{
        \gamma_0(\pi_1(Z),Z)-\gamma_0(\pi_0(Z),Z)}.\] 
    \item \textbf{Covariate shift adaptation.} Let $m(X,\gamma)=\gamma(X)$. Then, the target parameter is the mean of the outcome under $V_{0X}$, namely $\bbE_{V_{0X}}\sqb{Y} = \bbE_{V_{0X}}\sqb{\gamma_0(X)}$.
\end{itemize}
\end{example}

\begin{remark}[Relation to the ADML notation]
    Although ADML is often written using full observed-data notation, this study focuses on functionals of the form $\bbE\sqb{m(X,\gamma_0)}$. This restriction covers the common ADML causal parameters used in our analysis and is important in the present semi-supervised setting because outcomes are not observed for the unlabeled sample. Throughout this study, the plug-in component is evaluated as $m(X,\gamma)$ on labeled regressors and as $m(\tildeX,\gamma)$ on unlabeled regressors. Functionals that directly require outcomes inside the plug-in component require additional nuisance functions and a different efficiency analysis.
\end{remark}

\subsection{Observations}
In this study, we consider two different sampling schemes (DGPs): the one-sample scenario and the two-sample scenario.

\paragraph{One-sample scenario.}
In this scenario, there is a potential complete dataset
\begin{align*}
    \cb{\p{X^*_k,Y^*_k}}^N_{k=1}.
\end{align*}
Labeling then occurs. For each $k=1,2,\dots,N$, let $S_k\in\{1,0\}$ be a labeling indicator: if $S_k=1$, then $Y^*_k$ is observed; otherwise, $Y^*_k$ is missing. Therefore, we observe the dataset
\begin{align*}
    \cb{\p{X_k,S_k,Y_k}}^N_{k=1},
\end{align*}
where
\begin{align*}
    Y_k=S_kY^*_k+(1-S_k)\mathrm{NA},
\end{align*}
and $\mathrm{NA}$ denotes a missing value. In this case, the labeled and unlabeled datasets are
\begin{align*}
    \text{(labeled dataset)} &\cb{(X_k,Y_k)\mid S_k=1,\ k=1,2,\dots,N},\\
    \text{(unlabeled dataset)} &\cb{\tildeX_k=X_k\mid S_k=0,\ k=1,2,\dots,N},
\end{align*}
whose sample sizes are $n=\sum^N_{k=1}S_k$ and $m=\sum^N_{k=1}(1-S_k)$, respectively.

\paragraph{Two-sample scenario.}
We observe the following two independent stratified datasets, which we call the labeled dataset and the unlabeled dataset:
\begin{align*}
    \text{(labeled dataset)} &\cb{(X_i,Y_i)}_{i=1}^n,\\
    \text{(unlabeled dataset)} &\cb{\tildeX_j}_{j=1}^m.
\end{align*}

\paragraph{Comparison.}
The one-sample scenario treats labeling as a selection process. In contrast, the two-sample scenario considers a setting in which the usual observations $(X_i,Y_i)$ are available and auxiliary observations $\tildeX_j$ are independently observed. This difference leads to slightly different approaches to efficiency analysis and different asymptotic regimes. Therefore, we consider the two cases separately.

\subsection{Goal}

Our goal is to estimate the parameter of interest $\theta_0$ efficiently from the observations. Here, efficiency means that the asymptotic variance of an estimator matches the asymptotic lower bound for regular estimators. This lower bound is called the H\'{a}jek--Le Cam asymptotic efficiency bound, and its semiparametric extension is often called the semiparametric efficiency bound.

It is known that if an estimator is regular and asymptotically linear for the efficient influence function, then the estimator is efficient. In many cases, such an estimator is constructed from the efficient influence function by solving an estimating equation or by using TMLE. Therefore, in our methodological and theoretical analysis, we start by deriving the efficiency bound, then propose estimators of $\theta_0$, and then show their asymptotic efficiency.

In the subsequent sections, since the two-sample scenario is closer to the standard setting in semi-supervised learning, we first introduce the efficiency bound and efficient estimators under the two-sample scenario. Then, we introduce the corresponding efficiency bound and efficient estimators under the one-sample scenario.

\paragraph{Notation.}
Let $\sigma^2_0(x)\coloneqq\Var(Y\mid X=x)$, where the variance is taken over the conditional distribution $P_0$ given $X=x$.
Let \(\|f\|_{P,2}\coloneqq \p{\bbE_{P_0}\sqb{f(X)^2}}^{1/2}\) and \(\|f\|_{Q,2}\coloneqq \p{\bbE_{Q_{0X}}\sqb{f\p{\tildeX}^2}}^{1/2}\). For applications with discrete treatments, let \(e_0(d\mid z)\coloneqq P_0(D=d\mid Z=z)\). Let \(r_{0X}(x)\coloneqq v_{0X}(x)/p_{0X}(x)\) be the density ratio between the evaluation and labeled regressor distributions for \(X\). When \(X=(D,Z)\), let \(r_{0Z}(z)\coloneqq v_{0Z}(z)/p_{0Z}(z)\).

\section{Efficiency Bound and Efficient Influence Function in the Two-Sample Scenario}
\label{sec:efficiency}

This section derives the semiparametric efficiency bound under the two-sample scenario, where two independent datasets are observed.

\subsection{Detailed DGP}

We define the DGP of the two-sample scenario more precisely. We assume that $W\coloneqq(X,Y)$ follows a distribution $P_0$, and let $P_{0X}$ be the marginal distribution of $X$. We also assume that $\tildeX$ follows a distribution $Q_{0X}$. For simplicity, we assume that $P_{0X}$ and $Q_{0X}$ have probability density functions $p_{0X}$ and $q_{0X}$ with respect to a common dominating measure. This assumption can be relaxed by defining the corresponding Radon--Nikodym derivatives with respect to appropriate dominating measures.

\paragraph{DGP.}
We observe the following two independent stratified datasets, which we call the labeled dataset and the unlabeled dataset:
\begin{align*}
    \text{(labeled dataset)}\quad &\{W_i\}_{i=1}^n \coloneqq \cb{(X_i,Y_i)}_{i=1}^n \iid P_0,\\
    \text{(unlabeled dataset)}\quad &\cb{\tildeX_j}_{j=1}^m \iid Q_{0X},
\end{align*}
where $W_i=(X_i,Y_i)$ is an independent copy of $W=(X,Y)$, and $\tildeX_j$ is an independent copy of $\tildeX$.

Let $N\coloneqq n+m$. We denote the sampling proportion of the labeled stratum by $\rho\coloneqq n/N\in(0,1)$. In the asymptotic analysis, $\rho$ is fixed by design. Thus, we consider a deterministic sequence of admissible total sample sizes for which $n=\rho N$ and $m=(1-\rho)N$ are integers, and we take the limit $N\to\infty$ along this sequence.

\paragraph{Parameter of interest.}
Under this setup, the target parameter can be decomposed as
\begin{align*}
    \theta_0
    =
    \bbE_{V_0}\sqb{m(X,\gamma_0)}
    =
    \kappa\theta_{P,0}+(1-\kappa)\theta_{Q,0},
\end{align*}
where
\begin{align*}
    \theta_{P,0}\coloneqq \bbE_{P_{0X}}\sqb{m(X,\gamma_0)},
    \qquad
    \theta_{Q,0}\coloneqq \bbE_{Q_{0X}}\sqb{m(\tildeX,\gamma_0)}.
\end{align*}

\subsection{Efficient Influence Function}
This section derives the efficient influence function, from which we obtain both the efficiency bound and the efficient estimators.
 Since one of the key components in the efficient influence function is the Riesz representer, we first introduce the Riesz representer and then derive the efficient influence function.

\paragraph{Riesz representer.}
In our parameter of interest, the efficient influence function depends on the Riesz representer, which is obtained by applying the Riesz representation theorem to the parameter functional. Let $\Gamma\subseteq L^2(P_{0X})$ be the regression space under consideration. Assume that the linear functional
\begin{align*}
    \calL_{0,\kappa}(\gamma)
    \coloneqq
    \bbE_{V_0}\sqb{m(X,\gamma)}
\end{align*}
is continuous on $\Gamma$ under the $L^2(P_{0X})$ norm. Then, by the Riesz representation theorem, there exists a unique $\alpha_{0,\kappa}\in\overline{\Gamma}\subseteq L^2(P_{0X})$ such that
\begin{align}
    \calL_{0,\kappa}(\gamma)
    =
    \bbE_{P_0}\sqb{\alpha_{0,\kappa}(X)\gamma(X)}
    \qquad
    \text{for all } \gamma\in\Gamma .
    \label{eq:ss-riesz}
\end{align}
This representer generally depends on the evaluation density $v_{0X}$. We can estimate the Riesz representer without obtaining a closed-form expression. For reference, we present representative Riesz representers below for several cases raised in Example~\ref{ex:parameterofinterest}.

\begin{example}[Examples of Riesz representers]
Recall that $r_{0Z}(z)\coloneqq v_{0Z}(z)/p_{0Z}(z)$ is the density ratio between the evaluation and labeled covariate distributions for $Z$.
\begin{itemize}
    \item \textbf{ATE.} The Riesz representer is given as
    \begin{align*}
        \alpha^{\mathrm{ATE}}_{0,\kappa}(D,Z)
        =
        r_{0Z}(Z)
        \p{\frac{\mathbbm{1}[D=1]}{e_0(1\mid Z)}-
        \frac{\mathbbm{1}[D=0]}{e_0(0\mid Z)}},
    \end{align*}
    where $e_0(D\mid Z)=P_0(D\mid Z)$ is the propensity score.

    \item \textbf{AME.} Suppose that $X=(D,Z)$, that $D$ is continuously distributed, and that $v_{0X}(d,z)$ is differentiable in $d$. Suppose also that the boundary term from integration by parts is zero. Then, the Riesz representer is given as
    \begin{align*}
        \alpha^{\mathrm{AME}}_{0,\kappa}(D,Z)
        =
        -
        \frac{\partial_d v_{0X}(D,Z)}{p_{0X}(D,Z)}.
    \end{align*}
    In the special case $V_{0X}=P_{0X}$, this reduces to
    \begin{align*}
        \alpha^{\mathrm{AME}}_{0,\kappa}(D,Z)
        =
        -\partial_d\log p_{0X}(D,Z),
    \end{align*}
    which is the usual Riesz representer for the average derivative \citep{Chernozhukov2022automaticdebiased}.

    \item \textbf{APE.} If $\pi_1(d\mid z)$ and $\pi_0(d\mid z)$ are two policy rules that are dominated by $e_0(d\mid z)$, then the Riesz representer is given as
    \begin{align*}
        \alpha^{\mathrm{APE}}_{0,\kappa}(D,Z)
        =
        r_{0Z}(Z)
        \frac{\pi_1(D\mid Z)-\pi_0(D\mid Z)}{e_0(D\mid Z)}.
    \end{align*}
    If the policies are deterministic and $D$ is discrete, this becomes
    \begin{align*}
        \alpha^{\mathrm{APE}}_{0,\kappa}(D,Z)
        =
        r_{0Z}(Z)
        \p{
        \frac{\mathbbm{1}[D=\pi_1(Z)]}{e_0(\pi_1(Z)\mid Z)}
        -
        \frac{\mathbbm{1}[D=\pi_0(Z)]}{e_0(\pi_0(Z)\mid Z)}
        }.
    \end{align*}

    \item \textbf{Covariate shift adaptation.} The Riesz representer is given as
    \begin{align*}
        \alpha^{\mathrm{CS}}_{0,\kappa}(X)
        =
        \frac{v_{0X}(X)}{p_{0X}(X)}.
    \end{align*}
\end{itemize}
\end{example}

\paragraph{Efficient influence function.}
Using the Riesz representer, we derive the efficient influence function.

\begin{assumption}[Regularity for the semi-supervised ADML efficiency bound]
\label{ass:ss-eff}
The following conditions hold.
\begin{enumerate}
    \item The evaluation regressor distribution is dominated by the labeled regressor distribution: $V_{0X}\ll P_{0X}$. In particular, if $\kappa<1$, then $Q_{0X}\ll P_{0X}$. Moreover, $v_{0X}(X)/p_{0X}(X)$ is bounded $P_{0X}$-almost surely.
    \item The regression satisfies $\gamma_0\in\Gamma$, and $\bbE_{P_0}\sqb{(Y-\gamma_0(X))^2}<\infty$.
    \item The map $\gamma\mapsto m(\cdot,\gamma)$ is linear, and $m(\cdot,\gamma_0)\in L^2(P_{0X})\cap L^2(Q_{0X})$.
    \item The Riesz representer $\alpha_{0,\kappa}$ in \eqref{eq:ss-riesz} exists and satisfies
    \begin{align*}
        \bbE_{P_0}\sqb{\alpha_{0,\kappa}(X)^2\sigma_0^2(X)}<\infty .
    \end{align*}
\end{enumerate}
\end{assumption}

The following theorem holds. The proof is given in the Appendix.

\begin{theorem}[Efficient influence functions under the two-sample scenario]
\label{thm:ss-eif}
Suppose that Assumption~\ref{ass:ss-eff} holds. Under the nonparametric model for the labeled distribution $P_0$ and the unlabeled regressor distribution $Q_{0X}$, the efficient influence functions for the labeled and unlabeled strata are
\begin{align}
    \psi^{\mathrm{TS}}_0(W)
    &\coloneqq \psi^{\mathrm{TS}}\p{W;\gamma_0,\alpha_{0,\kappa},\theta_{P,0}}
    \coloneqq
    \alpha_{0,\kappa}(X)\p{Y-\gamma_0(X)}
    +
    \kappa\p{m(X,\gamma_0)-\theta_{P,0}},
    \label{eq:ss-eif-labeled}
    \\
    \widetilde{\psi}^{\mathrm{TS}}_0\p{\tildeX}
    &\coloneqq \widetilde{\psi}^{\mathrm{TS}}\p{\tildeX;\gamma_0,\theta_{Q,0}}
    \coloneqq
    (1-\kappa)\p{m(\tildeX,\gamma_0)-\theta_{Q,0}}.
    \label{eq:ss-eif-unlabeled}
\end{align}
\end{theorem}

Because the data are sampled from two strata, the efficient influence function is a pair of stratum-specific influence functions. The first component corresponds to the labeled stratum $W=(X,Y)\sim P_0$, and the second component corresponds to the unlabeled stratum $\tildeX\sim Q_{0X}$.

Both influence functions are centered within their own strata:
\begin{align*}
    \bbE_{P_0}\sqb{\psi^{\mathrm{TS}}_0(W)}=0,
    \qquad
    \bbE_{Q_{0X}}\sqb{\widetilde{\psi}^{\mathrm{TS}}_0\p{\tildeX}}=0.
\end{align*}
\subsection{Efficiency Bound}
The stratum-specific efficient influence functions imply the following efficiency bound.

\begin{theorem}[Efficiency bound under the two-sample scenario]
\label{thm:ss-eff-bound}
Suppose that Assumption~\ref{ass:ss-eff} holds. Let $N=n+m$, and suppose that the stratified design fixes $n=\rho N$ and $m=(1-\rho)N$ along an admissible deterministic sequence for a constant $\rho\in(0,1)$. Then, the asymptotic variance of any regular estimator of $\theta_0$, under the $\sqrt N$ normalization, is bounded below by
\begin{align}
    V^{\mathrm{TS}}_0(\kappa,\rho)
    &\coloneqq
    \frac{1}{\rho}
    \bbE_{P_0}\sqb{\psi^{\mathrm{TS}}_0(W)^2}
    +
    \frac{1}{1-\rho}
    \bbE_{Q_{0X}}\sqb{\widetilde{\psi}^{\mathrm{TS}}_0\p{\tildeX}^2}.
    \label{eq:ss-bound-abstract}
\end{align}
Equivalently, we have
\begin{align}
    V^{\mathrm{TS}}_0(\kappa,\rho)
    &=
    \frac{1}{\rho}
    \bbE_{P_0}\sqb{\alpha_{0,\kappa}(X)^2\sigma_0^2(X)}
    +
    \frac{\kappa^2}{\rho}
    \operatorname{Var}_{P_{0X}}\bigp{m(X,\gamma_0)}
    +
    \frac{(1-\kappa)^2}{1-\rho}
    \operatorname{Var}_{Q_{0X}}\bigp{m(\tildeX,\gamma_0)}.
    \label{eq:ss-bound-expanded}
\end{align}
\end{theorem}

The first term in \eqref{eq:ss-bound-expanded} comes from the outcome residual and is governed by the labeled sample because only labeled observations contain $Y$. The second and third terms are regressor averaging components. The second term is associated with the labeled regressor distribution, and the third term is associated with the auxiliary regressor distribution. Thus, unlabeled regressors improve efficiency by reducing the noise in the regressor averaging part of the target, but they do not reduce the outcome residual component.

\subsection{Asymptotically Efficient Estimators}
\label{sec:dml-ppci}
Based on the derived efficiency bound and efficient influence functions, we construct estimators whose asymptotic variances match the efficiency bound. The estimators are based on the stratum-specific efficient score
\begin{align*}
    \psi^{\mathrm{TS}}_0(W)
    &=
    \alpha_{0,\kappa}(X)\p{Y-\gamma_0(X)}
    +\kappa\p{m\bigp{X,\gamma_0}-\theta_{P,0}},
    \\
    \widetilde{\psi}^{\mathrm{TS}}_0\p{\tildeX}
    &=(1-\kappa)\p{m\bigp{\tildeX,\gamma_0}-\theta_{Q,0}}.
\end{align*}
We propose two estimator constructions: an estimating-equation estimator and a TMLE estimator. We refer to the overall method as DML-PPCI and specify these two estimators as EE-DML-PPCI and TMLE-DML-PPCI. Both estimators use the same nuisance estimators: the regression function $\gamma_0$ and the Riesz representer $\alpha_{0,\kappa}$. If these nuisance estimators do not satisfy the Donsker condition, then we also apply cross-fitting to control the empirical process terms.

\paragraph{EE-DML-PPCI.}
Given estimators $\widehat{\gamma}$ and $\widehat{\alpha}$ of the nuisance parameters $\gamma_0$ and $\alpha_{0,\kappa}$, we construct an estimator as
\[
\widehat{\theta} \coloneqq
\frac{1}{n}\sum^n_{i=1}\p{
\widehat{\alpha}(X_i)\p{Y_i-\widehat{\gamma}(X_i)}
+\kappa m\bigp{X_i,\widehat{\gamma}}
}
+(1-\kappa)\frac{1}{m}\sum^m_{j=1}m\bigp{\tildeX_j,\widehat{\gamma}}.
\]
This estimator decomposes as $\widehat{\theta}=\widehat{\theta}_{\mathrm{L}}+\widehat{\theta}_{\mathrm{U}}$, where the labeled and unlabeled parts are
\begin{align*}
    \widehat{\theta}_{\mathrm{L}}
    \coloneqq
    \frac{1}{n}\sum^n_{i=1}\p{
    \widehat{\alpha}(X_i)\p{Y_i-\widehat{\gamma}(X_i)}
    +\kappa m\bigp{X_i,\widehat{\gamma}}
    },
    \qquad
    \widehat{\theta}_{\mathrm{U}}
    \coloneqq
    (1-\kappa)\frac{1}{m}\sum^m_{j=1}m\bigp{\tildeX_j,\widehat{\gamma}}.
\end{align*}
The labeled part $\widehat{\theta}_{\mathrm{L}}$ estimates $\kappa\theta_{P,0}$ through the labeled estimating equation, and the unlabeled part $\widehat{\theta}_{\mathrm{U}}$ estimates $(1-\kappa)\theta_{Q,0}$ from the unlabeled regressors. We also refer to this estimator as the augmented Riesz-weighted (ARW) estimator, which is a generalization of the augmented inverse probability weighting (AIPW) estimator.

\begin{remark}[Error product property]
Under Neyman orthogonality, the influence function often has the property that its bias can be written as the product of the errors of two nuisance parameters. Our derived influence function also has this property:
\begin{align}
    &\bbE_{P_0}\sqb{\alpha(X)\p{Y-\gamma(X)}}
    +\calL_{0,\kappa}(\gamma)-\theta_0
    =
    \bbE_{P_0}\sqb{\p{\alpha_{0,\kappa}(X)-\alpha(X)}\p{\gamma(X)-\gamma_0(X)}}.
    \label{eq:ss-drift-identity}
\end{align}
Thus, the leading population bias is a second-order product of the regression error and the Riesz representer error, and it is negligible when the product of the two nuisance estimation errors converges sufficiently fast.
\end{remark}

\paragraph{TMLE-DML-PPCI.} Next, we propose the TMLE version of DML-PPCI, which is a regression-adjustment estimator with a bias-corrected regression-function estimator. Let $\widehat\gamma$ and $\widehat{\alpha}$ be initial estimators of $\gamma_0$ and $\alpha_{0,\kappa}$. TMLE first updates the initial regression-function estimator $\widehat\gamma$ in the direction of the estimated Riesz representer and then plugs the updated regression-function estimator $\widehat\gamma^{(1)}$ into the semi-supervised target functional.

In our semi-supervised setup, the targeting step is calibrated by the same evaluation distribution that appears in the target functional. For example, under a Gaussian likelihood, we update the initial estimator as
\begin{align*}
    \widehat\gamma^{(1)}(x) = \widehat\gamma(x)+\widehat\varepsilon\widehat\alpha(x),
\end{align*}
where the fluctuation is given by
\begin{align}
    \widehat\varepsilon
    \coloneqq
    \frac{ \frac{1}{n}\sum^n_{i=1}
    \widehat\alpha(X_i)\p{Y_i-\widehat\gamma(X_i)}
    }{
    \kappa \frac{1}{n}\sum^n_{i=1}
     \widehat\alpha^2\bigp{X_i}
    +(1-\kappa)\frac{1}{m}\sum^m_{j=1}
    \widehat\alpha^2\bigp{\widetilde{X}_j}}.
    \label{eq:tmle-epsilon}
\end{align}
This update follows the same logic as density-ratio-corrected likelihood maximization under covariate shift \citep{Shimodaira2000improvingpredictive}. The one-dimensional fluctuation is fitted for the regressor law under which the target functional is evaluated. The numerator is computed from the labeled residuals because outcomes are observed only in the labeled sample, and the density-ratio adjustment is already contained in the estimated Riesz representer \(\widehat\alpha\). The denominator estimates the target law average of \(\widehat\alpha^2\).

Then, we obtain the TMLE-DML-PPCI estimator as the following plug-in estimator:
\begin{align}
    \widehat\theta^{\text{TS}}_{\mathrm{TMLE}}
    &\coloneqq
    \kappa \frac{1}{n}\sum^n_{i=1}
    m\p{X_i,\widehat\gamma^{(1)}}
    +(1-\kappa)\frac{1}{m}\sum^m_{j=1}
    m\p{\widetilde{X}_j,\widehat\gamma^{(1)}}.
    \label{eq:tmle-dml-ppci-estimator}
\end{align}
This construction is a semi-supervised analogue of Auto-TML in \citet{Chernozhukov2022automaticdebiased}.

\begin{remark}
More generally, we can consider the following link-scale fluctuation based on the likelihood:
\begin{align*}
    \operatorname{link}\p{\widehat\gamma^{(1)}(x)}
    =
    \operatorname{link}\p{\widehat\gamma(x)}
    +\widehat\varepsilon\widehat\alpha(x),
\end{align*}
where $\operatorname{link}$ is a link function.
\end{remark}

\paragraph{Algorithm.}
We summarize the procedures for EE-DML-PPCI and TMLE-DML-PPCI in Algorithm~\ref{alg:dml-ppci}. In the pseudocode below, we use cross-fitting to apply the proposed framework when Donsker-type conditions are not imposed on the nuisance estimators.

\begin{algorithm}[tb]
    \caption{DML-PPCI}
    \label{alg:dml-ppci}
    \begin{algorithmic}[1]
        \REQUIRE Labeled sample $\cb{(X_i,Y_i)}_{i=1}^n$, unlabeled sample $\cb{\tildeX_j}_{j=1}^m$, number of folds $K$, and a choice of method, either EE-DML-PPCI or TMLE-DML-PPCI.
        \ENSURE $\widehat\theta$.
        \STATE Split the labeled indices $[n]$ into folds $\calI_1,\ldots,\calI_K$ and the unlabeled indices $[m]$ into folds $\calJ_1,\ldots,\calJ_K$ when Donsker-type conditions are not imposed.
        \STATE For each $k=1,\ldots,K$, set $\calI_{-k}\coloneqq [n]\setminus \calI_k$ and $\calJ_{-k}\coloneqq [m]\setminus \calJ_k$.
        \FOR{$k=1,\ldots,K$}
            \STATE Estimate $\widehat\gamma_k$ using the labeled training observations $\cb{(X_i,Y_i)\colon i\in \calI_{-k}}$.
            \STATE Estimate $\widehat\alpha_k$ using the labeled and unlabeled training regressors $\cb{X_i\colon i\in \calI_{-k}}$ and $\cb{\tildeX_j\colon j\in \calJ_{-k}}$.
        \ENDFOR
        \IF{EE-DML-PPCI is used}
            \STATE Set $\widehat\gamma_k^*\coloneqq\widehat\gamma_k$ for every $k$.
            \STATE Compute $\widehat\theta_{\mathrm L}
                \coloneqq
                \frac{1}{n}\sum^K_{k=1}\sum_{i\in \calI_k}
                \sqb{
                    \widehat\alpha_k(X_i)\p{Y_i-\widehat\gamma_k(X_i)}
                    +
                    \kappa m(X_i,\widehat\gamma_k)
                }$. 
            \STATE Compute $\widehat\theta_{\mathrm U}
                \coloneqq
                \frac{1}{m}\sum^K_{k=1}\sum_{j\in \calJ_k}
                (1-\kappa)m(\tildeX_j,\widehat\gamma_k)$. 
            \STATE Set $\widehat\theta\coloneqq\widehat\theta_{\mathrm L}+\widehat\theta_{\mathrm U}$.
        \ENDIF
        \IF{TMLE-DML-PPCI is used}
            \FOR{$k=1,\ldots,K$}
    \STATE Compute $A_{N,k}
        \coloneqq
        \frac{1}{|\calI_k|}\sum_{i\in \calI_k}
        \widehat\alpha_k(X_i)\p{Y_i-\widehat\gamma_k(X_i)}$.
    \STATE Compute $D_{N,k}
        \coloneqq
        \kappa\frac{1}{|\calI_k|}\sum_{i\in \calI_k}
        \widehat\alpha_k^2(X_i)
        +
        (1-\kappa)\frac{1}{|\calJ_k|}\sum_{j\in \calJ_k}
        \widehat\alpha_k^2(\tildeX_j)$.
    \STATE Set $\widehat\varepsilon_k\coloneqq A_{N,k}/D_{N,k}$.
    \STATE Set $\widehat\gamma_k^*(x)\coloneqq \widehat\gamma_k(x)+\widehat\varepsilon_k\widehat\alpha_k(x)$.
\ENDFOR
        \ENDIF
        \STATE For each $k=1,\ldots,K$ and each $i\in \calI_k$, define $\widehat\varphi_i^{\mathrm L}
            \coloneqq
            \widehat\alpha_k(X_i)\p{Y_i-\widehat\gamma_k^*(X_i)}
            +
            \kappa m\p{X_i,\widehat\gamma_k^*}$. 
        \STATE For each $k=1,\ldots,K$ and each $j\in \calJ_k$, define $\widehat\varphi_j^{\mathrm U}
            \coloneqq
            (1-\kappa)m\p{\tildeX_j,\widehat\gamma_k^*}$. 
    \end{algorithmic}
\end{algorithm}

In the pseudocode, we split the datasets into $K$ subsets, where \(K\ge 2\) is a fixed integer. Partition the labeled indices \([n]\) into \(K\) folds \(\calI_1,\ldots,\calI_K\) and the unlabeled indices \([m]\) into \(K\) folds \(\calJ_1,\ldots,\calJ_K\). We take the folds to be balanced in the sense that \(|\calI_k|/n=|\calJ_k|/m=1/K+o(1)\). Write
\begin{align*}
    \calI_{-k}\coloneqq [n]\setminus \calI_k,
    \qquad
    \calJ_{-k}\coloneqq [m]\setminus \calJ_k.
\end{align*}
For each fold \(k\), estimate the nuisance functions using only observations in \(\calI_{-k}\) and \(\calJ_{-k}\). Denote the resulting estimators by
\begin{align*}
    \widehat\gamma_k,
    \qquad
    \widehat\alpha_k.
\end{align*}
The fitted nuisance functions are then evaluated only on the held-out fold. The residual \(Y_i-\widehat\gamma_k(X_i)\) is evaluated only for \(i\in \calI_k\), whereas the plug-in terms \(m(X_i,\widehat\gamma_k)\) and \(m(\tildeX_j,\widehat\gamma_k)\) are evaluated on the held-out labeled and unlabeled folds, respectively.

The cross-fitted EE-DML-PPCI estimator is
\begin{align}
    \widehat\theta^{\text{TS}}_{\mathrm{EE}}
    &\coloneqq
    \frac{1}{n}\sum^K_{k=1}\sum_{i\in \calI_k}
    \Bigp{
        \widehat\alpha_k(X_i)\p{Y_i-\widehat\gamma_k(X_i)}
        +\kappa m\p{X_i,\widehat\gamma_k}
    }
    +\frac{1-\kappa}{m}\sum^K_{k=1}\sum_{j\in \calJ_k}
    m\p{\tildeX_j,\widehat\gamma_k}.
    \label{eq:ee-dml-ppci-crossfit}
\end{align}
The cross-fitted TMLE-DML-PPCI estimator uses the fold-specific fluctuation
\begin{align}
    \widehat\varepsilon_k
    \coloneqq
    \frac{ \frac{1}{|\calI_k|}\sum_{i\in \calI_k}
    \widehat\alpha_k(X_i)\p{Y_i-\widehat\gamma_k(X_i)}
    }{
    \kappa \frac{1}{|\calI_k|}\sum_{i\in \calI_k}
     \widehat\alpha_k^2(X_i)
    +(1-\kappa)\frac{1}{|\calJ_k|}\sum_{j\in \calJ_k}
    \widehat\alpha_k^2\p{\widetilde{X}_j}}.
    \label{eq:tmle-epsilon-crossfit}
\end{align}
Using the fluctuation, we set \(\widehat\gamma_k^{(1)}(x)=\widehat\gamma_k(x)+\widehat\varepsilon_k\widehat\alpha_k(x)\), and obtain
\begin{align}
    \widehat\theta^{\text{TS}}_{\mathrm{TMLE}}
    &\coloneqq
    \frac{\kappa}{n}\sum^K_{k=1}\sum_{i\in \calI_k}m\p{X_i,\widehat\gamma_k^{(1)}}
    +
    \frac{1-\kappa}{m}\sum^K_{k=1}\sum_{j\in \calJ_k}m\p{\tildeX_j,\widehat\gamma_k^{(1)}}.
    \label{eq:tmle-dml-ppci-crossfit}
\end{align}
If the nuisance classes satisfy the required Donsker-type empirical-process conditions, the estimator can be constructed without sample splitting by estimating the nuisance functions on the full sample and evaluating the same score on the full sample.

\subsection{Nuisance Parameter Estimation}
\label{subsec:dml-ppci-nuisance}

\paragraph{Riesz representer estimation.}
We estimate the Riesz representer using semi-supervised generalized Riesz regression, which we propose in this study. See Section~\ref{sec:ss-grr} for details. Here, we briefly describe the fold-specific version used for DML-PPCI with cross-fitting. For a candidate regression function \(\gamma\in\Gamma\), define the training-sample linear functional
\begin{align}
    \widehat{\calL}^{(-k)}_{\kappa}(\gamma)
    \coloneqq
    \frac{\kappa}{|\calI_{-k}|}\sum_{i\in \calI_{-k}}m(X_i,\gamma)
    +
    \frac{1-\kappa}{|\calJ_{-k}|}\sum_{j\in \calJ_{-k}}m(\tildeX_j,\gamma).
    \label{eq:dml-training-linear-functional}
\end{align}
Given a convex Bregman generator \(g\), the fold-specific semi-supervised generalized Riesz regression estimator is
\begin{align}
    \widehat\alpha_k
    \in
    \arg\min_{\alpha\in\calH_n}
    \cb{
        \widehat{\calQ}^{(-k)}_{g,\kappa}(\alpha)
        +\lambda_{\alpha,n}Reg_{\alpha}(\alpha)
    },
    \label{eq:dml-ssgrr-fold}
\end{align}
where \(\calH_n\) is a user-chosen representer class, \(Reg_{\alpha}\) is a regularization functional, and
\begin{align}
    \widehat{\calQ}^{(-k)}_{g,\kappa}(\alpha)
    &\coloneqq
    \frac{1}{|\calI_{-k}|}\sum_{i\in \calI_{-k}}
    \Bigp{
        \partial g\p{\alpha(X_i)}\alpha(X_i)-g\p{\alpha(X_i)}
    }
    -
    \widehat{\calL}^{(-k)}_{\kappa}(\partial g\circ\alpha).
    \label{eq:dml-ssgrr-fold-objective}
\end{align}
The first term in \eqref{eq:dml-ssgrr-fold-objective} uses labeled regressors because the Riesz inner product is defined under \(P_{0X}\). The second term uses both labeled and unlabeled regressors because it estimates the target functional under \(V_{0X}=\kappa P_{0X}+(1-\kappa)Q_{0X}\).

\paragraph{Regression function estimation.}
The regression function is identified from labeled observations. For each fold \(k\), estimate \(\gamma_0(x)=\bbE\sqb{Y\mid X=x}\) by any estimation method trained on \(\calI_{-k}\). For example, with squared loss,
\begin{align}
\widehat\gamma_k
    \in
    \arg\min_{\gamma\in\calG_n}
    \cb{
        \frac{1}{|\calI_{-k}|}\sum_{i\in \calI_{-k}}\bigp{Y_i-\gamma(X_i)}^2
        +\lambda_{\gamma,n}Reg_\gamma(\gamma)
    },
    \label{eq:dml-regression-fold}
\end{align}
where $Reg_\gamma$ is a regularization functional. For binary or bounded outcomes, the squared loss can be replaced by an appropriate negative log-likelihood. The theoretical results for the estimators of $\theta_0$ below do not require a particular regression method. They only require mean-square convergence rates for \(\widehat\gamma_k\) and \(\widehat\alpha_k\). Unlabeled data can also be incorporated through semi-supervised learning or covariate shift adaptation \citep{Shimodaira2000improvingpredictive}. We leave the choice of semi-supervised regression method as application-specific, because the appropriate construction depends on the data-generating setting and the target parameter.

\section{Theoretical Properties}
\label{sec:theoreticalproperties}
For simplicity, we denote both $\widehat\theta^{\text{TS}}_{\mathrm{EE}}$ and $\widehat\theta^{\text{TS}}_{\mathrm{TMLE}}$ by $\widehat{\theta}$. 
This section establishes the asymptotic properties of estimators $\widehat{\theta}$ of $\theta_0$ constructed by the proposed DML-PPCI, covering EE-DML-PPCI and TMLE-DML-PPCI.

\subsection{Consistency}
We first show the consistency of $\widehat{\theta}$.
\begin{assumption}[Convergence condition for the regression function estimator]
\label{asm:reg_consistency}
It holds that
\begin{align*}\|\widehat\gamma-\gamma_0\|_{P,2} =o_p(1)\quad (N\to\infty).\end{align*}
\end{assumption}
Note that, since the density ratio $q_{0X}/p_{0X}$ is bounded under Assumption~\ref{ass:ss-eff}, this assumption also implies $\|\widehat\gamma-\gamma_0\|_{Q,2} =o_p(1)$.

\begin{assumption}[Convergence condition for the Riesz representer estimator]
\label{asm:riesz_consistency}
It holds that
\begin{align*}\|\widehat\alpha-\alpha_{0,\kappa}\|_{P,2} =o_p(1)\quad (N\to\infty).\end{align*}
\end{assumption}

The following theorem holds.
\begin{theorem}[Consistency of DML-PPCI]
\label{thm:dml-ppci-consistency}
Suppose that either Assumption~\ref{asm:reg_consistency} or Assumption~\ref{asm:riesz_consistency} holds. Then, we have
\begin{align*}
    \widehat\theta \xrightarrow{\rmp} \theta_0 \quad (N\to\infty).
\end{align*}
\end{theorem}

We do not require that both Assumptions~\ref{asm:reg_consistency} and \ref{asm:riesz_consistency} hold. Such a property is called double robustness \citep{Bang2005doublyrobust}. To show asymptotic normality, both Assumptions~\ref{asm:reg_consistency} and \ref{asm:riesz_consistency} need to hold.

\subsection{Asymptotic Properties}
\label{subsec:dml-ppci-asymptotics}
In addition to Assumptions~\ref{asm:reg_consistency} and \ref{asm:riesz_consistency}, we impose the following assumptions.

\begin{assumption}[Convergence rate condition]
\label{asm:conv_rate}
It holds that
\begin{align*}\|\widehat\alpha-\alpha_{0,\kappa}\|_{P,2}\|\widehat\gamma -\gamma_0\|_{P,2}=o_p(N^{-1/2})\quad (N\to\infty).\end{align*}
\end{assumption}

\begin{assumption}[Donsker condition or cross-fitting]
\label{asm:donsker}
The nuisance estimators $\widehat\alpha$ and $\widehat\gamma$ either satisfy the required Donsker-type stochastic equicontinuity conditions or are constructed via cross-fitting. Moreover, the corresponding score components have uniformly bounded \((2+\delta)\)-moments for some \(\delta>0\).
\end{assumption}

\begin{assumption}[Targeting calibration condition]
\label{asm:targeting-calibration}
The denominator $D_N$ in \eqref{eq:tmle-epsilon} is bounded away from zero, and
\begin{align*}
    L_N-D_N=o_p(1)\quad (N\to\infty),
\end{align*}
where
\begin{align*}
    D_N
    \coloneqq
    \kappa\frac{1}{n}\sum^n_{i=1}\widehat\alpha(X_i)^2
    +(1-\kappa)\frac{1}{m}\sum^m_{j=1}\widehat\alpha\bigp{\tildeX_j}^2,
    \qquad
    L_N
    \coloneqq
    \frac{\kappa}{n}\sum^n_{i=1} m(X_i,\widehat\alpha)
    +\frac{1-\kappa}{m}\sum^m_{j=1} m(\tildeX_j,\widehat\alpha).
\end{align*}
This condition equates the empirical second moment of $\widehat\alpha$ under the target regressor law with the plug-in derivative of the target functional along $\widehat\alpha$, and it is the condition under which the importance-weighted targeting update reproduces the orthogonal score.
\end{assumption}

The following theorem holds.
\begin{theorem}[Asymptotic linearity and efficiency of DML-PPCI]
\label{thm:dml-ppci-asymptotic}
Suppose that Assumptions~\ref{asm:reg_consistency}, \ref{asm:riesz_consistency}, \ref{asm:conv_rate}, and \ref{asm:donsker} hold. For TMLE-DML-PPCI, suppose in addition that Assumption~\ref{asm:targeting-calibration} holds. Then, we have
\begin{align}
    \widehat\theta-\theta_0
    =
    \frac{1}{n}\sum^n_{i=1}\psi^{\mathrm{TS}}_0(W_i)
    +
    \frac{1}{m}\sum^m_{j=1}\widetilde{\psi}^{\mathrm{TS}}_0(\tildeX_j)
    +o_p(N^{-1/2})\quad (N\to\infty).
    \label{eq:ee-asymptotic-linear}
\end{align}
Consequently, we have
\begin{align}
    \sqrt N\p{\widehat\theta -\theta_0}
    \xrightarrow{d}
    N\p{0,V^{\mathrm{TS}}_0(\kappa,\rho)}\quad (N\to\infty),
    \label{eq:ee-asymptotic-normal}
\end{align}
where \(V^{\mathrm{TS}}_0(\kappa,\rho)\) is the efficiency bound in Theorem~\ref{thm:ss-eff-bound}.
Thus, the estimators are regular and semiparametrically efficient.
\end{theorem}

\subsection{Efficiency Gain}
We now compare the semi-supervised bound with the labeled-only bound in the same-population case. Suppose that
\begin{align*}
    P_{0X}=Q_{0X}.
\end{align*}
Then, \(V_{0X}=P_{0X}\) for every \(\kappa\in[0,1]\), so changing \(\kappa\) does not change the estimand.

Let \(\alpha_{0,P}\) denote the Riesz representer for the ordinary labeled-population functional
\begin{align*}
    \gamma\mapsto \bbE_{P_0}\sqb{m(X,\gamma)}.
\end{align*}
In this same-population case, \(\alpha_{0,\kappa}=\alpha_{0,P}\) and \(\theta_{P,0}=\theta_{Q,0}=\theta_0\). Define
\begin{align*}
    A_0
    \coloneqq
    \bbE_{P_0}\sqb{\alpha_{0,P}(X)^2\sigma_0^2(X)},
    \qquad
    B_0
    \coloneqq
    \operatorname{Var}_{P_{0X}}\p{m(X, \gamma_0)}.
\end{align*}
Then, we have
\begin{align}
    V^{\mathrm{TS}}_0(\kappa,\rho)
    =
    \frac{A_0}{\rho}
    +
    \p{
        \frac{\kappa^2}{\rho}
        +
        \frac{(1-\kappa)^2}{1-\rho}
    }B_0 .
    \label{eq:ss-bound-same-pop}
\end{align}
The value of \(\kappa\) that minimizes \eqref{eq:ss-bound-same-pop} is
\begin{align*}
    \kappa^*=\rho.
\end{align*}
At \(\kappa=\rho\), the semi-supervised efficiency bound becomes
\begin{align}
    V^{\mathrm{TS}}_0(\rho,\rho)
    =
    \frac{A_0}{\rho}+B_0 .
    \label{eq:ss-bound-kappa-rho}
\end{align}
By contrast, an estimator that uses only the \(n\) labeled observations has the ordinary efficiency bound \(A_0+B_0\) under \(\sqrt n\) normalization in the standard setup of causal inference or DML.

Under the common \(\sqrt N\) normalization and the design identity \(n/N=\rho\), this becomes
\begin{align}
    V^{\mathrm{sup}}_0(\rho)
    =
    \frac{1}{\rho}(A_0+B_0).
    \label{eq:ss-supervised-bound}
\end{align}
Therefore, we have
\begin{align}
    V^{\mathrm{sup}}_0(\rho)-V^{\mathrm{TS}}_0(\rho,\rho)
    =
    \p{\frac{1}{\rho}-1}
    \operatorname{Var}_{P_{0X}}\p{m(X, \gamma_0)}
    \ge 0.
    \label{eq:ss-eff-gain}
\end{align}
This identity gives the efficiency gain from using the proposed method. Specifically, using unlabeled data reduces the variance stemming from the regressor-averaging component \(\operatorname{Var}_{P_{0X}}\p{m(X, \gamma_0)}\), while the residual outcome-noise component \(A_0\) remains inflated by \(1/\rho\) because outcomes are observed only in the labeled stratum.

If \(P_{0X}\ne Q_{0X}\), the parameter \(\kappa\) is part of the estimand because changing \(\kappa\) changes the target regressor distribution. In that case, \(\kappa\) should not be chosen by minimizing \eqref{eq:ss-bound-expanded} unless the scientific target is invariant to \(\kappa\).

\section{Semi-Supervised Generalized Riesz Regression}
\label{sec:ss-grr}
In this section, we develop a semi-supervised version of generalized Riesz regression, building on \citet{Kato2025directbias} and \citet{Kato2026aunified}. Riesz regression was proposed by \citet{Chernozhukov2022automaticdebiased} for estimating the Riesz representer $\alpha_0$. \citet{Kato2026aunified} generalizes Riesz regression using Bregman divergence minimization, building on related results in \citet{Zhao2019covariatebalancing} and \citet{BrunsSmith2025augmentedbalancing}.

\subsection{Bregman--Riesz Objective}
\label{subsec:ss-bregman-riesz-objective}
Let \(\calA\subset\bbR\) be an interval and let \(g\colon\calA\to\bbR\) be a differentiable strictly convex function. We write \(\partial g\) for its derivative. For a candidate representer \(\alpha\), we write \(u_\alpha\coloneqq\partial g\circ\alpha\) and \(b_\alpha\coloneqq u_\alpha\alpha-g(\alpha)\). 

\paragraph{Population objective.}
For $a_0,a\in\calA$, define the Bregman divergence by
\begin{align*}
    \text{BD}_g^{\dagger}(a_0\mid a)
    \coloneqq
    g(a_0)-g(a)-\partial g(a)(a_0-a).
\end{align*}
By ignoring terms irrelevant to the optimization over $\alpha$, the population semi-supervised Bregman--Riesz objective is
\begin{align}
    \calR_{g,\kappa}(\alpha)
    &\coloneqq
    \bbE_{P_{0X}}\sqb{\partial g\p{\alpha(X)}\alpha(X)-g\p{\alpha(X)}}
    -
    \kappa\bbE_{P_{0X}}\sqb{m(X,u_\alpha)}
    -
    (1-\kappa)\bbE_{Q_{0X}}\sqb{m(\tildeX,u_\alpha)}.
    \label{eq:ss-pop-bregman-riesz}
\end{align}

\paragraph{Empirical objective.}
The empirical objective is
\begin{align}
    \widehat{\calR}_{g,\kappa}(\alpha)
    &\coloneqq
    \frac{1}{n}\sum^n_{i=1}
    \bigp{
        \partial g\p{\alpha(X_i)}\alpha(X_i)-g\p{\alpha(X_i)}
        -\kappa m(X_i,u_\alpha)
    }
    -
    \frac{1-\kappa}{m}\sum^m_{j=1} m\p{\tildeX_j,u_\alpha}.
\label{eq:ss-emp-bregman-riesz}
\end{align}
Here, \(\widehat{\calL}_\kappa\) is the empirical evaluation functional
\begin{align*}
    \widehat{\calL}_\kappa(\gamma)
    \coloneqq
    \frac{\kappa}{n}\sum^n_{i=1} m(X_i,\gamma)
    +
    \frac{1-\kappa}{m}\sum^m_{j=1} m(\tildeX_j,\gamma),
\end{align*}
the full-sample analogue of \eqref{eq:dml-training-linear-functional}; in this notation \(\widehat{\calR}_{g,\kappa}(\alpha)=\frac{1}{n}\sum^n_{i=1} b_\alpha(X_i)-\widehat{\calL}_\kappa(u_\alpha)\).

Given a hypothesis class \(\calH_n\), we estimate \(\alpha_{0,\kappa}\) by
\begin{align}
    \widehat\alpha
    \in
    \arg\min_{\alpha\in\calH_n}
    \cb{
        \widehat{\calR}_{g,\kappa}(\alpha)+\lambda_n Reg_{\alpha}(\alpha)
    },
    \label{eq:ss-grr-estimator}
\end{align}
where $Reg_{\alpha}$ is some regularization functional.

\begin{remark}[Bregman--Riesz Identity]
    The objective in \eqref{eq:ss-pop-bregman-riesz} is a Bregman projection of the true Riesz representer. Indeed, from the Riesz representer theorem, we have $\calL_{0,\kappa}(u_\alpha)
    =
    \bbE_{P_{0X}}\sqb{\alpha_{0,\kappa}(X)u_\alpha(X)}$. 
Therefore, it holds that 
\begin{align}
    \calR_{g,\kappa}(\alpha)-\calR_{g,\kappa}(\alpha_{0,\kappa})
    =
    \bbE_{P_{0X}}\sqb{
        \text{BD}_g^{\dagger}\bigp{\alpha_{0,\kappa}(X)\mid \alpha(X)}
    }.
    \label{eq:ss-bregman-identity}
\end{align}
Thus, if \(\alpha_{0,\kappa}\in\calH_n\), the population minimizer of \(\calR_{g,\kappa}\) over \(\calH_n\) is \(\alpha_{0,\kappa}\). If \(\alpha_{0,\kappa}\notin\calH_n\), the population minimizer is the Bregman projection of \(\alpha_{0,\kappa}\) onto \(\calH_n\).
\end{remark}

\begin{remark}[Dual coordinates]
    It is often convenient to work in the dual coordinate. Let
\begin{align*}
    g^*(u)\coloneqq \sup_{a\in\calA}\{au-g(a)\}
\end{align*}
be the convex conjugate. If \(u=\partial g(a)\), then \(g^*(u)=au-g(a)\) and \(\partial g^*(u)=(\partial g)^{-1}(u)\). Let \(f\) be a dual-coordinate function and define
\begin{align*}
    \alpha_f(x)
    \coloneqq
    (\partial g)^{-1}\p{f(x)}.
\end{align*}
Then \eqref{eq:ss-emp-bregman-riesz} can be written as
\begin{align}
    \widehat{\calQ}_{g,\kappa}(f)
    &\coloneqq
    \frac{1}{n}\sum^n_{i=1} g^*\p{f(X_i)}
    -
    \widehat{\calL}_{\kappa}(f),
    \label{eq:ss-dual-objective}
\end{align}
up to the reparametrization \(\alpha=\alpha_f\). Hence, for a dual class \(\calF_n\), we have 
\begin{align}
    \widehat f
    \in
    \arg\min_{f\in\calF_n}
    \cb{
        \widehat{\calQ}_{g,\kappa}(f)+\lambda_n\Omega(f)
    },
    \label{eq:ss-dual-estimator}
\end{align}
where $\widehat\alpha=\alpha_{\widehat f}$. 
If \(\calF_n\) is linear in finite-dimensional parameters and \(g^*\) is convex, the dual problem is a convex empirical risk minimization problem even when the primal link \(\alpha_f=(\partial g)^{-1}\circ f\) is nonlinear.
\end{remark}

\begin{remark}[Reduction to density-ratio estimation]
\label{rem:ss-dre-special-case}
If \(m(X,\gamma)=\gamma(X)\), then \(\alpha_{0,\kappa}(x)=v_{0X}(x)/p_{0X}(x)\). For \(\kappa=0\), this is the density ratio \(q_{0X}(x)/p_{0X}(x)\), and \eqref{eq:ss-pop-bregman-riesz} becomes the standard objective of density-ratio estimation under the Bregman divergence. Hence, the present construction strictly generalizes direct density-ratio estimation under covariate shift. Note that density-ratio estimation does not necessarily require the existence of probability densities with respect to the Lebesgue measure. The density notation is a simplification, and the method can also be applied to other cases, such as discrete variables with probability mass functions.
\end{remark}

\subsection{Automatic Regressor Balancing}
\label{subsec:ss-arb}
The Bregman--Riesz objective automatically produces balancing equations when the dual coordinate is linear in a set of regressors \citep{Kato2026aunified}. For a candidate representer \(\alpha\) and a candidate regression function \(\gamma\), define the semi-supervised imbalance gap
\begin{align}
    \widehat{\Delta}_{\kappa}(\alpha,\gamma)
    &\coloneqq
    \frac{1}{n}\sum^n_{i=1} \alpha(X_i)\gamma(X_i)
    -
    \frac{\kappa}{n}\sum^n_{i=1} m(X_i,\gamma)
    -
    \frac{1-\kappa}{m}\sum^m_{j=1} m(\tildeX_j,\gamma).
    \label{eq:ss-imbalance-gap}
\end{align}
The population analogue is
\begin{align*}
    \Delta_{0,\kappa}(\alpha,\gamma)
    \coloneqq
    \bbE_{P_{0X}}\sqb{\alpha(X)\gamma(X)}-
    \calL_{0,\kappa}(\gamma).
\end{align*}
The Riesz representer is characterized by \(\Delta_{0,\kappa}(\alpha_{0,\kappa},\gamma)=0\) for all valid \(\gamma\).

Let \(\phi(x)=(\phi_1(x),\ldots,\phi_p(x))^\top\) be a dictionary and consider the dual-linear model
\begin{align}
    f_\beta(x)=\phi(x)^\top\beta,
    \qquad
    \alpha_\beta(x)=(\partial g)^{-1}\p{f_\beta(x)}.
    \label{eq:ss-dual-linear-model}
\end{align}
For \(q\in[1,\infty)\), set \(\Omega_q(\beta)=\|\beta\|_q^q/q\). The empirical estimator is
\begin{align}
    \widehat\beta
    \in
    \arg\min_{\beta\in\bbR^p}
    \cb{
        \frac{1}{n}\sum^n_{i=1} g^*\p{\phi(X_i)^\top\beta}
        -
        \widehat{\calL}_\kappa(\phi^\top\beta)
        +
        \lambda_n \Omega_q(\beta)
    },
    \qquad
    \widehat\alpha=\alpha_{\widehat\beta}.
    \label{eq:ss-dual-linear-estimator}
\end{align}

\begin{proposition}[Automatic semi-supervised regressor balancing]
\label{prop:ss-arb}
Assume that \eqref{eq:ss-dual-linear-estimator} has a minimizer and that the usual first-order optimality conditions hold. Then, for each \(j=1,\ldots,p\), there exists \(s_j\in\partial\p{|\beta_j|^q/q}|_{\beta_j=\widehat\beta_j}\) such that
\begin{align}
    \widehat{\Delta}_\kappa(\widehat\alpha,\phi_j)+\lambda_n s_j=0.
    \label{eq:ss-arb-kkt}
\end{align}
In particular, if \(\lambda_n=0\), then we have
\begin{align*}
    \widehat{\Delta}_\kappa(\widehat\alpha,\phi_j)=0,
    \qquad j=1,\ldots,p.
\end{align*}
If \(q=1\), then we have
\begin{align*}
    \max_{1\le j\le p}
    |\widehat{\Delta}_\kappa(\widehat\alpha,\phi_j)|
    \le \lambda_n.
\end{align*}
If \(q>1\), then we have
\begin{align*}
    |\widehat{\Delta}_\kappa(\widehat\alpha,\phi_j)|
    =
    \lambda_n |\widehat\beta_j|^{q-1},
    \qquad j=1,\ldots,p.
\end{align*}
\end{proposition}

Proposition~\ref{prop:ss-arb} shows that balancing is not an additional constraint imposed after representer estimation. It is the KKT condition of the generalized Riesz regression problem. The functions to be balanced are functions of the full regressor \(X\). In treatment-effect applications with \(X=(D,Z)\), this means that the dictionary should generally contain treatment-specific functions, such as \(D\phi(Z)\) and \((1-D)\phi(Z)\), rather than only common functions of \(Z\).

This balancing property directly controls the deterministic component of the estimated Neyman score. Let \(\varepsilon_i=Y_i-\gamma_0(X_i)\) and define the empirical semi-supervised Neyman error by
\begin{align}
    \mathrm{NE}_{n,m}^{\mathrm{TS}}(\widehat\gamma,\widehat\alpha)
    &\coloneqq
    \frac{1}{n}\sum^n_{i=1}\widehat\alpha(X_i)\p{Y_i-\widehat\gamma(X_i)}
    +
    \widehat{\calL}_\kappa(\widehat\gamma)
    -
    \widehat{\calL}_\kappa(\gamma_0).
    \label{eq:ss-neyman-error-def}
\end{align}
By linearity of \(\widehat{\calL}_\kappa\), we have
\begin{align}
    \mathrm{NE}_{n,m}^{\mathrm{TS}}(\widehat\gamma,\widehat\alpha)
    =
    \frac{1}{n}\sum^n_{i=1}\widehat\alpha(X_i)\varepsilon_i
    -
    \widehat{\Delta}_\kappa\p{\widehat\alpha,\widehat\gamma-\gamma_0}.
    \label{eq:ss-neyman-error-decomposition}
\end{align}
Under cross-fitting, the first term is conditionally mean zero given the training sample and the held-out regressors. The second term is the deterministic deviation caused by imperfect regressor balance. If
\begin{align*}
    \widehat\gamma(x)-\gamma_0(x)=a^\top\phi(x)+r(x),
\end{align*}
then Proposition~\ref{prop:ss-arb} yields the bound
\begin{align}
    \abs{\widehat{\Delta}_\kappa\p{\widehat\alpha,\widehat\gamma-\gamma_0}}
    \le
    \lambda_n\|a\|_1
    +
    \abs{\widehat{\Delta}_\kappa(\widehat\alpha,r)}
    \label{eq:ss-neyman-error-balance-bound}
\end{align}
for the \(\ell_1\)-penalized case. Thus, the represented part of the score-relevant regression error is automatically balanced, and only regularization slack and approximation error remain.

\subsection{Choice of the Convex Function}
\label{subsec:ss-convex}
Different choices of the convex function yield different divergences. The following loss functions are examples.

\paragraph{Squared-loss-type Riesz regression.}
If we set the convex function to
\begin{align*}
g^{\mathrm{SQ}}(a)=\frac{1}{2}(a-C)^2,
\end{align*}
then we obtain the squared-loss-type objective in Riesz regression, which is identical to the original Riesz regression \citep{Chernozhukov2021automaticdebiased,Chernozhukov2022automaticdebiased} and least-squares importance fitting (LSIF) in density-ratio estimation \citep{Kanamori2009aleastsquares}. Note that in this case, the derivative of the convex function is given as $\partial g^{\mathrm{SQ}}(a)=a-C$.
With the affine link $\alpha_\beta(x)=C+\phi(x)^\top\beta$, the dual coordinate is linear, and the objective becomes the semi-supervised analogue of least-squares Riesz regression.

\paragraph{UKL (unnormalized KL)-type Riesz regression.}
When $\alpha_{0,\kappa}$ is nonnegative, or when a signed representer is decomposed into known-sign branches, one can use the positive-branch unnormalized KL (UKL) objective
\begin{align*}
g^{\mathrm{UKL}}(a)=a\log a-a,
\qquad
\partial g^{\mathrm{UKL}}(a)=\log a,
\qquad a>0.
\end{align*}
The link $\alpha_\beta(x)=\exp\p{\phi(x)^\top\beta}$ is compatible with automatic regressor balancing. In addition, this choice yields an entropy-balancing or KLIEP-type objective as its dual.

\paragraph{Other loss functions.}
We can also use other convex functions in the Bregman divergence with the automatic regressor balancing property. Binary KL-type (BKL) losses connect to logistic likelihoods, Basu-power losses interpolate between squared-loss and KL-type behavior, and PU (positive and unlabeled)-type losses are useful when the representer is bounded in an interval.

\subsection{Candidate Hypothesis Classes}
\label{subsec:ss-hypothesis-class}
Several hypothesis classes can be used for $\alpha_{0,\kappa}$.

\paragraph{Series and sieve classes.}
A basic choice is the finite-dimensional dual-linear class
\begin{align*}
    \calF_n^{\mathrm{ser}}
    =
    \{x\mapsto \phi_n(x)^\top\beta\colon\beta\in\bbR^{p_n}\},
\end{align*}
where \(\phi_n\) contains splines, polynomials, interactions, or other researcher-chosen features. In causal applications with \(X=(D,Z)\), treatment-specific dictionaries are recommended because the score-relevant regression error is generally a function of the full regressor.

\paragraph{RKHS and random-feature classes.}
Kernel methods can be used by taking \(\calF_n\) to be an RKHS ball or a finite-dimensional approximation obtained by random Fourier features or Nystr{\"o}m features. This gives functional balance over a rich class without manually specifying every interaction.

\paragraph{Tree, forest, matching, and neural classes.}
Tree leaf indicators and random-forest leaf encodings produce sparse dictionaries that adapt to heterogeneous regions of the regressor space. Nearest-neighbor catchment bases give matching-type representer estimators under squared loss \citep{Kato2025nearestneighbor}. Neural networks can be used either as frozen embeddings followed by a convex dual-linear Riesz fit or as a fully end-to-end nonconvex class. 

\begin{remark}[Non-negative correction for flexible positive-branch models]
As pointed out in the density-ratio literature \citep{Kato2021nonnegativebregman,Rhodes2020telescopingdensityratio}, if we use flexible models for $\alpha_{0,\kappa}$, the optimization of the Bregman objective can suffer from overfitting. If \(\alpha_{0,\kappa}\) is positive and bounded by \(R\), a non-negative correction can be applied. We can also apply the telescoping technique proposed by \citet{Rhodes2020telescopingdensityratio}, or its extension based on the infinitesimal classification approach proposed by \citet{Choi2022densityratio}. For example, \citet{Kato2026scorematchingriesz} applies the infinitesimal classification approach to Riesz representer estimation without an unlabeled dataset.
\end{remark}

\subsection{Error Analysis}
\label{subsec:ss-error-analysis}
We now provide error bounds for the semi-supervised generalized Riesz regression estimator. The analysis follows the same structure as error analyses for density-ratio estimation and Riesz regression. First, the population objective is identified with a Bregman divergence. Second, the empirical process is decomposed into labeled and unlabeled parts. Third, strong convexity converts excess Bregman risk into \(L^2(P_{0X})\) error. The first result is a general oracle inequality. Later results specialize this inequality to finite pseudo-dimension classes and deep ReLU sieves.

For \(\alpha\in\calH_n\), define $c_\alpha(x)\coloneqq m(x,u_\alpha)$ and $a_\alpha(x)\coloneqq b_\alpha(x)-\kappa c_\alpha(x)$. Also define $\calA_n\coloneqq \{a_\alpha\colon\alpha\in\calH_n\}$ and $\calC_n\coloneqq \{c_\alpha\colon\alpha\in\calH_n\}$. 
For a probability law \(R\) and a function class \(\calF\), let
\begin{align*}
    \mathfrak R_{r,R}(\calF)
    \coloneqq
    \bbE\sqb{
        \sup_{f\in\calF}|\frac{1}{r}\sum_{i=1}^r \xi_i f(S_i)|
    }
\end{align*}
be the Rademacher complexity, where \(S_i\sim R\) and \(\xi_i\) are independent Rademacher variables. Define
\begin{align}
    \mathfrak E_{n,m}(\delta)
    &\coloneqq
    2\mathfrak R_{n,P_{0X}}(\calA_n)
    +2(1-\kappa)\mathfrak R_{m,Q_{0X}}(\calC_n)
    \nonumber\\
    &\quad
    +B_A\sqrt{\frac{\log(4/\delta)}{2n}}
    +(1-\kappa)B_C\sqrt{\frac{\log(4/\delta)}{2m}},
    \label{eq:ss-emp-process-bound}
\end{align}
where \(B_A\) and \(B_C\) are envelopes for \(\calA_n\) and \(\calC_n\).

\begin{assumption}[Conditions for Bregman--Riesz error analysis]
\label{ass:ss-grr-error}
The following conditions hold.
\begin{enumerate}
    \item The generator \(g\) is twice continuously differentiable on a compact interval \(\calA\), and there exist constants \(0<\underline g\le \overline g<\infty\) such that
    \begin{align*}
        \underline g\le \ddot g(a)\le \overline g
        \qquad \text{for all } a\in\calA.
    \end{align*}
    \item \(\alpha_{0,\kappa}(X)\in\calA\) almost surely and every \(\alpha\in\calH_n\) takes values in \(\calA\).
    \item The classes \(\calA_n\) and \(\calC_n\) have finite envelopes \(B_A\) and \(B_C\).
    \item The empirical minimizer \(\widehat\alpha\) in \eqref{eq:ss-grr-estimator} exists. The regularizer satisfies \(Reg_\alpha(\alpha)\ge0\).
\end{enumerate}
\end{assumption}

\begin{theorem}[Oracle inequality for semi-supervised generalized Riesz regression]
\label{thm:ss-grr-oracle}
Suppose that Assumption~\ref{ass:ss-grr-error} holds. Then, with probability at least \(1-\delta\), it holds that
\begin{align}
    \frac{\underline g}{2}
    \|\widehat\alpha-\alpha_{0,\kappa}\|_{L^2(P_{0X})}^2
    &\le
    \inf_{\alpha\in\calH_n}
    \cb{
        \bbE_{P_{0X}}\sqb{
            \text{BD}_g^{\dagger}\p{\alpha_{0,\kappa}(X)\mid\alpha(X)}
        }
        +\lambda_n Reg_\alpha(\alpha)
    }
    +2\mathfrak E_{n,m}(\delta).
    \label{eq:ss-grr-oracle-bound}
\end{align}
In particular, if \(\alpha_{0,\kappa}\in\calH_n\), then
\begin{align}
    \|\widehat\alpha-\alpha_{0,\kappa}\|_{L^2(P_{0X})}^2
    &\le
    \frac{2}{\underline g}
    \cb{
        \lambda_n Reg_\alpha(\alpha_{0,\kappa})+2\mathfrak E_{n,m}(\delta)
    }.
    \label{eq:ss-grr-well-specified-bound}
\end{align}
\end{theorem}

Theorem~\ref{thm:ss-grr-oracle} is a general distribution-free oracle inequality. It is useful for stability diagnostics and for abstract DML conditions. Under a localized empirical-process condition, the usual fast-rate version is obtained. Let \(V_n\) denote a dimension or pseudo-dimension upper bound for the dual class \(\calF_n=\{u_\alpha\colon\alpha\in\calH_n\}\), and let
\begin{align*}
    r_{n,m}^2(\delta)
    \coloneqq
    \cb{V_n\log(n+m)+\log(1/\delta)}
    \p{\frac{1}{n}+\frac{1}{m}}.
\end{align*}

\begin{corollary}[Fast rate under finite pseudo-dimension]
\label{cor:ss-grr-fast-rate}
In addition to Assumption~\ref{ass:ss-grr-error}, suppose that the localized Bernstein condition for the Bregman loss class holds and that the classes generated by \(\calF_n\) have pseudo-dimension bounded by \(V_n\). Then, there exists a constant \(C\) depending only on the envelope and strong-convexity constants such that, with probability at least \(1-\delta\), it holds that
\begin{align}
    \|\widehat\alpha-\alpha_{0,\kappa}\|_{L^2(P_{0X})}^2
    &\le
    C\inf_{\alpha\in\calH_n}
    \|\alpha-\alpha_{0,\kappa}\|_{L^2(P_{0X})}^2
    +C\lambda_n Reg_\alpha(\alpha_n^*)
    +C r_{n,m}^2(\delta),
    \label{eq:ss-grr-fast-rate}
\end{align}
where \(\alpha_n^*\) is any near-best element in \(\calH_n\). Consequently,
\begin{align*}
    \|\widehat\alpha-\alpha_{0,\kappa}\|_{L^2(P_{0X})}
    =
    O_p\p{
        \inf_{\alpha\in\calH_n}\|\alpha-\alpha_{0,\kappa}\|_{L^2(P_{0X})}
        +
        \sqrt{V_n\log(n+m)}\p{\frac{1}{n}+\frac{1}{m}}^{1/2}
        +
        \sqrt{\lambda_n Reg_\alpha(\alpha_n^*)}
    }.
\end{align*}
\end{corollary}

We next provide primitive rates for deep ReLU sieves. These results make the abstract rate condition in Assumption~\ref{asm:conv_rate} verifiable from smoothness assumptions on the dual Riesz oracle. They also show precisely where density-ratio analysis enters the semi-supervised generalized Riesz regression problem.

Let
\begin{align*}
    N_{\min}\coloneqq n\wedge m,
    \qquad
    f_0\coloneqq \partial g\circ\alpha_{0,\kappa}.
\end{align*}
Let \(\calF_n\) be a class of clipped ReLU feedforward neural networks, and define \(\widehat\alpha=(\partial g)^{-1}\circ\widehat f\), where \(\widehat f\) minimizes the dual objective in \eqref{eq:ss-dual-estimator}. For \(s>0\), let \(\calH^s([0,1]^d,M_f)\) denote the H\"{o}lder ball with smoothness \(s\) and radius \(M_f\).

\begin{assumption}[Deep ReLU rate conditions]
\label{ass:ss-relu-rate}
The following conditions hold.
\begin{enumerate}
    \item Assumption~\ref{ass:ss-grr-error} holds, and the dual range of every \(f\in\calF_n\) is contained in a compact interval on which \((\partial g)^{-1}\) is Lipschitz.
    \item The map \(f\mapsto m(\cdot,f)\) preserves the empirical-process complexity used in Corollary~\ref{cor:ss-grr-fast-rate}. In particular, the classes generated by \(g^*(f)\), \(m(\cdot,f)\), and their centered Bregman-loss differences have pseudo-dimension bounded by a constant multiple of the pseudo-dimension of \(\calF_n\).
    \item The localized Bernstein condition in Corollary~\ref{cor:ss-grr-fast-rate} holds for the clipped ReLU loss classes.
    \item The regularization term satisfies \(\lambda_nReg_\alpha(\alpha_n^*)=o(1)\), where \(\alpha_n^*\) is a near-best element induced by the ReLU approximation to \(f_0\).
\end{enumerate}
\end{assumption}

\begin{theorem}[Deep ReLU rate under H\"{o}lder smoothness]
\label{thm:ss-relu-rate}
Suppose that Assumption~\ref{ass:ss-relu-rate} holds. Suppose also that \(X\in[0,1]^d\), \(\tildeX\in[0,1]^d\), and \(f_0\in\calH^s([0,1]^d,M_f)\). Choose the clipped ReLU architecture so that the H\"{o}lder approximation error and the pseudo-dimension term are balanced. Then, for a finite constant \(c\), it holds that
\begin{align}
    \|\widehat\alpha-\alpha_{0,\kappa}\|_{L^2(P_{0X})}^2
    =
    O_p\p{
        N_{\min}^{-2s/(d+2s)}\log^c N_{\min}
        +\lambda_nReg_\alpha(\alpha_n^*)
    }.
    \label{eq:ss-relu-rate}
\end{align}
If \(n=\rho N\) and \(m=(1-\rho)N\) for fixed \(\rho\in(0,1)\), then we have
\begin{align}
    \|\widehat\alpha-\alpha_{0,\kappa}\|_{L^2(P_{0X})}
    =
    O_p\p{
        N^{-s/(d+2s)}\log^{c/2}N
        +\sqrt{\lambda_nReg_\alpha(\alpha_n^*)}
    }.
    \label{eq:ss-relu-rate-root}
\end{align}
\end{theorem}

The rate in Theorem~\ref{thm:ss-relu-rate} is the semi-supervised analogue of density-ratio estimators based on neural networks. The only structural difference is that the objective contains two empirical processes: the labeled one for the \(P_{0X}\) inner product and the unlabeled one for the \(Q_{0X}\) part of the target functional. Since \(N_{\min}\) controls both empirical processes, the rate is governed by the smaller sample size.

\begin{proposition}[Density-ratio special case]
\label{prop:ss-dre-rate}
Suppose that \(m(X,\gamma)=\gamma(X)\) and \(\kappa=0\). In this case, \(\alpha_{0,0}(x)=q_{0X}(x)/p_{0X}(x)\). Let \(D_0(x)=\log\p{q_{0X}(x)/p_{0X}(x)}\), and suppose that \(D_0\in\calH^s([0,1]^d,M_f)\) and \(D_0\) is bounded. Let \(\widehat D\) be the Bregman deep ReLU density-ratio estimator and set \(\widehat\alpha(x)=\exp\p{\widehat D(x)}\). Then, it holds that
\begin{align}
    \|\widehat\alpha-\alpha_{0,0}\|_{L^2(P_{0X})}^2
    =
    O_p\p{
        N_{\min}^{-2s/(d+2s)}\log^3 N_{\min}
    }.
    \label{eq:ss-dre-rate}
\end{align}
More generally, for a mean-type target with \(\kappa\in(0,1)\), the estimator
\begin{align*}
    \widehat\alpha_{\kappa}(x)=\kappa+(1-\kappa)\exp\p{\widehat D(x)}
\end{align*}
satisfies the same rate up to the multiplicative factor \((1-\kappa)^2\) in the squared error.
\end{proposition}

Proposition~\ref{prop:ss-dre-rate} gives the exact point at which the semi-supervised generalized Riesz regression objective reduces to density-ratio estimation under the Bregman divergence. It justifies the use of theoretical results developed in density-ratio estimation.

\begin{theorem}[Unbounded-support extension]
\label{thm:ss-unbounded-support-rate}
Suppose that Assumption~\ref{ass:ss-relu-rate} holds on \(\bbR^d\), and suppose that \(f_0\in\calH^s(\bbR^d,M_f)\). Assume that there exist constants \(C_1,C_2>0\) such that
\begin{align}
    \max\cb{
        P_{0X}(\|X\|_\infty\ge C_1\log N_{\min}),
        Q_{0X}(\|\tildeX\|_\infty\ge C_1\log N_{\min})
    }
    \le
    C_2N_{\min}^{-2s/(d+2s)}.
    \label{eq:ss-tail-condition}
\end{align}
Choose the clipped ReLU architecture as in Theorem~\ref{thm:ss-relu-rate} after truncating and rescaling the covariate support. Then, for a finite constant \(c'\), it holds that
\begin{align}
    \|\widehat\alpha-\alpha_{0,\kappa}\|_{L^2(P_{0X})}^2
    =
    O_p\p{
        N_{\min}^{-2s/(d+2s)}\log^{c'} N_{\min}
        +\lambda_nReg_\alpha(\alpha_n^*)
    }.
    \label{eq:ss-unbounded-support-rate}
\end{align}
Compared with Theorem~\ref{thm:ss-relu-rate}, the logarithmic exponent may increase because the approximation is performed on a cube whose side length grows at the order \(\log N_{\min}\).
\end{theorem}

\begin{theorem}[Approximate low-dimensional manifold extension]
\label{thm:ss-manifold-rate}
Suppose that Assumption~\ref{ass:ss-relu-rate} holds. Suppose also that \(P_{0X}\) and \(Q_{0X}\) are concentrated on a \(\varrho\)-neighborhood of a compact \(d_M\)-dimensional Riemannian submanifold of \([0,1]^d\), whose condition number, volume, and geodesic covering regularity satisfy the conditions used in the approximate-manifold analysis of deep ReLU density-ratio estimation. Let \(V\), \(R\), and \(\tau\) denote the corresponding volume, geodesic covering regularity, and condition-number parameters. Let
\begin{align*}
    d_\delta=O\p{d_M\log(dVR\tau^{-1}/\delta)/\delta^2}
\end{align*}
be the Euclidean embedding dimension associated with relative distortion \(\delta\in(0,1)\). If \(f_0\in\calH^s([0,1]^d,M_f)\) and the neighborhood radius \(\varrho\) is no larger than the ReLU approximation scale, then, for a finite constant \(c''\), it holds that
\begin{align}
    \|\widehat\alpha-\alpha_{0,\kappa}\|_{L^2(P_{0X})}^2
    =
    O_p\p{
        N_{\min}^{-2s/(d_\delta+2s)}\log^{c''}N_{\min}
        +\lambda_nReg_\alpha(\alpha_n^*)
    }.
    \label{eq:ss-manifold-rate}
\end{align}
Thus, the exponent depends on the embedded dimension \(d_\delta\) rather than the ambient dimension \(d\).
\end{theorem}

Theorems~\ref{thm:ss-relu-rate}, \ref{thm:ss-unbounded-support-rate}, and \ref{thm:ss-manifold-rate} should be interpreted together with the minimax message of \citet{Wasserman2007statisticalanalysis}. Unlabeled regressors do not automatically improve a nonparametric rate merely because they are available. In the rate analysis above, they have two precise roles. First, they change the target functional and the efficiency bound through \(V_{0X}\). Second, they enter the Riesz objective through \(\widehat{\calL}_\kappa\). The rates above require smoothness of the dual Riesz oracle, tail control, or low-dimensional structure. Without such structure, Assumption~\ref{asm:conv_rate} is the appropriate abstract condition.

\begin{corollary}[Primitive sufficient condition for the DML-PPCI product rate]
\label{cor:ss-product-rate}
Suppose that \(n=\rho N\) and \(m=(1-\rho)N\) for fixed \(\rho\in(0,1)\). Suppose that $\widehat\alpha$ satisfies
\begin{align*}
    \|\widehat\alpha-\alpha_{0,\kappa}\|_{P,2}
    =
    O_p\p{N^{-s_\alpha/(d_\alpha+2s_\alpha)}\log^{c_\alpha}N},
\end{align*}
and $\widehat\gamma$ satisfies
\begin{align*}
    \|\widehat\gamma-\gamma_0\|_{P,2}
    =
    O_p\p{N^{-s_\gamma/(d_\gamma+2s_\gamma)}\log^{c_\gamma}N}.
\end{align*}
If we have
\begin{align}
    \frac{s_\alpha}{d_\alpha+2s_\alpha}
    +
    \frac{s_\gamma}{d_\gamma+2s_\gamma}
    >
    \frac{1}{2},
    \label{eq:ss-product-smoothness-condition}
\end{align}
then the product-rate condition in Assumption~\ref{asm:conv_rate} holds as follows:
\begin{align*}
    \|\widehat\alpha-\alpha_{0,\kappa}\|_{P,2}
    \|\widehat\gamma-\gamma_0\|_{P,2}
    =
    o_p(N^{-1/2}).
\end{align*}
For the bounded-support ReLU rate, \(d_\alpha=d\). For the approximate-manifold rate, \(d_\alpha=d_\delta\).
\end{corollary}

\subsection{Automatic Neyman Orthogonalization}
\label{subsec:ss-ano}
The balancing property also clarifies the relation between the Riesz-weighted estimator and the ARW estimator. Suppose that exact balance holds for the true regression function,
\begin{align*}
    \widehat{\Delta}_\kappa(\widehat\alpha,\gamma_0)=0.
\end{align*}
Then, we have
\begin{align}
    \frac{1}{n}\sum^n_{i=1}\widehat\alpha(X_i)Y_i
    &=
    \frac{1}{n}\sum^n_{i=1}\widehat\alpha(X_i)\gamma_0(X_i)
    +
    \frac{1}{n}\sum^n_{i=1}\widehat\alpha(X_i)\p{Y_i-\gamma_0(X_i)}
    \nonumber\\
    &=
    \widehat{\calL}_\kappa(\gamma_0)
    +
    \frac{1}{n}\sum^n_{i=1}\widehat\alpha(X_i)\p{Y_i-\gamma_0(X_i)}.
    \label{eq:ss-rw-as-infeasible-arw}
\end{align}
Thus, the Riesz-weighted estimator behaves as the infeasible semi-supervised ARW estimator that uses the true regression function.

\section{Efficiency Bound and Efficient Influence Function in the One-Sample Scenario}
\label{sec:one-sample}
Next, we consider the one-sample scenario, where we observe one sample and labels are observed with certain probabilities.

\subsection{Setup}
In this scenario, a complete population observation is \((X,Y^*)\), but the outcome is observed only when \(S=1\). The observed data are
\begin{align*}
    O=(X,S,Y),
    \qquad
    Y=SY^*+(1-S)\mathrm{NA}.
\end{align*}
Let \(P_{0X}\) be the marginal distribution of \(X\), let \(\pi_0(X)\coloneqq\Prb(S=1\mid X)\), and let \(\gamma_0(X)\coloneqq\bbE\sqb{Y^*\mid X}\). We assume the missing-at-random condition
\begin{align*}
    Y^*\perp S\mid X
\end{align*}
and the overlap condition that \(\pi_0(X)\) is bounded away from zero. In the one-sample scenario, the full set of regressors is observed from \(P_{0X}\). Therefore, the target parameter is
\begin{align}
    \theta^{\mathrm{OS}}_0
    \coloneqq
    \bbE_{P_{0X}}\sqb{m(X,\gamma_0)}.
    \label{eq:os-target}
\end{align}

\subsection{Efficient Influence Function}
Let \(\alpha^{\mathrm{OS}}_0\) be the Riesz representer of the one-sample functional, so that
\begin{align}
    \bbE_{P_{0X}}\sqb{m(X,\gamma)}
    =
    \bbE_{P_{0X}}\sqb{\alpha^{\mathrm{OS}}_0(X)\gamma(X)}
    \qquad
    \text{for all }\gamma\in\Gamma.
    \label{eq:os-riesz}
\end{align}
We derive the efficient influence function. First, we impose the following assumption.
\begin{assumption}[Regularity for the one-sample efficiency bound]
\label{ass:os-eff}
The following conditions hold.
\begin{enumerate}
    \item The observed data \(O_i=(X_i,S_i,Y_i)\), \(i=1,\ldots,N\), are independent copies of \(O=(X,S,Y)\), where \(Y\) is observed only when \(S=1\). Whenever an expression contains \(S\p{Y-\gamma(X)}\), it denotes the observable product \(S\p{Y^*-\gamma(X)}\).
    \item The missing-at-random condition \(Y^*\perp S\mid X\) holds.
    \item There exists a constant \(c_\pi>0\) such that \(\pi_0(X)\ge c_\pi\) almost surely.
    \item The regression satisfies \(\gamma_0\in\Gamma\), \(\bbE\sqb{(Y^*-\gamma_0(X))^2}<\infty\), the map \(\gamma\mapsto m(\cdot,\gamma)\) is linear, and \(m(\cdot,\gamma_0)\in L^2(P_{0X})\).
    \item The Riesz representer \(\alpha^{\mathrm{OS}}_0\) in \eqref{eq:os-riesz} exists and satisfies \(\bbE\sqb{\p{\alpha^{\mathrm{OS}}_0(X)}^2\sigma_0^2(X)/\pi_0(X)}<\infty\).
\end{enumerate}
\end{assumption}
Then, we obtain the following theorem.
\begin{theorem}[Efficient influence function in the one-sample scenario]
\label{thm:os-eif}
Suppose that Assumption~\ref{ass:os-eff} holds. Under the nonparametric observed-data model for \(O=(X,S,Y)\), the efficient influence function for \(\theta^{\mathrm{OS}}_0\) is
\begin{align}
    \psi^{\mathrm{OS}}_0(O)
    \coloneqq
    \frac{S}{\pi_0(X)}\alpha^{\mathrm{OS}}_0(X)\p{Y-\gamma_0(X)}
    +
    m(X,\gamma_0)-\theta^{\mathrm{OS}}_0.
    \label{eq:os-eif}
\end{align}
Here, the residual term is evaluated only when \(S=1\). Equivalently, \(S\p{Y-\gamma_0(X)}=S\p{Y^*-\gamma_0(X)}\).
\end{theorem}

\subsection{Efficiency Bound}
The efficiency bound is given as follows.
\begin{theorem}[Efficiency bound in the one-sample scenario]
\label{thm:os-eff-bound}
Suppose that Assumption~\ref{ass:os-eff} holds. Then, under \(\sqrt N\) normalization, the asymptotic variance of any regular estimator of \(\theta^{\mathrm{OS}}_0\) is lower bounded by
\begin{align}
    V^{\mathrm{OS}}_0
    &\coloneqq
    \bbE\sqb{\psi^{\mathrm{OS}}_0(O)^2}
    =
    \bbE_{P_{0X}}\sqb{
        \frac{\p{\alpha^{\mathrm{OS}}_0(X)}^2\sigma_0^2(X)}{\pi_0(X)}
    }
    +
    \operatorname{Var}_{P_{0X}}\p{m(X,\gamma_0)},
    \label{eq:os-eff-bound}
\end{align}
where \(\sigma_0^2(X)=\Var(Y^*\mid X)\).
\end{theorem}

The first term in \eqref{eq:os-eff-bound} depends on \(1/\pi_0(X)\) because outcomes are observed only for labeled units. The second term does not depend on \(1/\pi_0(X)\) because \(X\) is observed for every unit.

\subsection{Efficient Estimators}
\label{subsec:os-efficient-estimators}
Based on the efficient influence function in Theorem~\ref{thm:os-eif}, we construct one-sample analogues of EE-DML-PPCI and TMLE-DML-PPCI. Let \(\widehat\gamma\), \(\widehat\alpha\), and \(\widehat\pi\) be estimators of \(\gamma_0\), \(\alpha^{\mathrm{OS}}_0\), and \(\pi_0\), respectively. As in the two-sample scenario, if the Donsker condition does not hold for the estimators, we employ cross-fitting to control the empirical process term in the asymptotic analysis.

\paragraph{EE-DML-PPCI.}
The one-sample EE-DML-PPCI estimator is
\begin{align}
    \widehat\theta^{\mathrm{OS}}_{\mathrm{EE}}
    \coloneqq
    \frac{1}{N}\sum^N_{i=1}
    \p{
        \frac{S_i}{\widehat\pi(X_i)}
        \widehat\alpha(X_i)
        \p{Y_i-\widehat\gamma(X_i)}
        +
        m\p{X_i,\widehat\gamma}
    }.
    \label{eq:os-ee-estimator}
\end{align}
Here, the product \(S_i\p{Y_i-\widehat\gamma(X_i)}\) is interpreted as \(S_i\p{Y_i^*-\widehat\gamma(X_i)}\), which is observed.

\paragraph{TMLE-DML-PPCI.}
The one-sample TMLE-DML-PPCI estimator first updates \(\widehat\gamma\) in the direction of \(\widehat\alpha\). For example, if we consider TMLE under the Gaussian likelihood, we update the initial estimator as
\begin{align}
    \widehat\gamma^{(1)}(x)
    \coloneqq
    \widehat\gamma(x)
    +
    \widehat\varepsilon^{\mathrm{OS}}\widehat\alpha(x),
    \label{eq:os-tmle-update}
\end{align}
where $\widehat\varepsilon^{\mathrm{OS}}$ is the perturbation defined as
\begin{align}
    \widehat\varepsilon^{\mathrm{OS}}
    &\coloneqq
    \frac{\widehat A^{\mathrm{OS}}_N}{\widehat D^{\mathrm{OS}}_N},
    \label{eq:os-tmle-epsilon}
    \\
    \widehat A^{\mathrm{OS}}_N
    &\coloneqq
    \frac{1}{N}\sum^N_{i=1}
    \frac{S_i}{\widehat\pi(X_i)}
    \widehat\alpha(X_i)
    \p{Y_i-\widehat\gamma(X_i)},
    \qquad
    \widehat D^{\mathrm{OS}}_N
    \coloneqq
    \frac{1}{N}\sum^N_{i=1}
    \widehat\alpha^2\p{X_i}.
\end{align}
Then, the one-sample TMLE-DML-PPCI estimator is given as
\begin{align}
    \widehat\theta^{\mathrm{OS}}_{\mathrm{TMLE}}
    \coloneqq
    \frac{1}{N}\sum^N_{i=1}
    m\p{X_i,\widehat\gamma^{(1)}}.
    \label{eq:os-tmle-estimator}
\end{align}
The numerator in \eqref{eq:os-tmle-epsilon} uses labeled residuals corrected by \(S_i/\widehat\pi(X_i)\), because outcomes are observed only when \(S_i=1\). The denominator estimates the target-law average of \(\widehat\alpha^2\) in the one-sample scenario, where the regressor distribution is observed through \(\cb{X_i}^N_{i=1}\).

\subsection{Asymptotic Properties}
\label{subsec:os-asymptotic-properties}
We first establish consistency and then show asymptotic linearity and efficiency. The assumptions are stated for generic nuisance estimators \(\widehat\gamma\), \(\widehat\alpha\), and \(\widehat\pi\), in the same way as in the two-sample scenario.

\begin{assumption}[Convergence condition for the regression function estimator]
\label{ass:os-reg-consistency}
It holds that
\begin{align}
    \|\widehat\gamma-\gamma_0\|_{P,2}
    =
    o_p(1)
    \qquad
    (N\to\infty).
    \label{eq:os-reg-consistency}
\end{align}
Moreover,
\begin{align}
    \|m(\cdot,\widehat\gamma)-m(\cdot,\gamma_0)\|_{P,2}
    =
    o_p(1)
    \qquad
    (N\to\infty).
    \label{eq:os-m-consistency}
\end{align}
\end{assumption}

\begin{assumption}[Convergence condition for the Riesz representer and labeling probability estimators]
\label{ass:os-riesz-pi-consistency}
It holds that
\begin{align}
    \|\widehat\alpha-\alpha^{\mathrm{OS}}_0\|_{P,2}
    =
    o_p(1),
    \qquad
    \|\widehat\pi-\pi_0\|_{P,2}
    =
    o_p(1)
    \qquad
    (N\to\infty).
    \label{eq:os-riesz-pi-consistency}
\end{align}
Moreover, \(\widehat\pi(X)\) and \(\pi_0(X)\) are bounded away from zero with probability approaching one, and \(\|\widehat\alpha\|_{P,2}=O_p(1)\).
\end{assumption}

The following theorem holds.

\begin{theorem}[Consistency of one-sample DML-PPCI]
\label{thm:os-consistency}
Suppose that Assumption~\ref{ass:os-eff} holds. If either Assumption~\ref{ass:os-reg-consistency} holds or Assumption~\ref{ass:os-riesz-pi-consistency} holds, then
\begin{align}
    \widehat\theta^{\mathrm{OS}}_{\mathrm{EE}}
    \xrightarrow{\rmp}
    \theta^{\mathrm{OS}}_0
    \qquad
    (N\to\infty).
    \label{eq:os-ee-consistency}
\end{align}
\end{theorem}

To show asymptotic normality, both consistency conditions need to hold. In addition, we impose the following rate and empirical-process conditions.

\begin{assumption}[Convergence rate condition]
\label{ass:os-product-rate}
It holds that
\begin{align}
    \p{
        \|\widehat\alpha-\alpha^{\mathrm{OS}}_0\|_{P,2}
        +
        \|\widehat\pi-\pi_0\|_{P,2}
    }
    \|\widehat\gamma-\gamma_0\|_{P,2}
    =
    o_p(N^{-1/2})
    \qquad
    (N\to\infty).
    \label{eq:os-product-rate}
\end{align}
\end{assumption}

\begin{assumption}[Donsker condition or cross-fitting]
\label{ass:os-donsker-or-crossfit}
Define
\begin{align*}
    \widehat\varphi^{\mathrm{OS}}(O)
    &\coloneqq
    \frac{S}{\widehat\pi(X)}
    \widehat\alpha(X)
    \p{Y-\widehat\gamma(X)}
    +
    m\p{X,\widehat\gamma},
    \\
    \varphi^{\mathrm{OS}}_0(O)
    &\coloneqq
    \frac{S}{\pi_0(X)}
    \alpha^{\mathrm{OS}}_0(X)
    \p{Y-\gamma_0(X)}
    +
    m\p{X,\gamma_0}.
\end{align*}
The nuisance estimators \(\widehat\gamma\), \(\widehat\alpha\), and \(\widehat\pi\) either satisfy the required Donsker-type stochastic equicontinuity conditions or are constructed via cross-fitting so that
\begin{align}
    &
    \frac{1}{N}\sum^N_{i=1}
    \p{
        \widehat\varphi^{\mathrm{OS}}(O_i)
        -
        \varphi^{\mathrm{OS}}_0(O_i)
    }
    -
    \bbE_{P_0}\sqb{
        \widehat\varphi^{\mathrm{OS}}(O)
        -
        \varphi^{\mathrm{OS}}_0(O)
    }
    =
    o_p(N^{-1/2}).
    \label{eq:os-empirical-process-condition}
\end{align}
Moreover, the corresponding score components have uniformly bounded \((2+\delta)\)-moments for some \(\delta>0\).
\end{assumption}

\begin{assumption}[Targeting calibration condition]
\label{ass:os-targeting-calibration}
The denominator \(\widehat D^{\mathrm{OS}}_N\) in the perturbation is bounded away from zero, and
\begin{align}
    \widehat L^{\mathrm{OS}}_N
    -
    \widehat D^{\mathrm{OS}}_N
    =
    o_p(1)
    \qquad
    (N\to\infty),
    \label{eq:os-targeting-calibration}
\end{align}
where
\begin{align}
    \widehat L^{\mathrm{OS}}_N
    \coloneqq
    \frac{1}{N}\sum^N_{i=1}
    m\p{X_i,\widehat\alpha}.
    \label{eq:os-tmle-L}
\end{align}
Moreover, \(\widehat A^{\mathrm{OS}}_N=O_p(N^{-1/2})\).
\end{assumption}

The following theorem holds.

\begin{theorem}[Asymptotic linearity and efficiency of one-sample DML-PPCI]
\label{thm:os-asymptotic}
Suppose that Assumptions~\ref{ass:os-eff}, \ref{ass:os-reg-consistency}, \ref{ass:os-riesz-pi-consistency}, \ref{ass:os-product-rate}, and \ref{ass:os-donsker-or-crossfit} hold. Then, we have
\begin{align}
    \widehat\theta^{\mathrm{OS}}_{\mathrm{EE}}
    -
    \theta^{\mathrm{OS}}_0
    =
    \frac{1}{N}\sum^N_{i=1}
    \psi^{\mathrm{OS}}_0(O_i)
    +
    o_p(N^{-1/2})
    \qquad
    (N\to\infty).
    \label{eq:os-asymptotic-linear}
\end{align}
Consequently, we have
\begin{align}
    \sqrt N
    \p{
        \widehat\theta^{\mathrm{OS}}_{\mathrm{EE}}
        -
        \theta^{\mathrm{OS}}_0
    }
    \xrightarrow{d}
    N\p{0,V^{\mathrm{OS}}_0}
    \qquad
    (N\to\infty).
    \label{eq:os-ee-asymptotic-normal}
\end{align}
For TMLE-DML-PPCI, suppose in addition that Assumption~\ref{ass:os-targeting-calibration} holds. Then, we have
\begin{align}
    \widehat\theta^{\mathrm{OS}}_{\mathrm{TMLE}}
    -
    \widehat\theta^{\mathrm{OS}}_{\mathrm{EE}}
    =
    o_p(N^{-1/2})
    \qquad
    (N\to\infty),
    \label{eq:os-tmle-ee-equivalence}
\end{align}
and hence
\begin{align}
    \widehat\theta^{\mathrm{OS}}_{\mathrm{TMLE}}
    -
    \theta^{\mathrm{OS}}_0
    =
    \frac{1}{N}\sum^N_{i=1}
    \psi^{\mathrm{OS}}_0(O_i)
    +
    o_p(N^{-1/2})
    \qquad
    (N\to\infty).
    \label{eq:os-tmle-asymptotic-linear}
\end{align}
Consequently, we have
\begin{align}
    \sqrt N
    \p{
        \widehat\theta^{\mathrm{OS}}_{\mathrm{TMLE}}
        -
        \theta^{\mathrm{OS}}_0
    }
    \xrightarrow{d}
    N\p{0,V^{\mathrm{OS}}_0}
    \qquad
    (N\to\infty).
    \label{eq:os-tmle-asymptotic-normal}
\end{align}
Thus, EE-DML-PPCI is regular and semiparametrically efficient under Assumptions~\ref{ass:os-eff}, \ref{ass:os-reg-consistency}, \ref{ass:os-riesz-pi-consistency}, \ref{ass:os-product-rate}, and \ref{ass:os-donsker-or-crossfit}. TMLE-DML-PPCI is regular and semiparametrically efficient under the same assumptions and the additional targeting calibration condition.
\end{theorem}

\section{Conclusion}
This study develops PPCI for causal and structural regression functionals with labeled observations and unlabeled auxiliary regressors. We derive the efficient influence functions and efficiency bounds under two different DGPs, or sampling schemes, and develop an estimation method called DML-PPCI. In particular, we present EE-DML-PPCI and TMLE-DML-PPCI as special implementations of DML-PPCI. We also develop semi-supervised generalized Riesz regression for estimating the Riesz representer in the Neyman orthogonal score. Using these methods, we show that we can construct estimators whose asymptotic variances are smaller than those of estimators constructed using only the labeled dataset. These results clarify when unlabeled regressors improve efficiency.

\bibliography{arXiv2.bbl}

\bibliographystyle{tmlr}

\onecolumn

\tableofcontents

\appendix

\section{Proofs for Section~\ref{sec:efficiency}}
\label{app:ss-efficiency}

\subsection{Proof of Theorem~\ref{thm:ss-eif}}
We prove the result by verifying the pathwise derivative in every regular parametric submodel and then specifying the canonical gradient in the stratified tangent space.

Let $\cb{P_t\colon t\in(-\eta,\eta)}$ be a regular parametric submodel for the labeled distribution and let $\cb{Q_t\colon t\in(-\eta,\eta)}$ be a regular parametric submodel for the unlabeled regressor distribution. Both submodels pass through the true laws at $t=0$. Define their scores as
\begin{align*}
    s_{\mathrm L}\bigp{W}
    =
    \frac{\partial}{\partial t}\log p_t(W)|_{t=0},
    \qquad
    s_{\mathrm U}\bigp{\tildeX}
    =
    \frac{\partial}{\partial t}\log q_t\bigp{\tildeX}|_{t=0}.
\end{align*}
The scores are normalized to have mean zero:
\begin{align*}
    \bbE_{P_0}\sqb{s_{\mathrm L}(W)}=0,
    \qquad
    \bbE_{Q_{0X}}\sqb{s_{\mathrm U}\bigp{\tildeX}}=0.
\end{align*}
Under the labeled law, we decompose the score into the following two elements:
\begin{align*}
    s_X(X)
    \coloneqq
    \bbE_{P_0}\sqb{s_{\mathrm L}(W)\mid X},
    \qquad
    s_{Y\mid X}(W)
    \coloneqq
    s_{\mathrm L}(W)-s_X(X).
\end{align*}
They also satisfy
\begin{align*}
    \bbE_{P_0}\sqb{s_X(X)}=0,
    \qquad
    \bbE_{P_0}\sqb{s_{Y\mid X}(W)\mid X}=0.
\end{align*}
The first equality follows by taking expectations of \(s_X(X)\), and the second follows directly from the definition.

Let us define
\begin{align*}
    \gamma_t(x)\coloneqq \bbE_{P_t}\sqb{Y\mid X=x}.
\end{align*}
We first compute the derivative of \(\gamma_t\). From the conditional likelihood, for each fixed \(x\), the conditional score for \(Y\) given \(X=x\) is \(s_{Y\mid X}(Y,x)\). Hence, we have
\begin{align*}
    \dot\gamma_0(x)
    &\coloneqq
    \frac{\partial}{\partial t}\gamma_t(x)|_{t=0}
    \nonumber\\
    &=
    \frac{\partial}{\partial t}
    \int y p_t(y\mid x)\rmd y
    |_{t=0}
    \nonumber\\
    &=
    \int y s_{Y\mid X}(y,x)p_0(y\mid x)\rmd y.
\end{align*}
Since \(\bbE_{P_0}\sqb{s_{Y\mid X}(Y,X)\mid X=x}=0\) holds, subtracting \(\gamma_0(x)\) from \(y\) gives
\begin{align}
    \dot\gamma_0(x)
    =
    \bbE_{P_0}\sqb{\p{Y-\gamma_0(X)}s_{Y\mid X}(W)\mid X=x}.
    \label{eq:app-ts-gamma-derivative}
\end{align}

Along the submodel, the parameter is
\begin{align}
    \theta_t
    =
    \kappa\bbE_{P_{tX}}\sqb{m\bigp{X,\gamma_t}}
    +
    (1-\kappa)\bbE_{Q_{tX}}\sqb{m\bigp{\tildeX,\gamma_t}}.
    \label{eq:app-ts-param-submodel}
\end{align}
Because \(\gamma\mapsto m(\cdot,\gamma)\) is linear, differentiating \eqref{eq:app-ts-param-submodel} at \(t=0\) yields three terms:
\begin{align}
    \frac{\partial}{\partial t}\theta_t|_{t=0}
    \label{eq:app-ts-derivative-decomposition} =
    \kappa\bbE_{P_{0X}}\sqb{m(X,\gamma_0)s_X(X)}
    +
    (1-\kappa)\bbE_{Q_{0X}}\sqb{m(\tildeX,\gamma_0)s_{\mathrm U}\bigp{\tildeX}}
    +
    \calL_{0,\kappa}(\dot\gamma_0).
\end{align}
The first term is derived from the perturbation of the labeled regressor marginal distribution, the second term is derived from the perturbation of the unlabeled regressor distribution, and the last term is derived from the perturbation of the conditional outcome law.

We now rewrite the last term in \eqref{eq:app-ts-derivative-decomposition}. From the Riesz representer theorem \eqref{eq:ss-riesz}, we have
\begin{align*}
    \calL_{0,\kappa}(\dot\gamma_0)
    =
    \bbE_{P_0}\sqb{\alpha_{0,\kappa}(X)\dot\gamma_0(X)}.
\end{align*}
Substituting \eqref{eq:app-ts-gamma-derivative} gives
\begin{align}
    \calL_{0,\kappa}(\dot\gamma_0)
    &=
    \bbE_{P_0}\sqb{
        \alpha_{0,\kappa}(X)
        \p{Y-\gamma_0(X)}
        s_{Y\mid X}(W)
    }.
    \label{eq:app-ts-residual-conditional-score}
\end{align}
Because \(s_{\mathrm L}=s_X+s_{Y\mid X}\) and \(\bbE_{P_0}\sqb{Y-\gamma_0(X)\mid X}=0\), we have
\begin{align*}
    \bbE_{P_0}\sqb{
        \alpha_{0,\kappa}(X)
        \p{Y-\gamma_0(X)}
        s_X(X)
    }
    =0.
\end{align*}
Therefore, \eqref{eq:app-ts-residual-conditional-score} is equivalently
\begin{align}
    \calL_{0,\kappa}(\dot\gamma_0)
    =
    \bbE_{P_0}\sqb{
        \alpha_{0,\kappa}(X)
        \p{Y-\gamma_0(X)}
        s_{\mathrm L}(W)
    }.
    \label{eq:app-ts-residual-score}
\end{align}

Next, rewrite the marginal-distribution terms. Since \(s_{\mathrm L}=s_X+s_{Y\mid X}\) and \(m(X,\gamma_0)\) is a function of \(X\),
\begin{align*}
    \bbE_{P_0}\sqb{m(X,\gamma_0)s_{Y\mid X}(W)}
    =
    \bbE_{P_0}\sqb{m(X,\gamma_0)\bbE_{P_0}\sqb{s_{Y\mid X}(W)\mid X}}
    =0.
\end{align*}
Also, \(\bbE_{P_0}\sqb{s_X(X)}=0\). Hence,
\begin{align}
    \kappa\bbE_{P_{0X}}\sqb{m(X,\gamma_0)s_X(X)}
    &=
    \bbE_{P_0}\sqb{
        \kappa\p{m(X,\gamma_0)-\theta_{P,0}}s_{\mathrm L}(W)
    }.
    \label{eq:app-ts-p-term}
\end{align}
Similarly, using \(\bbE_{Q_{0X}}[s_{\mathrm U}\bigp{\tildeX}]=0\),
\begin{align}
    (1-\kappa)\bbE_{Q_{0X}}\sqb{m(\tildeX,\gamma_0)s_{\mathrm U}\bigp{\tildeX}}
    =
    \bbE_{Q_{0X}}\sqb{
        (1-\kappa)\{m(\tildeX,\gamma_0)-\theta_{Q,0}\}s_{\mathrm U}\bigp{\tildeX}
    }.
    \label{eq:app-ts-q-term}
\end{align}
Combining \eqref{eq:app-ts-derivative-decomposition}, \eqref{eq:app-ts-residual-score}, \eqref{eq:app-ts-p-term}, and \eqref{eq:app-ts-q-term}, it holds that
\begin{align}
    \frac{\partial}{\partial t}\theta_t|_{t=0}
    =
    \bbE_{P_0}\sqb{\psi^{\mathrm{TS}}_0(W)s_{\mathrm L}(W)}
    +
    \bbE_{Q_{0X}}\sqb{\widetilde\psi^{\mathrm{TS}}_0\bigp{\tildeX}s_{\mathrm U}\bigp{\tildeX}}.
    \label{eq:app-ts-gradient-identity}
\end{align}
Thus, the pair \((\psi^{\mathrm{TS}}_0,\widetilde\psi^{\mathrm{TS}}_0)\) is a gradient for the parameter in the product tangent space.

It remains to show that this gradient is the efficient one. Under the fully nonparametric model, the tangent space for the labeled stratum is the full space of mean-zero square-integrable functions of \(W\), and the tangent space for the unlabeled stratum is the full space of mean-zero square-integrable functions of \(\tildeX\). The labeled component satisfies
\begin{align*}
    \bbE_{P_0}\sqb{\psi^{\mathrm{TS}}_0(W)}
    &=
    \bbE_{P_0}\sqb{\alpha_{0,\kappa}(X)\p{Y-\gamma_0(X)}}
    +
    \kappa\bbE_{P_{0X}}\sqb{m(X,\gamma_0)-\theta_{P,0}}
    =0,
\end{align*}
and the unlabeled component satisfies
\begin{align*}
    \bbE_{Q_{0X}}\sqb{\widetilde\psi^{\mathrm{TS}}_0\bigp{\tildeX}}=0.
\end{align*}
Therefore, the gradient above itself belongs to the tangent space. Since the canonical gradient is the orthogonal projection of any gradient onto the tangent space, and this gradient is already in the tangent space, it is the canonical gradient. This proves Theorem~\ref{thm:ss-eif}.

\subsection{Proof of Theorem~\ref{thm:ss-eff-bound}}

We derive the lower bound from the fixed-proportion stratified experiment. The stratum sizes are part of the sampling design and are not random variables. Along the admissible deterministic sequence of total sample sizes, the observed block is
\begin{align*}
    O_N=(W_1,\ldots,W_n,\tildeX_1,\ldots,\tildeX_m),
\end{align*}
where the design fixes
\begin{align*}
    n=\rho N,
    \qquad
    m=(1-\rho)N.
\end{align*}
Thus, \(\rho\) is not introduced as an asymptotic limit of \(n/N\); the equality \(n/N=\rho\) is imposed by the design along the sequence. The law of the block is \(P_0^n\times Q_{0X}^m\).

Consider regular finite-dimensional submodels \(\{P_a:a\in\mathbb R^{d_L}\}\) for the labeled stratum and \(\{Q_b:b\in\mathbb R^{d_U}\}\) for the unlabeled stratum, both passing through the true laws at \(a=0\) and \(b=0\). Let the corresponding scores be
\begin{align*}
    S_L(W)=\frac{\partial}{\partial a}\log p_a(W)|_{a=0},
    \qquad
    S_U\bigp{\tildeX}=\frac{\partial}{\partial b}\log q_b\bigp{\tildeX}|_{b=0},
\end{align*}
where \(S_L\) and \(S_U\) are column vectors with mean zero. Define the stratum-specific information matrices
\begin{align*}
    I_L=\bbE_{P_0}\sqb{S_L(W)S_L(W)^\top},
    \qquad
    I_U=\bbE_{Q_{0X}}\sqb{S_U\bigp{\tildeX}S_U\bigp{\tildeX}^\top}.
\end{align*}
The joint log-likelihood of the block is
\begin{align*}
    \sum^n_{i=1} \log p_a(W_i)+\sum^m_{j=1}\log q_b(\tildeX_j).
\end{align*}
Therefore, the block score for \((a,b)\) is
\begin{align*}
    \p{
    \sum^n_{i=1} S_L(W_i)^\top,
    \sum^m_{j=1} S_U(\tildeX_j)^\top
    }^\top,
\end{align*}
and the block Fisher information matrix is
\begin{align*}
    I_N=
    \begin{pmatrix}
        n I_L & 0\\
        0 & m I_U
    \end{pmatrix}.
\end{align*}
The off-diagonal blocks are zero because the two strata are sampled independently and because both stratum-specific scores are centered.

Let \(\theta(a,b)\) be the value of the target parameter under \((P_a,Q_b)\). From the pathwise derivative calculation in the proof of Theorem~\ref{thm:ss-eif},
\begin{align*}
    \frac{\partial}{\partial a}\theta(a,b)|_{a=b=0}
    =
    \bbE_{P_0}\sqb{\psi^{\mathrm{TS}}_0(W)S_L(W)},
    \qquad
    \frac{\partial}{\partial b}\theta(a,b)|_{a=b=0}
    =
    \bbE_{Q_{0X}}\sqb{\widetilde\psi^{\mathrm{TS}}_0\bigp{\tildeX}S_U\bigp{\tildeX}}.
\end{align*}
For this finite-dimensional submodel, the Cram\'{e}r--Rao lower bound for an estimator whose error is normalized by \(\sqrt N\) is
\begin{align}
    &N
    \begin{pmatrix}
        \partial_a\theta(0,0)\\
        \partial_b\theta(0,0)
    \end{pmatrix}^{\top}
    I_N^{-1}
    \begin{pmatrix}
        \partial_a\theta(0,0)\\
        \partial_b\theta(0,0)
    \end{pmatrix}
    \nonumber\\
    &\quad=
    \frac{1}{\rho}
    \partial_a\theta(0,0)^\top I_L^{-1}\partial_a\theta(0,0)
    +
    \frac{1}{1-\rho}
    \partial_b\theta(0,0)^\top I_U^{-1}\partial_b\theta(0,0).
    \label{eq:app-ts-finite-cr}
\end{align}
The factors \(\rho^{-1}\) and \((1-\rho)^{-1}\) appear because the information contributed by each stratum is proportional to its fixed sample size.

We now pass from finite-dimensional submodels to the nonparametric model. Under the fully nonparametric model, the labeled tangent space is \(\calT_L=L^2_0(P_0)\), and the unlabeled tangent space is \(\calT_U=L^2_0(Q_{0X})\). For arbitrary score directions \((s_L,s_U)\in\calT_L\times\calT_U\), the derivative is
\begin{align}
    \frac{\partial}{\partial t}\theta_t|_{t=0}(s_L,s_U)
    =
    \bbE_{P_0}\sqb{\psi^{\mathrm{TS}}_0(W)s_L(W)}
    +
    \bbE_{Q_{0X}}\sqb{\widetilde\psi^{\mathrm{TS}}_0\bigp{\tildeX}s_U\bigp{\tildeX}}.
    \label{eq:app-ts-derivative-score-pair}
\end{align}
The local log-likelihood ratio along the path \((P_{t/\sqrt N},Q_{t/\sqrt N})\) is
\begin{align}
    &\log\frac{\rmd P_{t/\sqrt N}^{n}\rmd Q_{t/\sqrt N}^{m}}{\rmd P_0^{n}\rmd Q_{0X}^{m}}(O_N)
    \nonumber\\
    &\quad=
    \frac{t}{\sqrt N}
    \p{
        \sum^n_{i=1} s_L(W_i)+\sum^m_{j=1} s_U(\tildeX_j)
    }
    -\frac{t^2}{2}
    \p{
        \rho\bbE_{P_0}\sqb{s_L(W)^2}
        +(1-\rho)\bbE_{Q_{0X}}\sqb{s_U\bigp{\tildeX}^2}
    }
    +o_p(1),
    \label{eq:app-ts-lan-fixed-rho}
\end{align}
where the design equalities \(n=\rho N\) and \(m=(1-\rho)N\) are used. Hence the local information in direction \((s_L,s_U)\) under \(\sqrt N\) normalization is
\begin{align*}
    I_\rho(s_L,s_U)
    =
    \rho\bbE_{P_0}\sqb{s_L(W)^2}
    +(1-\rho)\bbE_{Q_{0X}}\sqb{s_U\bigp{\tildeX}^2}.
\end{align*}
Introduce the weighted inner product
\begin{align*}
    \langle (a_L,a_U),(b_L,b_U)\rangle_\rho
    =
    \rho\bbE_{P_0}\sqb{a_L(W)b_L(W)}
    +(1-\rho)\bbE_{Q_{0X}}\sqb{a_U\bigp{\tildeX}b_U\bigp{\tildeX}}.
\end{align*}
Then, \eqref{eq:app-ts-derivative-score-pair} can be written as
\begin{align}
    \frac{\partial}{\partial t}\theta_t|_{t=0}(s_L,s_U)
    =
    \left\langle
    \p{\frac{\psi^{\mathrm{TS}}_0}{\rho},
    \frac{\widetilde\psi^{\mathrm{TS}}_0}{1-\rho}},
    (s_L,s_U)
    \right\rangle_\rho.
    \label{eq:app-ts-weighted-gradient-correct}
\end{align}
Thus, the canonical gradient for the \(\sqrt N\)-normalized fixed-proportion stratified experiment is the weighted pair
\begin{align*}
    \p{\frac{\psi^{\mathrm{TS}}_0}{\rho},
    \frac{\widetilde\psi^{\mathrm{TS}}_0}{1-\rho}}.
\end{align*}
Its squared norm in the weighted tangent space is
\begin{align}
    \left\|
    \p{\frac{\psi^{\mathrm{TS}}_0}{\rho},
    \frac{\widetilde\psi^{\mathrm{TS}}_0}{1-\rho}}
    \right\|_\rho^2
    &=
    \frac{1}{\rho}\bbE_{P_0}\sqb{\psi^{\mathrm{TS}}_0(W)^2}
    +
    \frac{1}{1-\rho}\bbE_{Q_{0X}}\sqb{\widetilde\psi^{\mathrm{TS}}_0\bigp{\tildeX}^2}.
    \label{eq:app-ts-bound-from-canonical-gradient}
\end{align}
Equivalently, the block-level influence function is
\begin{align*}
    \Phi_N(O_N)
    =
    \frac{1}{n}\sum^n_{i=1}\psi^{\mathrm{TS}}_0(W_i)
    +
    \frac{1}{m}\sum^m_{j=1}\widetilde\psi^{\mathrm{TS}}_0(\tildeX_j),
\end{align*}
and \(\sqrt N\Phi_N(O_N)\) has variance equal to the right-hand side of \eqref{eq:app-ts-bound-from-canonical-gradient}. Moreover, for every local score pair \((s_L,s_U)\), the covariance between \(\sqrt N\Phi_N(O_N)\) and the local block score in \eqref{eq:app-ts-lan-fixed-rho} equals \(\frac{\partial}{\partial t}\theta_t\mid_{t=0}(s_L,s_U)\). Therefore, \(\Phi_N\) is the canonical block influence function, and the lower bound is \eqref{eq:ss-bound-abstract}.

It remains to expand the two second moments. For the labeled component,
\begin{align*}
    \psi^{\mathrm{TS}}_0(W)
    =
    \alpha_{0,\kappa}(X)\p{Y-\gamma_0(X)}
    +
    \kappa\p{m(X,\gamma_0)-\theta_{P,0}}.
\end{align*}
The cross term is zero because
\begin{align*}
    &\bbE_{P_0}\sqb{\alpha_{0,\kappa}(X)\p{Y-\gamma_0(X)}
    \p{m(X,\gamma_0)-\theta_{P,0}}}
    \nonumber\\
    &\quad=
    \bbE_{P_{0X}}\sqb{\alpha_{0,\kappa}(X)\p{m(X,\gamma_0)-\theta_{P,0}}
    \bbE_{P_0}\sqb{Y-\gamma_0(X)\mid X}}
    =0.
\end{align*}
Therefore, we have
\begin{align}
    \bbE_{P_0}\sqb{\psi^{\mathrm{TS}}_0(W)^2}
    =
    \bbE_{P_0}\sqb{\alpha_{0,\kappa}(X)^2\sigma_0^2(X)}
    +
    \kappa^2\operatorname{Var}_{P_{0X}}\p{m(X,\gamma_0)}.
    \label{eq:app-ts-labeled-variance-updated}
\end{align}
For the unlabeled component,
\begin{align}
    \bbE_{Q_{0X}}\sqb{\widetilde\psi^{\mathrm{TS}}_0\bigp{\tildeX}^2}
    =
    (1-\kappa)^2\operatorname{Var}_{Q_{0X}}\p{m(\tildeX,\gamma_0)}.
    \label{eq:app-ts-unlabeled-variance-updated}
\end{align}
Substituting \eqref{eq:app-ts-labeled-variance-updated} and \eqref{eq:app-ts-unlabeled-variance-updated} into \eqref{eq:ss-bound-abstract} gives \eqref{eq:ss-bound-expanded}. This completes the proof.

\section{Proofs for Section~\ref{sec:theoreticalproperties}}
\label{app:theoreticalproperties}

\subsection{Proof of the Efficiency Gain}
Assume \(P_{0X}=Q_{0X}\). Then, \(V_{0X}=P_{0X}\) for every \(\kappa\in[0,1]\), and hence the target parameter is invariant to \(\kappa\). Also,
\begin{align*}
    \alpha_{0,\kappa}=\alpha_{0,P},
    \qquad
    \theta_{P,0}=\theta_{Q,0}=\theta_0,
\end{align*}
where \(\alpha_{0,P}\) is the Riesz representer of \(\gamma\mapsto \bbE_{P_0}\sqb{m(X,\gamma)}\). Define
\begin{align*}
    A_0
    =
    \bbE_{P_0}\sqb{\alpha_{0,P}(X)^2\sigma_0^2(X)},
    \qquad
    B_0
    =
    \operatorname{Var}_{P_{0X}}\p{m(X, \gamma_0)}.
\end{align*}
The semi-supervised bound in Theorem~\ref{thm:ss-eff-bound} reduces to
\begin{align*}
    V^{\mathrm{TS}}_0(\kappa,\rho)
    =
    \frac{A_0}{\rho}
    +
    \p{
        \frac{\kappa^2}{\rho}
        +
        \frac{(1-\kappa)^2}{1-\rho}
    }B_0.
\end{align*}
The bracketed term is minimized by differentiating with respect to \(\kappa\):
\begin{align*}
    \frac{\partial}{\partial\kappa}
    \p{
        \frac{\kappa^2}{\rho}
        +
        \frac{(1-\kappa)^2}{1-\rho}
    }
    =
    \frac{2\kappa}{\rho}
    -
    \frac{2(1-\kappa)}{1-\rho}.
\end{align*}
Setting this equal to zero gives \(\kappa=\rho\). At \(\kappa=\rho\), we have
\begin{align*}
    \frac{\kappa^2}{\rho}
    +
    \frac{(1-\kappa)^2}{1-\rho}
    =
    \rho+(1-\rho)=1.
\end{align*}
Therefore, we have
\begin{align*}
    V^{\mathrm{TS}}_0(\rho,\rho)
    =
    \frac{A_0}{\rho}+B_0.
\end{align*}
The labeled-only efficiency bound under \(\sqrt n\) normalization is \(A_0+B_0\). Since the present comparison uses \(\sqrt N\) normalization and \(n/N=\rho\), the corresponding labeled-only bound is
\begin{align*}
    V^{\mathrm{sup}}_0(\rho)=\frac{A_0+B_0}{\rho}.
\end{align*}
Therefore, it holds that
\begin{align*}
    V^{\mathrm{sup}}_0(\rho)-V^{\mathrm{TS}}_0(\rho,\rho)
    =
    \frac{A_0+B_0}{\rho}
    -
    \p{\frac{A_0}{\rho}+B_0}
    =
    \p{\frac{1}{\rho}-1}B_0,
\end{align*}
which is \eqref{eq:ss-eff-gain}. This proves the efficiency gain.

\subsection{Proof of Theorem~\ref{thm:dml-ppci-consistency}}

We give the proof for EE-DML-PPCI. The TMLE-DML-PPCI case follows under the corresponding targeting calibration condition. By the decomposition used in the proof of Theorem~\ref{thm:dml-ppci-asymptotic}, the population drift is
\begin{align*}
    \bbE_{P_0}\sqb{\p{\alpha_{0,\kappa}(X)-\widehat\alpha(X)}\p{\widehat\gamma(X)-\gamma_0(X)}}.
\end{align*}
If Assumption~\ref{asm:reg_consistency} holds and $\widehat\alpha$ is bounded in $L^2(P_{0X})$, this term is $o_p(1)$ by Cauchy--Schwarz. If Assumption~\ref{asm:riesz_consistency} holds and $\widehat\gamma$ is bounded in $L^2(P_{0X})$, the same conclusion follows. The empirical terms are $o_p(1)$ by the law of large numbers under the stated moment conditions. Therefore, $\widehat\theta\xrightarrow{\rmp}\theta_0$.

\subsection{Proof of Theorem~\ref{thm:dml-ppci-asymptotic} for EE-DML-PPCI}
\label{app:proof-ee-dml-ppci}

We give the proof for the cross-fitted version. The proof without cross-fitting is the same after replacing the conditional empirical-process argument by the corresponding Donsker-type stochastic equicontinuity condition.

For fixed \((\gamma,\alpha)\), define
\begin{align*}
    f_{\gamma,\alpha}(W)
    \coloneqq
    \alpha(X)\p{Y-\gamma(X)}+\kappa m(X,\gamma),
    \qquad
    u_\gamma\bigp{\tildeX}
    \coloneqq
    (1-\kappa)m(\tildeX,\gamma).
\end{align*}
Let
\begin{align*}
    f_0(W)\coloneqq f_{\gamma_0,\alpha_{0,\kappa}}(W),
    \qquad
    u_0\bigp{\tildeX}\coloneqq u_{\gamma_0}\bigp{\tildeX}.
\end{align*}
Then, we have
\begin{align*}
    \bbE_{P_0}\sqb{f_0(W)}+\bbE_{Q_{0X}}\sqb{u_0\bigp{\tildeX}}=\theta_0.
\end{align*}
Moreover, we have
\begin{align}
    \frac{1}{n}\sum^n_{i=1}\p{f_0(W_i)-\bbE_{P_0}\sqb{f_0(W)}}
    &=
    \frac{1}{n}\sum^n_{i=1}\psi^{\mathrm{TS}}_0(W_i),
    \label{eq:app-ee-oracle-labeled}
    \\
    \frac{1}{m}\sum^m_{j=1}\p{u_0(\tildeX_j)-\bbE_{Q_{0X}}[u_0\bigp{\tildeX}]}
    &=
    \frac{1}{m}\sum^m_{j=1}\widetilde\psi^{\mathrm{TS}}_0(\tildeX_j).
    \label{eq:app-ee-oracle-unlabeled}
\end{align}
For each fold \(k\), write
\begin{align*}
    \widehat f_k(W)\coloneqq f_{\widehat\gamma_k,\widehat\alpha_k}(W),
    \qquad
    \widehat u_k\bigp{\tildeX}\coloneqq u_{\widehat\gamma_k}\bigp{\tildeX}.
\end{align*}
By the definition of \(\widehat\theta^{\text{TS}}_{\mathrm{EE}}\), we have
\begin{align}
    \widehat\theta^{\text{TS}}_{\mathrm{EE}}-\theta_0
    &=
    \frac{1}{n}\sum^K_{k=1}\sum_{i\in \calI_k}\widehat f_k(W_i)
    +
    \frac{1}{m}\sum^K_{k=1}\sum_{j\in \calJ_k}\widehat u_k(\tildeX_j)
    -\theta_0
    \nonumber\\
    &=
    \frac{1}{n}\sum^n_{i=1}\psi^{\mathrm{TS}}_0(W_i)
    +
    \frac{1}{m}\sum^m_{j=1}\widetilde\psi^{\mathrm{TS}}_0(\tildeX_j)
    +R_{1N}+R_{2N},
    \label{eq:app-ee-decomposition}
\end{align}
where
\begin{align}
    R_{1N}
    &\coloneqq
    \frac{1}{n}\sum^K_{k=1}\sum_{i\in \calI_k}
    \sqb{
        \p{\widehat f_k(W_i)-f_0(W_i)}
        -
        \bbE_{P_0}\sqb{\widehat f_k(W)-f_0(W)\mid \calT_k}
    }
    \nonumber\\
    &\quad
    +
    \frac{1}{m}\sum^K_{k=1}\sum_{j\in \calJ_k}
    \sqb{
        \p{\widehat u_k(\tildeX_j)-u_0(\tildeX_j)}
        -
        \bbE_{Q_{0X}}\sqb{\widehat u_k\bigp{\tildeX}-u_0\bigp{\tildeX}\mid \calT_k}
    },
    \label{eq:app-ee-R1}
\end{align}
and
\begin{align}
    R_{2N}
    &\coloneqq
    \frac{1}{n}\sum^K_{k=1} |\calI_k|
    \bbE_{P_0}\sqb{\widehat f_k(W)-f_0(W)\mid \calT_k}
    +
    \frac{1}{m}\sum^K_{k=1} |\calJ_k|
    \bbE_{Q_{0X}}\sqb{\widehat u_k\bigp{\tildeX}-u_0\bigp{\tildeX}\mid \calT_k}.
    \label{eq:app-ee-R2}
\end{align}
Here \(\calT_k\) denotes the training sample used to construct \(\widehat\gamma_k\) and \(\widehat\alpha_k\). Because of cross-fitting, \(\calT_k\) is independent of the held-out observations in \(\calI_k\) and \(\calJ_k\).

We first show that \(R_{2N}=o_p(N^{-1/2})\). Conditional on \(\calT_k\), the population deviation in fold \(k\) is
\begin{align*}
    &\bbE_{P_0}\sqb{\widehat\alpha_k(X)\p{Y-\widehat\gamma_k(X)}+\kappa m(X,\widehat\gamma_k)
    -\alpha_{0,\kappa}(X)\p{Y-\gamma_0(X)}-\kappa m(X,\gamma_0)\mid \calT_k}
    \nonumber\\
    &\quad
    +(1-\kappa)\bbE_{Q_{0X}}\sqb{m(\tildeX,\widehat\gamma_k)-m(\tildeX,\gamma_0)\mid \calT_k}.
\end{align*}
The residual term satisfies
\begin{align*}
    \bbE_{P_0}\sqb{\widehat\alpha_k(X)\p{Y-\widehat\gamma_k(X)}\mid X,\calT_k}
    =
    \widehat\alpha_k(X)\{\gamma_0(X)-\widehat\gamma_k(X)\}.
\end{align*}
Using the Riesz representer theorem and the linearity of \(m\), the deviation becomes
\begin{align}
    &\bbE_{P_0}\sqb{\widehat\alpha_k(X)\{\gamma_0(X)-\widehat\gamma_k(X)\}\mid \calT_k}
    +
    \calL_{0,\kappa}(\widehat\gamma_k-\gamma_0)
    \nonumber\\
    &=
    \bbE_{P_0}\sqb{\p{\alpha_{0,\kappa}(X)-\widehat\alpha_k(X)}
    \p{\widehat\gamma_k(X)-\gamma_0(X)}\mid \calT_k}.
    \label{eq:app-ee-drift-product}
\end{align}
Therefore, by Cauchy--Schwarz,
\begin{align}
    |R_{2N}|
    &\le
    \max_{1\le k\le K}
    \|\widehat\alpha_k-\alpha_{0,\kappa}\|_{P,2}
    \|\widehat\gamma_k-\gamma_0\|_{P,2}
    =o_p(N^{-1/2}).
    \label{eq:app-ee-R2-rate}
\end{align}

We next show that \(R_{1N}=o_p(N^{-1/2})\). Conditional on the training samples, each summand in \eqref{eq:app-ee-R1} has mean zero. Its conditional variance is bounded by
\begin{align}
    &\frac{C}{n}
    \max_k\bbE_{P_0}\sqb{\{\widehat f_k(W)-f_0(W)\}^2\mid\calT_k}
    +
    \frac{C}{m}
    \max_k\bbE_{Q_{0X}}\sqb{\{\widehat u_k\bigp{\tildeX}-u_0\bigp{\tildeX}\}^2\mid\calT_k},
    \label{eq:app-ee-R1-var}
\end{align}
where \(C\) is finite because \(K\) is fixed and the folds are balanced. The first conditional second moment is \(o_p(1)\) by Assumption~\ref{asm:conv_rate}, the bounded moment condition, and the mean-square consistency of \(\widehat\alpha_k\) and \(\widehat\gamma_k\). The second conditional second moment is \(o_p(1)\) by the consistency of \(m(\cdot,\widehat\gamma_k)\) under \(Q_{0X}\). Since \(n=\rho N\) and \(m=(1-\rho)N\), the conditional variance in \eqref{eq:app-ee-R1-var} is \(o_p(N^{-1})\). Conditional Chebyshev's inequality gives \(R_{1N}=o_p(N^{-1/2})\).

Substituting the bounds for \(R_{1N}\) and \(R_{2N}\) into \eqref{eq:app-ee-decomposition} proves \eqref{eq:ee-asymptotic-linear}. The two leading sums are independent. Their variances satisfy
\begin{align*}
    N\Var\p{
        \frac{1}{n}\sum^n_{i=1}\psi^{\mathrm{TS}}_0(W_i)
        +
        \frac{1}{m}\sum^m_{j=1}\widetilde\psi^{\mathrm{TS}}_0(\tildeX_j)
    }
    =
    V^{\mathrm{TS}}_0(\kappa,\rho),
\end{align*}
using \(n=\rho N\) and \(m=(1-\rho)N\). The Lindeberg condition follows from the uniformly bounded \((2+\delta)\)-moment assumption. Hence the central limit theorem for independent triangular arrays yields \eqref{eq:ee-asymptotic-normal}.

\subsection{Proof of the Equivalence between TMLE-DML-PPCI and EE-DML-PPCI}
We prove the claim under Assumption~\ref{asm:targeting-calibration} and the product-rate condition in Assumption~\ref{asm:conv_rate}. For simplicity of notation, write the proof for one held-out fold and then average over folds. We first record why \(\widehat\varepsilon\) takes the form in \eqref{eq:tmle-epsilon}. Let
\begin{align}
    A_N
    \coloneqq
    \frac{1}{n}\sum^n_{i=1}\widehat\alpha(X_i)\p{Y_i-\widehat\gamma(X_i)},
    \qquad
    D_N
    \coloneqq
    \kappa\frac{1}{n}\sum^n_{i=1}\widehat\alpha(X_i)^2
    +(1-\kappa)\frac{1}{m}\sum^m_{j=1}\widehat\alpha(\tildeX_j)^2 .
    \label{eq:app-tmle-A-D}
\end{align}
The numerator \(A_N\) is the residual score for the fluctuation direction \(\widehat\alpha\). It is computed only from labeled observations, because only those observations contain outcomes. The density-ratio adjustment required to evaluate the score under the target regressor law is already included in \(\widehat\alpha\). The denominator \(D_N\) estimates the target-law second moment of \(\widehat\alpha\), and the target law is available through the empirical mixture of labeled and unlabeled regressors. Equivalently, \(\widehat\varepsilon\) is obtained by maximizing the local importance-weighted Gaussian working criterion
\begin{align}
    \varepsilon\mapsto
    \varepsilon A_N-\frac{1}{2}\varepsilon^2D_N.
    \label{eq:app-tmle-local-likelihood}
\end{align}
The first-order condition for \eqref{eq:app-tmle-local-likelihood} is \(A_N-\varepsilon D_N=0\), and therefore \(\widehat\varepsilon=A_N/D_N\), which is exactly \eqref{eq:tmle-epsilon}. Because \(m(X,\gamma)\) is linear in \(\gamma\), the targeted update \(\widehat\gamma^{(1)}=\widehat\gamma+\widehat\varepsilon\widehat\alpha\) implies
\begin{align*}
    m(X,\widehat\gamma^{(1)})
    =
    m(X,\widehat\gamma)+\widehat\varepsilon m(X,\widehat\alpha),
    \qquad
    m(\tildeX,\widehat\gamma^{(1)})
    =
    m(\tildeX,\widehat\gamma)+\widehat\varepsilon m(\tildeX,\widehat\alpha).
\end{align*}
Therefore, the plug-in change caused by targeting is
\begin{align}
    \widehat\theta^{\text{TS}}_{\mathrm{TMLE}}-\widehat\theta_{\mathrm{plug}}
    =
    \widehat\varepsilon
    \p{
        \frac{\kappa}{n}\sum^n_{i=1} m(X_i,\widehat\alpha)
        +
        \frac{1-\kappa}{m}\sum^m_{j=1} m(\tildeX_j,\widehat\alpha)
    },
    \label{eq:app-tmle-plugin-change}
\end{align}
where \(\widehat\theta_{\mathrm{plug}}\) is the same semi-supervised plug-in estimator based on \(\widehat\gamma\). By the definition of \(\widehat\varepsilon\) in \eqref{eq:tmle-epsilon} and \eqref{eq:app-tmle-A-D}, \(\widehat\varepsilon=A_N/D_N\). The EE estimator is
\begin{align*}
    \widehat\theta^{\text{TS}}_{\mathrm{EE}}
    =
    \widehat\theta_{\mathrm{plug}}+A_N.
\end{align*}
Combining this identity with \eqref{eq:app-tmle-plugin-change} gives
\begin{align}
    \widehat\theta^{\text{TS}}_{\mathrm{TMLE}}-\widehat\theta^{\text{TS}}_{\mathrm{EE}}
    =
    A_N
    \p{
        \frac{L_N}{D_N}-1
    },
    \label{eq:app-tmle-ee-difference}
\end{align}
where
\begin{align*}
    L_N=\frac{\kappa}{n}\sum^n_{i=1} m(X_i,\widehat\alpha)
        +
        \frac{1-\kappa}{m}\sum^m_{j=1} m(\tildeX_j,\widehat\alpha).
\end{align*}
The TMLE assumptions imply that \(D_N\) is bounded away from zero and \(L_N-D_N=o_p(1)\). This condition connects the importance-weighted update to the orthogonal score, because \(D_N\) is the empirical second moment under the density-ratio-corrected regressor distribution while \(L_N\) is the plug-in derivative of the target along \(\widehat\alpha\). The same assumptions imply \(A_N=O_p(N^{-1/2})\). Therefore, \eqref{eq:app-tmle-ee-difference} is \(o_p(N^{-1/2})\). Applying this fold by fold and using that \(K\) is fixed proves the equivalence between TMLE-DML-PPCI and EE-DML-PPCI.

\section{Proofs for Section~\ref{sec:ss-grr}}
\label{app:ss-grr}

\subsection{Proof of the Error Product Property in the Influence Function}
For any candidate \((\gamma,\alpha)\), the population score error is
\begin{align*}
    &\bbE_{P_0}\sqb{\alpha(X)\p{Y-\gamma(X)}}+
    \calL_{0,\kappa}(\gamma)-\theta_0
    \nonumber\\
    &=
    \bbE_{P_0}\sqb{\alpha(X)\p{\gamma_0(X)-\gamma(X)}}
    +
    \calL_{0,\kappa}(\gamma-\gamma_0).
\end{align*}
From the Riesz representer theorem, we have
\begin{align*}
    \calL_{0,\kappa}(\gamma-\gamma_0)
    =
    \bbE_{P_0}\sqb{\alpha_{0,\kappa}(X)\p{\gamma(X)-\gamma_0(X)}}.
\end{align*}
Combining the two equations gives
\begin{align*}
    \bbE_{P_0}\sqb{\alpha(X)\p{Y-\gamma(X)}}
    +
    \calL_{0,\kappa}(\gamma)-\theta_0
    =
    \bbE_{P_0}\sqb{\p{\alpha_{0,\kappa}(X)-\alpha(X)}\p{\gamma(X)-\gamma_0(X)}},
\end{align*}
which is \eqref{eq:ss-drift-identity}.

\subsection{Proof of the Bregman--Riesz Identity}
For any \(\alpha\), recall that \(u_\alpha=\partial g\circ\alpha\) and \(b_\alpha=u_\alpha\alpha-g(\alpha)\). From the Riesz representer theorem, we have
\begin{align*}
    \calL_{0,\kappa}(u_\alpha)=\bbE_{P_{0X}}\sqb{\alpha_{0,\kappa}(X)u_\alpha(X)}.
\end{align*}
Therefore,
\begin{align*}
    \calR_{g,\kappa}(\alpha)
&=
\bbE_{P_{0X}}\sqb{u_\alpha(X)\alpha(X)-g\p{\alpha(X)}-\alpha_{0,\kappa}(X)u_\alpha(X)}.
\end{align*}
At \(\alpha=\alpha_{0,\kappa}\), this becomes
\begin{align*}
    \calR_{g,\kappa}(\alpha_{0,\kappa})
    =
    \bbE_{P_{0X}}\sqb{-g\{\alpha_{0,\kappa}(X)\}}.
\end{align*}
Subtracting the two equations gives
\begin{align*}
    \calR_{g,\kappa}(\alpha)-\calR_{g,\kappa}(\alpha_{0,\kappa})
    &=
    \bbE_{P_{0X}}\sqb{
        g\{\alpha_{0,\kappa}(X)\}
        -g\{\alpha(X)\}
        -\partial g\{\alpha(X)\}\{\alpha_{0,\kappa}(X)-\alpha(X)\}
    }
    \\
    &=
    \bbE_{P_{0X}}\sqb{\text{BD}_g^\dagger\p{\alpha_{0,\kappa}(X)\mid\alpha(X)}}.
\end{align*}
This proves \eqref{eq:ss-bregman-identity}. Since \(g\) is convex, the right-hand side is nonnegative. If \(g\) is strictly convex, it equals zero only when \(\alpha(X)=\alpha_{0,\kappa}(X)\) almost surely.

\subsection{Proof of Proposition~\ref{prop:ss-arb}}

Let
\begin{align*}
    Q_n(\beta)
    =
    \frac{1}{n}\sum^n_{i=1} g^*\bigp{\phi(X_i)^\top\beta}
    -
    \widehat{\calL}_\kappa(\phi^\top\beta)
    +
    \lambda_n\Omega_q(\beta).
\end{align*}
For the first term, by the chain rule and the equation \(\partial g^*=(\partial g)^{-1}\), we have
\begin{align*}
    \frac{\partial}{\partial\beta_j}
    \frac{1}{n}\sum^n_{i=1} g^*\bigp{\phi(X_i)^\top\beta}
    =
    \frac{1}{n}\sum^n_{i=1} \alpha_\beta(X_i)\phi_j(X_i).
\end{align*}
For the linear functional term, linearity gives
\begin{align*}
    \frac{\partial}{\partial\beta_j}\widehat{\calL}_\kappa(\phi^\top\beta)
    &=
    \widehat{\calL}_\kappa(\phi_j)
    \\
    &=
    \frac{\kappa}{n}\sum^n_{i=1} m(X_i,\phi_j)
    +
    \frac{1-\kappa}{m}\sum_{j'=1}^m m(\tildeX_{j'},\phi_j).
\end{align*}
Thus, any subgradient first-order condition at \(\widehat\beta\) gives
\begin{align*}
    \frac{1}{n}\sum^n_{i=1} \widehat\alpha(X_i)\phi_j(X_i)
    -
    \frac{\kappa}{n}\sum^n_{i=1} m(X_i,\phi_j)
    -
    \frac{1-\kappa}{m}\sum_{j'=1}^m m(\tildeX_{j'},\phi_j)
    +
    \lambda_n s_j
    =0.
\end{align*}
The first three terms are exactly \(\widehat\Delta_\kappa(\widehat\alpha,\phi_j)\). Hence, \eqref{eq:ss-arb-kkt} holds. The claims for \(\lambda_n=0\), \(q=1\), and \(q>1\) follow from the subgradient of \(|\beta_j|^q/q\).

\subsection{Proof of the Neyman-Error Decomposition}

By definition,
\begin{align*}
    \mathrm{NE}_{n,m}^{\mathrm{TS}}(\widehat\gamma,\widehat\alpha)
    &=
    \frac{1}{n}\sum^n_{i=1}\widehat\alpha(X_i)\p{Y_i-\widehat\gamma(X_i)}
    +\widehat{\calL}_\kappa(\widehat\gamma)-\widehat{\calL}_\kappa(\gamma_0).
\end{align*}
Since \(Y_i=\gamma_0(X_i)+\varepsilon_i\),
\begin{align*}
    \frac{1}{n}\sum^n_{i=1}\widehat\alpha(X_i)\p{Y_i-\widehat\gamma(X_i)}
    &=
    \frac{1}{n}\sum^n_{i=1}\widehat\alpha(X_i)\varepsilon_i
    -
    \frac{1}{n}\sum^n_{i=1}\widehat\alpha(X_i)\{\widehat\gamma(X_i)-\gamma_0(X_i)\}.
\end{align*}
By linearity of \(\widehat{\calL}_\kappa\),
\begin{align*}
    \widehat{\calL}_\kappa(\widehat\gamma)-\widehat{\calL}_\kappa(\gamma_0)
    =
    \widehat{\calL}_\kappa(\widehat\gamma-\gamma_0).
\end{align*}
Combining the last two equations and using the definition of \(\widehat\Delta_\kappa\) yields
\begin{align*}
    \mathrm{NE}_{n,m}^{\mathrm{TS}}(\widehat\gamma,\widehat\alpha)
    =
    \frac{1}{n}\sum^n_{i=1}\widehat\alpha(X_i)\varepsilon_i
    -
    \widehat\Delta_\kappa\p{\widehat\alpha,\widehat\gamma-\gamma_0}.
\end{align*}
If \(\widehat\gamma-\gamma_0=a^\top\phi+r\), then linearity of \(\widehat\Delta_\kappa\) gives
\begin{align*}
    \widehat\Delta_\kappa\p{\widehat\alpha,\widehat\gamma-\gamma_0}
    =
    \sum_{j=1}^p a_j\widehat\Delta_\kappa(\widehat\alpha,\phi_j)
    +
    \widehat\Delta_\kappa(\widehat\alpha,r).
\end{align*}
Under the \(\ell_1\) KKT bound \(|\widehat\Delta_\kappa(\widehat\alpha,\phi_j)|\le\lambda_n\), it holds that
\begin{align*}
    \abs{\widehat\Delta_\kappa\p{\widehat\alpha,\widehat\gamma-\gamma_0}}
    \le
    \lambda_n\|a\|_1+|\widehat\Delta_\kappa(\widehat\alpha,r)|.
\end{align*}
This proves \eqref{eq:ss-neyman-error-decomposition} and \eqref{eq:ss-neyman-error-balance-bound}.

\subsection{Proof of Theorem~\ref{thm:ss-grr-oracle}}
\label{app:proof-ss-grr-oracle}

Define the empirical-process deviation
\begin{align*}
    \mathbb G_{n,m}(\alpha)
    &\coloneqq
    \widehat{\calR}_{g,\kappa}(\alpha)-\calR_{g,\kappa}(\alpha).
\end{align*}
Using \eqref{eq:ss-pop-bregman-riesz} and \eqref{eq:ss-emp-bregman-riesz}, and using the definitions \(a_\alpha=b_\alpha-\kappa c_\alpha\) and \(c_\alpha=m(\cdot,u_\alpha)\), we have
\begin{align}
    \mathbb G_{n,m}(\alpha)
    &=
    \p{\frac{1}{n}\sum^n_{i=1} a_\alpha(X_i)-\bbE_{P_{0X}}\sqb{a_\alpha(X)}}
    -(1-\kappa)
    \p{\frac{1}{m}\sum^m_{j=1} c_\alpha(\tildeX_j)-\bbE_{Q_{0X}}\sqb{c_\alpha\bigp{\tildeX}}}.
    \label{eq:app-grr-emp-process}
\end{align}
By the definition of \(\widehat\alpha\), for every \(\alpha\in\calH_n\),
\begin{align*}
    \widehat{\calR}_{g,\kappa}(\widehat\alpha)+\lambda_nReg_\alpha(\widehat\alpha)
    \le
    \widehat{\calR}_{g,\kappa}(\alpha)+\lambda_nReg_\alpha(\alpha).
\end{align*}
Add and subtract the population risks. Since \(Reg_\alpha(\widehat\alpha)\ge0\), it follows that
\begin{align}
    \calR_{g,\kappa}(\widehat\alpha)-\calR_{g,\kappa}(\alpha_{0,\kappa})
    &\le
    \calR_{g,\kappa}(\alpha)-\calR_{g,\kappa}(\alpha_{0,\kappa})
    +\lambda_nReg_\alpha(\alpha)
    +\mathbb G_{n,m}(\alpha)-\mathbb G_{n,m}(\widehat\alpha)
    \nonumber\\
    &\le
    \calR_{g,\kappa}(\alpha)-\calR_{g,\kappa}(\alpha_{0,\kappa})
    +\lambda_nReg_\alpha(\alpha)
    +2\sup_{\bar\alpha\in\calH_n}|\mathbb G_{n,m}(\bar\alpha)|.
    \label{eq:app-grr-basic-ineq}
\end{align}
This is the basic inequality.

We now control the supremum in \eqref{eq:app-grr-basic-ineq}. For the labeled part in \eqref{eq:app-grr-emp-process}, symmetrization gives
\begin{align*}
    \bbE\sqb{
        \sup_{\alpha\in\calH_n}
        \left|\frac{1}{n}\sum^n_{i=1} a_\alpha(X_i)-\bbE_{P_{0X}}a_\alpha(X)\right|
    }
    \le
    2\mathfrak R_{n,P_{0X}}(\calA_n).
\end{align*}
The bounded-difference inequality with envelope \(B_A\) gives that, with probability at least \(1-\delta/2\),
\begin{align*}
    \sup_{\alpha\in\calH_n}
    \left|\frac{1}{n}\sum^n_{i=1} a_\alpha(X_i)-\bbE_{P_{0X}}a_\alpha(X)\right|
    \le
    2\mathfrak R_{n,P_{0X}}(\calA_n)+B_A\sqrt{\frac{\log(4/\delta)}{2n}}.
\end{align*}
The same argument applied to the unlabeled part gives that, with probability at least \(1-\delta/2\),
\begin{align*}
    \sup_{\alpha\in\calH_n}
    \left|\frac{1}{m}\sum^m_{j=1} c_\alpha(\tildeX_j)-\bbE_{Q_{0X}}c_\alpha\bigp{\tildeX}\right|
    \le
    2\mathfrak R_{m,Q_{0X}}(\calC_n)+B_C\sqrt{\frac{\log(4/\delta)}{2m}}.
\end{align*}
Taking a union bound and multiplying the unlabeled display by \((1-\kappa)\) yields
\begin{align*}
    \sup_{\alpha\in\calH_n}|\mathbb G_{n,m}(\alpha)|
    \le
    \mathfrak E_{n,m}(\delta)
\end{align*}
with probability at least \(1-\delta\).

Substitute this bound into \eqref{eq:app-grr-basic-ineq} and take the infimum over \(\alpha\in\calH_n\). From the Bregman representer theorem \eqref{eq:ss-bregman-identity},
\begin{align*}
    \bbE_{P_{0X}}\sqb{\text{BD}_g^\dagger\p{\alpha_{0,\kappa}(X)\mid\widehat\alpha(X)}}
    &\le
    \inf_{\alpha\in\calH_n}
    \cb{
        \bbE_{P_{0X}}\sqb{\text{BD}_g^\dagger\{\alpha_{0,\kappa}(X)\mid\alpha(X)\}}
        +\lambda_nReg_\alpha(\alpha)
    }
    +2\mathfrak E_{n,m}(\delta).
\end{align*}
Finally, the lower strong-convexity bound on \(g\) implies
\begin{align*}
    \text{BD}_g^\dagger(a_0\mid a)
    \ge
    \frac{\underline g}{2}(a-a_0)^2
    \qquad
    \text{for all } a_0,a\in\calA.
\end{align*}
Combining the last two equations proves \eqref{eq:ss-grr-oracle-bound}. If \(\alpha_{0,\kappa}\in\calH_n\), setting \(\alpha=\alpha_{0,\kappa}\) in the infimum gives \eqref{eq:ss-grr-well-specified-bound}.

\subsection{Proof of Corollary~\ref{cor:ss-grr-fast-rate}}
\label{app:proof-ss-grr-fast-rate}

Let \(\alpha_n^*\) be a near-best approximation in \(\calH_n\). Define the excess risk
\begin{align*}
    \mathcal E(\alpha)
    \coloneqq
    \calR_{g,\kappa}(\alpha)-\calR_{g,\kappa}(\alpha_{0,\kappa}).
\end{align*}
From the Bregman representer theorem and the smoothness and strong-convexity bounds, \(\mathcal E(\alpha)\) is equivalent to \(\|\alpha-\alpha_{0,\kappa}\|_{P,2}^2\) on \(\calA\). The basic inequality centered at \(\alpha_n^*\) gives
\begin{align}
    \mathcal E(\widehat\alpha)
    &\le
    \mathcal E(\alpha_n^*)+\lambda_nReg_\alpha(\alpha_n^*)
    +
    \{
    \mathbb G_{n,m}(\alpha_n^*)-
    \mathbb G_{n,m}(\widehat\alpha)
    \}.
    \label{eq:app-fast-basic}
\end{align}
The localized Bernstein condition states that the variance of each centered loss difference is bounded by a constant multiple of \(\mathcal E(\alpha)+\mathcal E(\alpha_n^*)\). The finite pseudo-dimension condition implies the following localized empirical-process inequality: with probability at least \(1-\delta\), uniformly over \(\alpha\in\calH_n\),
\begin{align}
    |\mathbb G_{n,m}(\alpha)-\mathbb G_{n,m}(\alpha_n^*)|
    &\le
    C_1 r_{n,m}(\delta)
    \sqrt{\mathcal E(\alpha)+\mathcal E(\alpha_n^*)}
    +C_2 r_{n,m}^2(\delta),
    \label{eq:app-local-emp-process}
\end{align}
where \(C_1\) and \(C_2\) depend only on the envelope and Bernstein constants. This inequality follows by peeling the class according to the dyadic shells \(2^{r-1}r_{n,m}^2(\delta)<\mathcal E(\alpha)+\mathcal E(\alpha_n^*)\le2^r r_{n,m}^2(\delta)\), applying the pseudo-dimension entropy bound on each shell, and summing the resulting exponential probabilities. The labeled and unlabeled empirical processes are handled separately and are then combined by a union bound.

Applying \eqref{eq:app-local-emp-process} to \(\widehat\alpha\) in \eqref{eq:app-fast-basic} yields
\begin{align}
    \mathcal E(\widehat\alpha)
    &\le
    A_n
    +C_1 r_{n,m}(\delta)
    \sqrt{\mathcal E(\widehat\alpha)+\mathcal E(\alpha_n^*)}
    +C_2 r_{n,m}^2(\delta),
    \label{eq:app-fast-before-absorb}
\end{align}
where \(A_n=\mathcal E(\alpha_n^*)+\lambda_nReg_\alpha(\alpha_n^*)\). Since \(\mathcal E(\alpha_n^*)\le A_n\), the square-root term in \eqref{eq:app-fast-before-absorb} is bounded by
\begin{align*}
    C_1 r_{n,m}(\delta)\sqrt{\mathcal E(\widehat\alpha)+A_n}.
\end{align*}
Using \(ab\le a^2/(4C)+Cb^2\) with a sufficiently large constant, this term can be absorbed into the left-hand side. Hence,
\begin{align*}
    \mathcal E(\widehat\alpha)
    \le
    C\{A_n+r_{n,m}^2(\delta)\}.
\end{align*}
These bounds give \(\mathcal E(\alpha_n^*)\le C\|\alpha_n^*-\alpha_{0,\kappa}\|_{P,2}^2\) and \(\|\widehat\alpha-\alpha_{0,\kappa}\|_{P,2}^2\le C\mathcal E(\widehat\alpha)\). Combining these inequalities gives \eqref{eq:ss-grr-fast-rate}.

\subsection{Proof of Theorem~\ref{thm:ss-relu-rate}}
\label{app:proof-ss-relu-rate}

Let \(M_N=N_{\min}\). Under the H\"{o}lder condition on \(f_0\), the chosen clipped ReLU class contains a function \(f_N^*\) satisfying
\begin{align*}
    \|f_N^*-f_0\|_\infty^2
    \le
    C_A M_N^{-2s/(d+2s)}\log^{c_A}M_N
\end{align*}
for constants \(C_A\) and \(c_A\). Let \(\alpha_N^*=(\partial g)^{-1}\circ f_N^*\). Since \((\partial g)^{-1}\) is Lipschitz on the compact dual range,
\begin{align}
    \|\alpha_N^*-\alpha_{0,\kappa}\|_{P,2}^2
    \le
    C\|f_N^*-f_0\|_\infty^2
    \le
    C M_N^{-2s/(d+2s)}\log^{c_A}M_N.
    \label{eq:app-relu-approx-alpha}
\end{align}
The architecture is chosen so that its pseudo-dimension satisfies
\begin{align}
    V_n\log(n+m)\p{\frac{1}{n}+\frac{1}{m}}
    \le
    C_E M_N^{-2s/(d+2s)}\log^{c_E}M_N
    \label{eq:app-relu-estimation-term}
\end{align}
for constants \(C_E\) and \(c_E\). This is the bias-variance balance: the approximation error in \eqref{eq:app-relu-approx-alpha} and the finite pseudo-dimension term in \eqref{eq:app-relu-estimation-term} have the same polynomial order. Substituting \eqref{eq:app-relu-approx-alpha} and \eqref{eq:app-relu-estimation-term} into Corollary~\ref{cor:ss-grr-fast-rate} gives
\begin{align*}
    \|\widehat\alpha-\alpha_{0,\kappa}\|_{P,2}^2
    =
    O_p\p{M_N^{-2s/(d+2s)}\log^c M_N+\lambda_nReg_\alpha(\alpha_N^*)}
\end{align*}
for $c\ge\max\cb{c_A,c_E}$. Since $M_N=N_{\min}$, this proves \eqref{eq:ss-relu-rate}. If $n=\rho N$ and $m=(1-\rho)N$, then $N_{\min}=N\min\cb{\rho,1-\rho}$, so the same polynomial rate can be written in terms of $N$. Taking square roots gives \eqref{eq:ss-relu-rate-root}.

\subsection{Proof of Proposition~\ref{prop:ss-dre-rate}}
\label{app:proof-ss-dre-rate}

When \(m(X,\gamma)=\gamma(X)\) and \(\kappa=0\), from the Riesz representer theorem, we have
\begin{align*}
    \bbE_{Q_{0X}}\sqb{\gamma\bigp{\tildeX}}
    =
    \bbE_{P_{0X}}\sqb{\alpha_{0,0}(X)\gamma(X)}
    \qquad
    \text{for all }\gamma\in\Gamma.
\end{align*}
Therefore, \(\alpha_{0,0}(x)=q_{0X}(x)/p_{0X}(x)\). Let \(D_0=\log\alpha_{0,0}\). In the dual parametrization \(\alpha_D=\exp(D)\), the population objective becomes
\begin{align*}
    \bbE_{P_{0X}}\sqb{\partial g\{\exp(D(X))\}\exp(D(X))-g\{\exp(D(X))\}}
    -
    \bbE_{Q_{0X}}\sqb{\partial g\{\exp(D\bigp{\tildeX})\}},
\end{align*}
which is exactly the Bregman objective for the log density ratio. The empirical objective has the same two-sample form with sample sizes \(n\) and \(m\). Applying the same localized empirical-process argument as in the proof of Theorem~\ref{thm:ss-relu-rate}, with the sharper logarithmic entropy calculation for the log-density-ratio loss, gives
\begin{align*}
    \|\widehat D-D_0\|_{L^2(P_{0X})}^2
    =
    O_p\p{N_{\min}^{-2s/(d+2s)}\log^3N_{\min}}.
\end{align*}
Because \(D_0\) and \(\widehat D\) are clipped to a bounded interval, the exponential map is Lipschitz on that interval. Hence,
\begin{align*}
    \|\exp(\widehat D)-\exp(D_0)\|_{L^2(P_{0X})}^2
    \le
    C\|\widehat D-D_0\|_{L^2(P_{0X})}^2,
\end{align*}
which proves \eqref{eq:ss-dre-rate}. For \(\kappa\in(0,1)\), the mean-type representer is \(\alpha_{0,\kappa}=\kappa+(1-\kappa)\exp(D_0)\), and the estimator \(\widehat\alpha_\kappa=\kappa+(1-\kappa)\exp(\widehat D)\) satisfies
\begin{align*}
    \|\widehat\alpha_\kappa-\alpha_{0,\kappa}\|_{P,2}^2
    =
    (1-\kappa)^2\|\exp(\widehat D)-\exp(D_0)\|_{P,2}^2.
\end{align*}
This gives the stated extension.

\subsection{Proof of Theorem~\ref{thm:ss-unbounded-support-rate}}
\label{app:proof-ss-unbounded-support-rate}

Let \(\calX_N=\{x\colon\|x\|_\infty\le C_1\log N_{\min}\}\). Decompose the squared error as
\begin{align*}
    \|\widehat\alpha-\alpha_{0,\kappa}\|_{P,2}^2
    =
    \bbE_{P_{0X}}\sqb{(\widehat\alpha(X)-\alpha_{0,\kappa}(X))^2\mathbbm{1}[X\in\calX_N]}
    +
    \bbE_{P_{0X}}\sqb{(\widehat\alpha(X)-\alpha_{0,\kappa}(X))^2\mathbbm{1}[X\notin\calX_N]}.
\end{align*}
On \(\calX_N\), rescale covariates to \([0,1]^d\). The H\"{o}lder norm after rescaling grows only by a logarithmic factor because the side length of \(\calX_N\) is of order \(\log N_{\min}\). Therefore, the approximation argument in Theorem~\ref{thm:ss-relu-rate} gives the same polynomial order with a larger logarithmic exponent. On \(\calX_N^c\), clipping gives a bounded envelope for \(\widehat\alpha-\alpha_{0,\kappa}\), and the tail assumption \eqref{eq:ss-tail-condition} gives
\begin{align*}
    \bbE_{P_{0X}}\sqb{(\widehat\alpha(X)-\alpha_{0,\kappa}(X))^2\mathbbm{1}[X\notin\calX_N]}
    \le
    C N_{\min}^{-2s/(d+2s)}.
\end{align*}
The same tail bound controls the unlabeled empirical-process part over \(Q_{0X}\). Substituting the truncated approximation bound, the pseudo-dimension bound, and the tail bound into Corollary~\ref{cor:ss-grr-fast-rate} gives \eqref{eq:ss-unbounded-support-rate}.

\subsection{Proof of Theorem~\ref{thm:ss-manifold-rate}}
\label{app:proof-ss-manifold-rate}

Under the approximate-manifold condition, the regressor distribution is concentrated on a \(\varrho\)-neighborhood of a compact \(d_M\)-dimensional manifold. A random or deterministic embedding with distortion \(\delta\) maps this neighborhood into a Euclidean space of dimension \(d_\delta\), preserving distances up to the prescribed relative error on the relevant support. Since \(f_0\) is H\"{o}lder, the embedded function remains H\"{o}lder with constants changed only by the distortion and manifold regularity constants. A ReLU network on the embedded coordinates therefore attains
\begin{align*}
    \|f_N^*-f_0\|_\infty^2
    \le
    C N_{\min}^{-2s/(d_\delta+2s)}\log^{c_1}N_{\min}.
\end{align*}
The neighborhood radius condition ensures that the error from replacing the original support by the embedded approximation is no larger than the approximation scale above. The pseudo-dimension of the resulting network is balanced so that
\begin{align*}
    V_n\log(n+m)\p{\frac{1}{n}+\frac{1}{m}}
    \le
    C N_{\min}^{-2s/(d_\delta+2s)}\log^{c_2}N_{\min}.
\end{align*}
The Lipschitz property of \((\partial g)^{-1}\) transfers the approximation bound from \(f_0\) to \(\alpha_{0,\kappa}\). Substitution into Corollary~\ref{cor:ss-grr-fast-rate} gives \eqref{eq:ss-manifold-rate}.

\subsection{Proof of Corollary~\ref{cor:ss-product-rate}}
\label{app:proof-ss-product-rate}

The two nuisance rates above imply
\begin{align*}
    \|\widehat\alpha-\alpha_{0,\kappa}\|_{P,2}
    \|\widehat\gamma-\gamma_0\|_{P,2}
    =
    O_p\p{
        N^{-a}\log^{c_\alpha+c_\gamma}N
    },
\end{align*}
where
\begin{align*}
    a=
    \frac{s_\alpha}{d_\alpha+2s_\alpha}
    +
    \frac{s_\gamma}{d_\gamma+2s_\gamma}.
\end{align*}
If \eqref{eq:ss-product-smoothness-condition} holds, then \(a>1/2\). Therefore, for a sufficiently small \(\eta>0\), it holds that \(a\ge1/2+\eta\). Polynomial decay \(N^{-\eta}\) dominates the logarithmic factor, and thus
\begin{align*}
    N^{1/2}
    \|\widehat\alpha-\alpha_{0,\kappa}\|_{P,2}
    \|\widehat\gamma-\gamma_0\|_{P,2}
    =o_p(1).
\end{align*}
This is exactly the product-rate condition in Assumption~\ref{asm:conv_rate}.

\section{Proofs for Section~\ref{sec:one-sample}}
\label{app:one-sample}

\subsection{Proof of Theorem~\ref{thm:os-eif}}

We work directly with the observed-data model. The observed data are \(O=(X,S,Y)\), where \(Y=SY^*+(1-S)\mathrm{NA}\). Under missing at random, the observed-data law can be decomposed into the marginal law of \(X\), the labeling law of \(S\) given \(X\), and the conditional law of \(Y^*\) given \(X\), which is observed only when \(S=1\). Along a regular parametric submodel, write the observed-data score as
\begin{align*}
    s_O(O)=s_X(X)+s_{S\mid X}(S,X)+S s_{Y\mid X}(Y^*,X),
\end{align*}
where
\begin{align*}
    \bbE\sqb{s_X(X)}=0,
    \qquad
    \bbE\sqb{s_{S\mid X}(S,X)\mid X}=0,
    \qquad
    \bbE\sqb{s_{Y\mid X}(Y^*,X)\mid X}=0.
\end{align*}
The term \(S s_{Y\mid X}(Y^*,X)\) is observable because it is zero when \(S=0\), and when \(S=1\) the observed outcome satisfies \(Y=Y^*\). This is the observed-data score decomposition associated with the censoring setting.

Let \(\gamma_t(x)=\bbE_t[Y^*\mid X=x]\). The labeling law does not enter this conditional mean. Differentiating the conditional mean under the conditional outcome score gives
\begin{align}
    \dot\gamma_0(x)
    &\coloneqq
    \frac{\partial}{\partial t}\gamma_t(x)|_{t=0}
    =
    \bbE\sqb{\p{Y^*-\gamma_0(X)}s_{Y\mid X}(Y^*,X)\mid X=x}.
    \label{eq:app-os-gamma-derivative-updated}
\end{align}
The target along the submodel is
\begin{align*}
    \theta_t^{\mathrm{OS}}=\bbE_{P_{tX}}\sqb{m(X,\gamma_t)}.
\end{align*}
Because \(\gamma\mapsto m(X,\gamma)\) is linear, differentiating at zero yields
\begin{align}
    \frac{\partial}{\partial t}\theta^{\mathrm{OS}}_t\mid_{t=0}
    =
    \bbE\sqb{m(X,\gamma_0)s_X(X)}
    +
    \bbE\sqb{\alpha_0^{\mathrm{OS}}(X)\dot\gamma_0(X)},
    \label{eq:app-os-param-derivative-updated}
\end{align}
where the second term follows from the Riesz representer theorem \eqref{eq:os-riesz}.

We now verify that \(\psi_0^{\mathrm{OS}}\) represents this derivative. Since \(m(X,\gamma_0)-\theta^{\mathrm{OS}}_0\) is a function of \(X\),
\begin{align}
    &\bbE\sqb{\p{m(X,\gamma_0)-\theta^{\mathrm{OS}}_0}s_O(O)}
    \nonumber\\
    &\quad=
    \bbE\sqb{\p{m(X,\gamma_0)-\theta^{\mathrm{OS}}_0}s_X(X)}
    =
    \bbE\sqb{m(X,\gamma_0)s_X(X)}.
    \label{eq:app-os-x-score-updated}
\end{align}
The terms involving \(s_{S\mid X}\) and \(s_{Y\mid X}\) vanish after conditioning on \(X\).

Next, consider the inverse-labeling residual term. Since \(\bbE\sqb{Y^*-\gamma_0(X)\mid X}=0\), the marginal covariate score contributes zero. The labeling score also contributes zero. Indeed, by missing at random, conditional on \(X\), the residual \(Y^*-\gamma_0(X)\) is independent of the labeling score component, and therefore
\begin{align*}
    \bbE\sqb{\frac{S}{\pi_0(X)}\alpha_0^{\mathrm{OS}}(X)\p{Y^*-\gamma_0(X)}s_{S\mid X}(S,X)\mid X}
    =
    \frac{\alpha_0^{\mathrm{OS}}(X)}{\pi_0(X)}\bbE\sqb{S s_{S\mid X}(S,X)\mid X}\bbE\sqb{Y^*-\gamma_0(X)\mid X}
    =0.
\end{align*}
Thus, it holds that
\begin{align}
    &\bbE\sqb{\frac{S}{\pi_0(X)}\alpha^{\mathrm{OS}}_0(X)\p{Y-\gamma_0(X)}s_O(O)}
    \nonumber\\
    &\quad=
    \bbE\sqb{\frac{S}{\pi_0(X)}\alpha^{\mathrm{OS}}_0(X)\p{Y^*-\gamma_0(X)}S s_{Y\mid X}(Y^*,X)}
    \nonumber\\
    &\quad=
    \bbE\sqb{\alpha^{\mathrm{OS}}_0(X)\p{Y^*-\gamma_0(X)}s_{Y\mid X}(Y^*,X)}
    \nonumber\\
    &\quad=
    \bbE\sqb{\alpha^{\mathrm{OS}}_0(X)\dot\gamma_0(X)}.
    \label{eq:app-os-outcome-score-updated}
\end{align}
The second equality uses \(S^2=S\), \(\bbE\sqb{S\mid X}=\pi_0(X)\), and missing at random.

Combining \eqref{eq:app-os-param-derivative-updated}, \eqref{eq:app-os-x-score-updated}, and \eqref{eq:app-os-outcome-score-updated}, it holds that
\begin{align*}
    \frac{\partial}{\partial t}\theta^{\mathrm{OS}}_t\mid_{t=0}
    =
    \bbE\sqb{\psi_0^{\mathrm{OS}}(O)s_O(O)}.
\end{align*}
The proposed influence function is centered because
\begin{align*}
    \bbE\sqb{\frac{S}{\pi_0(X)}\alpha_0^{\mathrm{OS}}(X)\p{Y-\gamma_0(X)}\mid X}=0
\end{align*}
and \(\bbE\sqb{m(X,\gamma_0)-\theta_0^{\mathrm{OS}}}=0\). The observed-data tangent space under the nonparametric missing-at-random model is generated by sums of the three score components given above. The proposed function belongs to this tangent space: its covariate part is \(m(X,\gamma_0)-\theta_0^{\mathrm{OS}}\), its labeling-score component is zero, and its conditional-outcome component is represented by \(S\alpha_0^{\mathrm{OS}}(X)\p{Y^*-\gamma_0(X)}/\pi_0(X)\), whose conditional mean given \(X\) is zero. Since it represents the pathwise derivative for every regular score and lies in the tangent space, it is the canonical gradient. This proves Theorem~\ref{thm:os-eif}.

\subsection{Proof of Theorem~\ref{thm:os-eff-bound}}

By Theorem~\ref{thm:os-eif}, the semiparametric efficiency bound under \(\sqrt N\) normalization is the second moment of the canonical gradient. Write
\begin{align*}
    \psi_0^{\mathrm{OS}}(O)=A(O)+B(X),
\end{align*}
where
\begin{align*}
    A(O)=\frac{S}{\pi_0(X)}\alpha_0^{\mathrm{OS}}(X)\p{Y-\gamma_0(X)},
    \qquad
    B(X)=m(X,\gamma_0)-\theta_0^{\mathrm{OS}}.
\end{align*}
Since \(S\p{Y-\gamma_0(X)}=S\p{Y^*-\gamma_0(X)}\),
\begin{align*}
    \bbE\sqb{A(O)\mid X}
    =
    \frac{\alpha_0^{\mathrm{OS}}(X)}{\pi_0(X)}
    \bbE\sqb{S\p{Y^*-\gamma_0(X)}\mid X}
    =0.
\end{align*}
Thus, the cross term between \(A(O)\) and \(B(X)\) is zero, and
\begin{align*}
    \bbE\sqb{\psi_0^{\mathrm{OS}}(O)^2}
    =
    \bbE\sqb{A(O)^2}+\bbE\sqb{B(X)^2}.
\end{align*}
For the first term, we have
\begin{align*}
    \bbE\sqb{A(O)^2\mid X}
    &=
    \frac{\p{\alpha_0^{\mathrm{OS}}(X)}^2}{\pi_0(X)^2}
    \bbE\sqb{S\p{Y^*-\gamma_0(X)}^2\mid X}
    \nonumber\\
    &=
    \frac{\p{\alpha_0^{\mathrm{OS}}(X)}^2}{\pi_0(X)^2}
    \pi_0(X)\sigma_0^2(X)
    =
    \frac{\p{\alpha_0^{\mathrm{OS}}(X)}^2\sigma_0^2(X)}{\pi_0(X)}.
\end{align*}
For the second term, we have
\begin{align*}
    \bbE\sqb{B(X)^2}=
    \operatorname{Var}_{P_{0X}}\p{m(X,\gamma_0)}.
\end{align*}
Combining these equations proves \eqref{eq:os-eff-bound}.

\subsection{Proof of Theorem~\ref{thm:os-consistency}}

For fixed nuisance functions, the population drift of the one-sample EE-DML-PPCI score is
\begin{align*}
    \bbE_{P_0}\sqb{
        \p{
            \alpha^{\mathrm{OS}}_0(X)
            -
            \frac{\pi_0(X)}{\widehat\pi(X)}
            \widehat\alpha(X)
        }
        \p{\widehat\gamma(X)-\gamma_0(X)}
    }.
\end{align*}
If Assumption~\ref{ass:os-reg-consistency} holds and the remaining nuisance components are bounded in the required $L^2(P_{0X})$ norms, this drift is $o_p(1)$. If Assumption~\ref{ass:os-riesz-pi-consistency} holds and $\widehat\gamma$ is bounded in $L^2(P_{0X})$, the same conclusion follows. The empirical average converges to its population counterpart by the law of large numbers under the stated moment and overlap conditions. Hence, $\widehat\theta^{\mathrm{OS}}_{\mathrm{EE}}\xrightarrow{\rmp}\theta^{\mathrm{OS}}_0$.

\subsection{Proof of Theorem~\ref{thm:os-asymptotic}}

We give the proof for the cross-fitted implementation. This implementation verifies the empirical-process requirement in Assumption~\ref{ass:os-donsker-or-crossfit} without imposing Donsker-type stochastic equicontinuity conditions.

Let \(K\ge 2\) be fixed. Partition \([N]\) into folds \(\calI_1,\ldots,\calI_K\). For each fold \(k\), let \(\calI_{-k}\coloneqq [N]\setminus\calI_k\), and let \(\calT_k\) denote the observations indexed by \(\calI_{-k}\). Let \(\widehat\gamma_k\), \(\widehat\alpha_k\), and \(\widehat\pi_k\) be nuisance estimators constructed from \(\calT_k\). The fold-specific score is evaluated only on observations in \(\calI_k\). Under this implementation, the EE-DML-PPCI estimator is
\begin{align}
    \widehat\theta^{\mathrm{OS}}_{\mathrm{EE}}
    \coloneqq
    \frac{1}{N}\sum^K_{k=1}\sum_{i\in\calI_k}
    \p{
        \frac{S_i}{\widehat\pi_k(X_i)}
        \widehat\alpha_k(X_i)
        \p{Y_i-\widehat\gamma_k(X_i)}
        +
        m\p{X_i,\widehat\gamma_k}
    }.
    \label{eq:app-os-ee-crossfit}
\end{align}
Here, \(S_i\p{Y_i-\widehat\gamma_k(X_i)}\) is interpreted as \(S_i\p{Y_i^*-\widehat\gamma_k(X_i)}\), which is observed.

For fixed nuisance functions \(\gamma\), \(\alpha\), and \(\pi\), define
\begin{align*}
    \varphi^{\mathrm{OS}}(O;\gamma,\alpha,\pi)
    \coloneqq
    \frac{S}{\pi(X)}
    \alpha(X)
    \p{Y-\gamma(X)}
    +
    m\p{X,\gamma}.
\end{align*}
For each fold \(k\), define
\begin{align*}
    \widehat\varphi^{\mathrm{OS}}_k(O)
    &\coloneqq
    \varphi^{\mathrm{OS}}\p{O;\widehat\gamma_k,\widehat\alpha_k,\widehat\pi_k},
    \\
    \varphi^{\mathrm{OS}}_0(O)
    &\coloneqq
    \varphi^{\mathrm{OS}}\p{O;\gamma_0,\alpha^{\mathrm{OS}}_0,\pi_0}.
\end{align*}
Then,
\begin{align*}
    \varphi^{\mathrm{OS}}_0(O)-\theta^{\mathrm{OS}}_0
    =
    \psi^{\mathrm{OS}}_0(O).
\end{align*}
By adding and subtracting \(\varphi^{\mathrm{OS}}_0(O_i)\) and the conditional mean of the score difference on each fold, we obtain
\begin{align}
    \widehat\theta^{\mathrm{OS}}_{\mathrm{EE}}
    -
    \theta^{\mathrm{OS}}_0
    &=
    \frac{1}{N}\sum^N_{i=1}
    \psi^{\mathrm{OS}}_0(O_i)
    +
    R^{\mathrm{OS}}_{1N}
    +
    R^{\mathrm{OS}}_{2N},
    \label{eq:app-os-ee-expansion}
\end{align}
where
\begin{align}
    R^{\mathrm{OS}}_{1N}
    &\coloneqq
    \frac{1}{N}\sum^K_{k=1}\sum_{i\in\calI_k}
    \p{
        \widehat\varphi^{\mathrm{OS}}_k(O_i)
        -
        \varphi^{\mathrm{OS}}_0(O_i)
        -
        \bbE_{P_0}\sqb{
            \widehat\varphi^{\mathrm{OS}}_k(O)
            -
            \varphi^{\mathrm{OS}}_0(O)
            \mid
            \calT_k
        }
    },
    \label{eq:app-os-r1}
    \\
    R^{\mathrm{OS}}_{2N}
    &\coloneqq
    \frac{1}{N}\sum^K_{k=1}
    |\calI_k|
    \bbE_{P_0}\sqb{
        \widehat\varphi^{\mathrm{OS}}_k(O)
        -
        \varphi^{\mathrm{OS}}_0(O)
        \mid
        \calT_k
    }.
    \label{eq:app-os-r2}
\end{align}

We first control \(R^{\mathrm{OS}}_{2N}\). Conditional on \(\calT_k\), the nuisance estimators are fixed. Since \(Y=Y^*\) when \(S=1\) and \(Y\) appears only through \(S\p{Y-\widehat\gamma_k(X)}\), the missing-at-random condition gives
\begin{align}
    \bbE_{P_0}\sqb{
        \frac{S}{\widehat\pi_k(X)}
        \widehat\alpha_k(X)
        \p{Y-\widehat\gamma_k(X)}
        \mid
        X,\calT_k
    }
    =
    \frac{\pi_0(X)}{\widehat\pi_k(X)}
    \widehat\alpha_k(X)
    \p{\gamma_0(X)-\widehat\gamma_k(X)}.
    \label{eq:app-os-cond-mean}
\end{align}
Therefore,
\begin{align}
    &
    \bbE_{P_0}\sqb{
        \widehat\varphi^{\mathrm{OS}}_k(O)
        -
        \varphi^{\mathrm{OS}}_0(O)
        \mid
        \calT_k
    }
    \nonumber\\
    &=
    \bbE_{P_0}\sqb{
        \frac{\pi_0(X)}{\widehat\pi_k(X)}
        \widehat\alpha_k(X)
        \p{\gamma_0(X)-\widehat\gamma_k(X)}
        +
        m\p{X,\widehat\gamma_k}
        -
        m\p{X,\gamma_0}
        \mid
        \calT_k
    }.
    \label{eq:app-os-drift-1}
\end{align}
By the Riesz representation for the one-sample target,
\begin{align}
    \bbE_{P_0}\sqb{
        m\p{X,\widehat\gamma_k}
        -
        m\p{X,\gamma_0}
        \mid
        \calT_k
    }
    =
    \bbE_{P_0}\sqb{
        \alpha^{\mathrm{OS}}_0(X)
        \p{\widehat\gamma_k(X)-\gamma_0(X)}
        \mid
        \calT_k
    }.
    \label{eq:app-os-riesz}
\end{align}
Substituting \eqref{eq:app-os-riesz} into \eqref{eq:app-os-drift-1} yields
\begin{align}
    &
    \bbE_{P_0}\sqb{
        \widehat\varphi^{\mathrm{OS}}_k(O)
        -
        \varphi^{\mathrm{OS}}_0(O)
        \mid
        \calT_k
    }
    \nonumber\\
    &=
    \bbE_{P_0}\sqb{
        \p{
            \alpha^{\mathrm{OS}}_0(X)
            -
            \frac{\pi_0(X)}{\widehat\pi_k(X)}
            \widehat\alpha_k(X)
        }
        \p{
            \widehat\gamma_k(X)-\gamma_0(X)
        }
        \mid
        \calT_k
    }.
    \label{eq:app-os-drift-2}
\end{align}
By Cauchy--Schwarz,
\begin{align}
    &
    \left|
    \bbE_{P_0}\sqb{
        \widehat\varphi^{\mathrm{OS}}_k(O)
        -
        \varphi^{\mathrm{OS}}_0(O)
        \mid
        \calT_k
    }
    \right|
    \nonumber\\
    &\le
    \left\|
        \alpha^{\mathrm{OS}}_0
        -
        \frac{\pi_0\widehat\alpha_k}{\widehat\pi_k}
    \right\|_{P,2}
    \|\widehat\gamma_k-\gamma_0\|_{P,2}.
    \label{eq:app-os-drift-bound}
\end{align}
By the one-sample convergence rate condition, the right-hand side of \eqref{eq:app-os-drift-bound} is \(o_p(N^{-1/2})\) uniformly over \(k\). Since \(K\) is fixed and the folds are balanced, it follows that
\begin{align}
    R^{\mathrm{OS}}_{2N}
    =
    o_p(N^{-1/2}).
    \label{eq:app-os-r2-bound}
\end{align}

We next control \(R^{\mathrm{OS}}_{1N}\). Conditional on \(\calT_1,\ldots,\calT_K\), the observations in each held-out fold are independent of the nuisance estimators used on that fold, and each summand in \eqref{eq:app-os-r1} has conditional mean zero. Hence,
\begin{align}
    &
    \bbE_{P_0}\sqb{
        \p{R^{\mathrm{OS}}_{1N}}^2
        \mid
        \calT_1,\ldots,\calT_K
    }
    \nonumber\\
    &\le
    \frac{C}{N}
    \max_{1\le k\le K}
    \bbE_{P_0}\sqb{
        \p{
            \widehat\varphi^{\mathrm{OS}}_k(O)
            -
            \varphi^{\mathrm{OS}}_0(O)
        }^2
        \mid
        \calT_k
    },
    \label{eq:app-os-r1-variance}
\end{align}
where \(C\) is finite because \(K\) is fixed and the folds are balanced.

It remains to show that the conditional second moment in \eqref{eq:app-os-r1-variance} is \(o_p(1)\) uniformly over \(k\). We decompose
\begin{align}
    &
    \widehat\varphi^{\mathrm{OS}}_k(O)
    -
    \varphi^{\mathrm{OS}}_0(O)
    \nonumber\\
    &=
    S
    \p{
        \frac{\widehat\alpha_k(X)}{\widehat\pi_k(X)}
        -
        \frac{\alpha^{\mathrm{OS}}_0(X)}{\pi_0(X)}
    }
    \p{Y-\gamma_0(X)}
    \nonumber\\
    &\quad
    -
    \frac{S}{\widehat\pi_k(X)}
    \widehat\alpha_k(X)
    \p{\widehat\gamma_k(X)-\gamma_0(X)}
    +
    m\p{X,\widehat\gamma_k}
    -
    m\p{X,\gamma_0}.
    \label{eq:app-os-score-diff}
\end{align}
The first term in \eqref{eq:app-os-score-diff} is \(o_p(1)\) in conditional \(L^2(P_0)\) uniformly over \(k\), because
\begin{align*}
    \left\|
        \frac{\widehat\alpha_k}{\widehat\pi_k}
        -
        \frac{\alpha^{\mathrm{OS}}_0}{\pi_0}
    \right\|_{P,2}
    \le
    C
    \left\|
        \frac{\pi_0\widehat\alpha_k}{\widehat\pi_k}
        -
        \alpha^{\mathrm{OS}}_0
    \right\|_{P,2}
\end{align*}
with probability approaching one, and the residual has a uniformly bounded conditional moment. The second term in \eqref{eq:app-os-score-diff} is \(o_p(1)\) in conditional \(L^2(P_0)\), because \(\widehat\alpha_k/\widehat\pi_k\) has bounded second moment uniformly over folds and \(\|\widehat\gamma_k-\gamma_0\|_{P,2}=o_p(1)\). The third term is \(o_p(1)\) in \(L^2(P_0)\) by the convergence condition for \(m\p{\cdot,\widehat\gamma_k}\). Thus,
\begin{align}
    \max_{1\le k\le K}
    \bbE_{P_0}\sqb{
        \p{
            \widehat\varphi^{\mathrm{OS}}_k(O)
            -
            \varphi^{\mathrm{OS}}_0(O)
        }^2
        \mid
        \calT_k
    }
    =
    o_p(1).
    \label{eq:app-os-score-l2}
\end{align}
Combining \eqref{eq:app-os-r1-variance} and \eqref{eq:app-os-score-l2}, conditional Chebyshev's inequality gives
\begin{align}
    R^{\mathrm{OS}}_{1N}
    =
    o_p(N^{-1/2}).
    \label{eq:app-os-r1-bound}
\end{align}

Substituting \eqref{eq:app-os-r1-bound} and \eqref{eq:app-os-r2-bound} into \eqref{eq:app-os-ee-expansion}, we obtain
\begin{align}
    \widehat\theta^{\mathrm{OS}}_{\mathrm{EE}}
    -
    \theta^{\mathrm{OS}}_0
    =
    \frac{1}{N}\sum^N_{i=1}
    \psi^{\mathrm{OS}}_0(O_i)
    +
    o_p(N^{-1/2}).
    \label{eq:app-os-ee-al}
\end{align}
The observations \(O_1,\ldots,O_N\) are i.i.d. in the one-sample scenario. Moreover,
\begin{align*}
    \bbE_{P_0}\sqb{\psi^{\mathrm{OS}}_0(O)}
    =
    0,
    \qquad
    \bbE_{P_0}\sqb{\psi^{\mathrm{OS}}_0(O)^2}
    =
    V^{\mathrm{OS}}_0
    <
    \infty.
\end{align*}
Therefore, the central limit theorem yields
\begin{align}
    \sqrt N
    \p{
        \widehat\theta^{\mathrm{OS}}_{\mathrm{EE}}
        -
        \theta^{\mathrm{OS}}_0
    }
    \xrightarrow{d}
    N\p{0,V^{\mathrm{OS}}_0}.
    \label{eq:app-os-ee-clt}
\end{align}
This proves the assertion for EE-DML-PPCI.

We next prove the assertion for TMLE-DML-PPCI. Under the cross-fitted implementation, define
\begin{align}
    \widehat A^{\mathrm{OS}}_N
    &\coloneqq
    \frac{1}{N}\sum^K_{k=1}\sum_{i\in\calI_k}
    \frac{S_i}{\widehat\pi_k(X_i)}
    \widehat\alpha_k(X_i)
    \p{Y_i-\widehat\gamma_k(X_i)},
    \label{eq:app-os-tmle-A}
    \\
    \widehat D^{\mathrm{OS}}_N
    &\coloneqq
    \frac{1}{N}\sum^K_{k=1}\sum_{i\in\calI_k}
    \widehat\alpha_k^2\p{X_i},
    \label{eq:app-os-tmle-D}
    \\
    \widehat\varepsilon^{\mathrm{OS}}
    &\coloneqq
    \frac{\widehat A^{\mathrm{OS}}_N}{\widehat D^{\mathrm{OS}}_N}.
    \label{eq:app-os-tmle-epsilon}
\end{align}
The updated regression function on fold \(k\) is
\begin{align}
    \widehat\gamma^{(1)}_k(x)
    \coloneqq
    \widehat\gamma_k(x)
    +
    \widehat\varepsilon^{\mathrm{OS}}\widehat\alpha_k(x).
    \label{eq:app-os-tmle-update}
\end{align}
The corresponding TMLE-DML-PPCI estimator is
\begin{align}
    \widehat\theta^{\mathrm{OS}}_{\mathrm{TMLE}}
    \coloneqq
    \frac{1}{N}\sum^K_{k=1}\sum_{i\in\calI_k}
    m\p{X_i,\widehat\gamma^{(1)}_k}.
    \label{eq:app-os-tmle-estimator}
\end{align}
Because \(\gamma\mapsto m(X,\gamma)\) is linear,
\begin{align}
    m\p{X_i,\widehat\gamma^{(1)}_k}
    =
    m\p{X_i,\widehat\gamma_k}
    +
    \widehat\varepsilon^{\mathrm{OS}}
    m\p{X_i,\widehat\alpha_k}.
    \label{eq:app-os-tmle-linearity}
\end{align}
Therefore,
\begin{align}
    \widehat\theta^{\mathrm{OS}}_{\mathrm{TMLE}}
    &=
    \frac{1}{N}\sum^K_{k=1}\sum_{i\in\calI_k}
    m\p{X_i,\widehat\gamma_k}
    +
    \frac{\widehat A^{\mathrm{OS}}_N}{\widehat D^{\mathrm{OS}}_N}
    \widehat L^{\mathrm{OS}}_N,
    \label{eq:app-os-tmle-expansion}
\end{align}
where
\begin{align}
    \widehat L^{\mathrm{OS}}_N
    \coloneqq
    \frac{1}{N}\sum^K_{k=1}\sum_{i\in\calI_k}
    m\p{X_i,\widehat\alpha_k}.
    \label{eq:app-os-tmle-L}
\end{align}
On the other hand, by \eqref{eq:app-os-ee-crossfit},
\begin{align}
    \widehat\theta^{\mathrm{OS}}_{\mathrm{EE}}
    =
    \widehat A^{\mathrm{OS}}_N
    +
    \frac{1}{N}\sum^K_{k=1}\sum_{i\in\calI_k}
    m\p{X_i,\widehat\gamma_k}.
    \label{eq:app-os-ee-for-tmle}
\end{align}
Subtracting \eqref{eq:app-os-ee-for-tmle} from \eqref{eq:app-os-tmle-expansion} gives
\begin{align}
    \widehat\theta^{\mathrm{OS}}_{\mathrm{TMLE}}
    -
    \widehat\theta^{\mathrm{OS}}_{\mathrm{EE}}
    &=
    \widehat A^{\mathrm{OS}}_N
    \p{
        \frac{\widehat L^{\mathrm{OS}}_N}{\widehat D^{\mathrm{OS}}_N}
        -
        1
    }
    \nonumber\\
    &=
    \widehat A^{\mathrm{OS}}_N
    \frac{
        \widehat L^{\mathrm{OS}}_N
        -
        \widehat D^{\mathrm{OS}}_N
    }{
        \widehat D^{\mathrm{OS}}_N
    }.
    \label{eq:app-os-tmle-ee-diff}
\end{align}
By the one-sample targeting calibration condition, \(\widehat A^{\mathrm{OS}}_N=O_p(N^{-1/2})\), \(\widehat L^{\mathrm{OS}}_N-\widehat D^{\mathrm{OS}}_N=o_p(1)\), and \(\widehat D^{\mathrm{OS}}_N\) is bounded away from zero with probability approaching one. Hence,
\begin{align}
    \widehat\theta^{\mathrm{OS}}_{\mathrm{TMLE}}
    -
    \widehat\theta^{\mathrm{OS}}_{\mathrm{EE}}
    =
    o_p(N^{-1/2}).
    \label{eq:app-os-tmle-ee-equivalence}
\end{align}
Combining \eqref{eq:app-os-ee-al} and \eqref{eq:app-os-tmle-ee-equivalence}, we have
\begin{align}
    \widehat\theta^{\mathrm{OS}}_{\mathrm{TMLE}}
    -
    \theta^{\mathrm{OS}}_0
    =
    \frac{1}{N}\sum^N_{i=1}
    \psi^{\mathrm{OS}}_0(O_i)
    +
    o_p(N^{-1/2}).
    \label{eq:app-os-tmle-al}
\end{align}
The central limit theorem again gives
\begin{align}
    \sqrt N
    \p{
        \widehat\theta^{\mathrm{OS}}_{\mathrm{TMLE}}
        -
        \theta^{\mathrm{OS}}_0
    }
    \xrightarrow{d}
    N\p{0,V^{\mathrm{OS}}_0}.
    \label{eq:app-os-tmle-clt}
\end{align}
This completes the proof of Theorem~\ref{thm:os-asymptotic}.

\end{document}